%% file: main.tex
\documentclass[nohyperref]{article}

\usepackage{microtype}
\usepackage{graphicx}
\usepackage{caption}
\usepackage{booktabs} 
\usepackage{subcaption}
\usepackage{adjustbox}

\usepackage{hyperref}

\usepackage[accepted]{icml2022}

\usepackage{amsmath}
\usepackage{amssymb}
\usepackage{mathtools}
\usepackage{amsthm}

\usepackage[capitalize,noabbrev]{cleveref}

\usepackage{wrapfig}

\usepackage{tikz}
\usepackage{pgfplots}
\pgfplotsset{compat=newest}
\usepgfplotslibrary{groupplots}

\usepackage{makecell}

\theoremstyle{plain}

\theoremstyle{definition}

\theoremstyle{remark}

\DeclareMathOperator{\DCA}{DCA}
\DeclareMathOperator{\wDCA}{wDCA}
\DeclareMathOperator{\gstats}{gstats}

\usepackage[textsize=tiny]{todonotes}
\usepackage[normalem]{ulem}

\definecolor{niceOrange}{rgb}{1, 0.6, 0}
\definecolor{niceOrange2}{rgb}{1, 0.49, 0}
\definecolor{kthGreen}{RGB}{98, 146, 46}
\definecolor{kthYellow}{RGB}{250,185, 25}
\definecolor{burgundy}{rgb}{0.5, 0.0, 0.13}

\DeclarePairedDelimiter\floor{\lfloor}{\rfloor}

\icmltitlerunning{GraphDCA Framework for Node Distribution Comparison}

\begin{document}

\twocolumn[
\icmltitle{GraphDCA -- a Framework for Node Distribution Comparison \\ in Real and Synthetic Graphs}



\icmlsetsymbol{equal}{*}
\icmlsetsymbol{second_equal}{$\dagger$}

\begin{icmlauthorlist}
\icmlauthor{Ciwan Ceylan}{equal,kth,seb}
\icmlauthor{Petra Poklukar}{equal,kth}
\icmlauthor{Hanna Hultin}{second_equal,kth,seb}
\icmlauthor{Alexander Kravchenko}{second_equal,kth}
\icmlauthor{Anastasia Varava}{}
\icmlauthor{Danica Kragic}{kth}
\end{icmlauthorlist}

\icmlaffiliation{kth}{KTH Royal Institute of Technology, Stockholm, Sweden}
\icmlaffiliation{seb}{SEB Group, Stockholm, Sweden}

\icmlcorrespondingauthor{Ciwan Ceylan}{ciwan@kth.se}
\icmlcorrespondingauthor{Petra Poklukar}{poklukar@kth.se}

\icmlkeywords{Graph Comparison, Node Structural Features, Graph Representation Learning, Graph Generative Models}

\vskip 0.3in
]



\printAffiliationsAndNotice{\icmlEqualContribution \icmlEqualSecondContribution} 

\begin{abstract}
We argue that when comparing two graphs, the distribution of node structural features is more informative than global graph statistics which are often used in practice, especially to evaluate graph generative models. Thus, we present GraphDCA -- a framework for evaluating similarity between graphs based on the alignment of their respective node representation sets. The sets are compared using a recently proposed method for comparing representation spaces, called Delaunay Component Analysis (DCA), which we extend to graph data. To evaluate our framework, we generate a benchmark dataset of graphs exhibiting different structural patterns and show, using three node structure feature extractors, that GraphDCA recognizes graphs with both similar and dissimilar local structure. We then apply our framework to evaluate three publicly available real-world graph datasets and demonstrate, using gradual edge perturbations, that GraphDCA satisfyingly captures gradually decreasing similarity, unlike global statistics. Finally, we use GraphDCA to evaluate two state-of-the-art graph generative models, NetGAN and CELL, and conclude that further improvements are needed for these models to adequately reproduce local structural features.  
\end{abstract}

\input{Sections/introduction.tex}

\input{Sections/related_work}

\input{Sections/graphdca_framework}
\input{Sections/experiments}

\input{Sections/discussion} 

\section*{Acknowledgements} 
This work has been supported by the Knut and Alice Wallenberg Foundation, Swedish Research Council and European Research Council. The authors thank Vladislav Polianskii, Miguel Vasco, Kambiz Ghoorchian and Ala Tarighati for their valuable suggestions and feedback.

\bibliography{references}
\bibliographystyle{icml2022}

\newpage
\appendix
\onecolumn
\input{Sections/appendix}

\end{document}

%% file: Sections/introduction.tex
\section{Introduction}


 

Currently, there is no universally accepted framework for comparing graphs in terms of the distributions of their local structural properties, often referred to as node roles \cite{rossi_role_2015}.
In particular, evaluation of graph generative models, which have potential uses in data anonymization and genetic bootstrap analysis, constitutes a difficult open problem 
since it is generally not possible to evaluate the quality of generated graphs using human intuition, e.g.\ via visual inspection as done 
for images \cite{salimans2016improved, fid_heusel17, prec_recall_sajjadi18}.
To evaluate generative models for graphs, researchers have resorted to comparison of global graph statistics \cite{bojchevski_netgan_2018, rendsburg_netgan_2020}.
However, as pointed out already by \citet{bojchevski_netgan_2018}, global statistics are not sufficient if one is also concerned about similarity in the local structures: two locally different graphs may still have similar global properties due to the averaging across all the nodes. 
\begin{figure}[t]
  \begin{center}
    \includegraphics[width=0.48\textwidth]{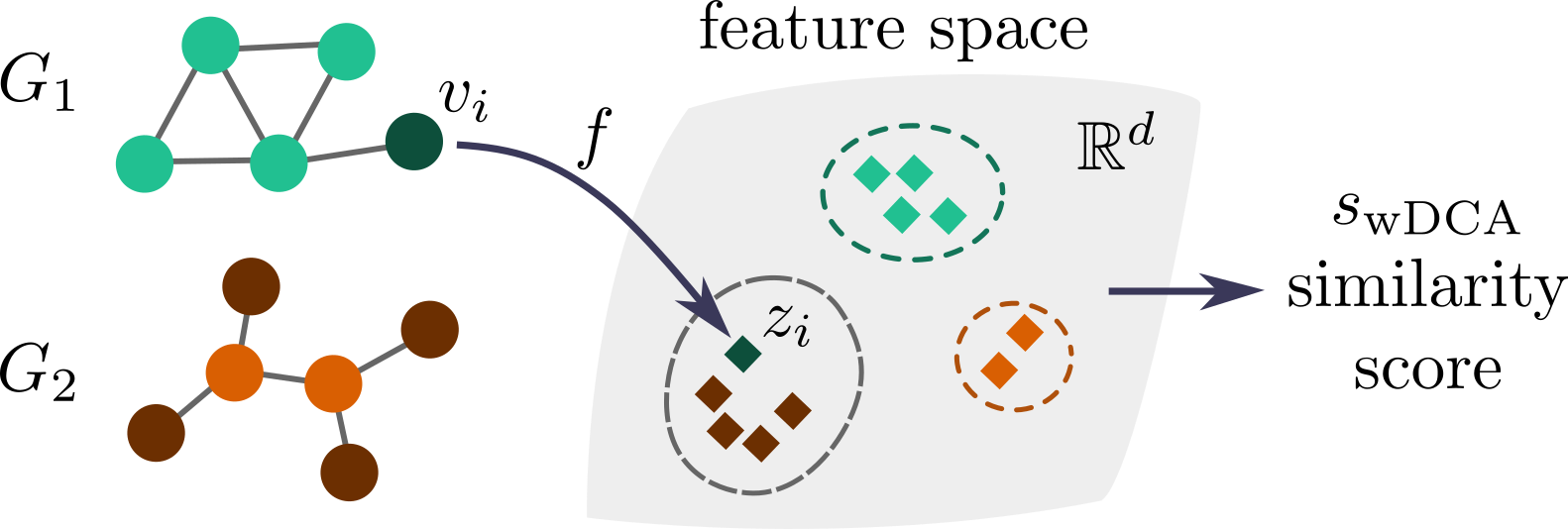}
  \end{center}
  \caption{Overview of the GraphDCA framework which compares two input graphs, $G_1$ and $G_2$ by calculating local structural similarity score $s_{\wDCA}$ based on representations $z_i$ corresponding to their nodes $v_i$ obtained by a feature extractor model $f$. }
\label{fig:graph_dca}
\vskip -0.2in
\end{figure} 

Comparing node role distributions is challenging as they are typically not uniform across different roles. 
In particular, the presence of rare roles is common in real-world graphs and important for their analysis \cite{powerlaw_Faloutsos_11, henderson_rolx_2012}. 
For instance, in social networks there usually exist "hub" nodes with high centrality whose presence is a crucial property of the graph's topology, while only constituting a small fraction of the total number of nodes \cite{newman2018networks}.
It is therefore necessary for similarity scores between graphs to account for different role importance as defined by downstream applications.

In this work, we present the \emph{GraphDCA} evaluation framework, visualized in Figure~\ref{fig:graph_dca}, for comparison of two graphs through the analysis of their respective node role representation sets.
GraphDCA takes inspiration from the recently proposed Delaunay Component Analysis (DCA) method \cite{anonymous2022delaunay} which compares two sets of image representations by analyzing the topological and geometric properties of manifolds described by the two sets.
To overcome the challenge of unbalanced node role distributions and promote rare roles in the comparison analysis, we extend their scores to take into account weights of node representations.


We show that the GraphDCA framework
can distinguish local structural similarities of undirected graphs without attributes when combined with a feature extractor capable of capturing such local structures. 
We construct a benchmark dataset of graphs, called \textsc{Groletest},  exhibiting qualitatively different connectivity patterns and thoroughly evaluate GraphDCA on it. To extract node representations, we combine GraphDCA with two state-of-the-art node role representation learning methods, GraphWave \cite{donnat_learning_2018} and Graph Contrastive Coding (GCC)~\cite{qiu_gcc_2020}.
We then apply gradual edge perturbation on three real-world graphs and demonstrate that GraphDCA favorably captures the decreasing similarity trend compared to global statistics which can exhibit large variance over different graphs and statistics. 
Furthermore, we employ GraphDCA to assess the quality of two recent graph generative models, NetGAN~\cite{bojchevski_netgan_2018} and CELL~\cite{rendsburg_netgan_2020}, trained on both \textsc{Groletest} and real-world data and demonstrate that these models struggle to capture local structure of the training graph. 
Our findings suggest that improvements are necessary for these models to fully reproduce graphs' local structural properties, in which case GraphDCA can serve as a complementary evaluation to global statistics.

%% file: Sections/related_work.tex
\section{Related Work}

\subsection{Graph Comparison}
Traditionally, definitions of graph similarity are based on the notion of graph isomorophism such as edit distance \cite{Sanfeliu_graph_edit83, gao2010survey} and maximum common subgraph \cite{BUNKE1998255}. 
These methods are prohibitively computationally expensive and reflect how different two graphs are, e.g.how difficult it is to obtain one graph by editing another, by finding a one-to-one correspondence between nodes, which is itself a complex problem. 
Instead, we are interested in comparing the distributions of different node roles capturing local structural properties.


Graph kernels \cite{vishwanathan2010graph, shervashidze2011weisfeiler}
are typically used in graph classification tasks involving a set of graphs, often with node attributes or labels \cite{wwl_graph_kernel_togninalli19, chen2020convolutional}.
Such kernels do not explicitly relate graph similarity to the distribution of local structural properties.
This is unlike the modular approach of GraphDCA which measures similarity in terms of different node representations.
For a recent overview of graph kernels, see \cite{nikolentzos2021graph}.

Graphlets are sometimes used in molecular applications to characterize properties of graphs and model node roles. A graphlet is a connected subgraph induced on a, usually tiny, set of nodes. Several methods for enumerating graphlets of certain size allow to characterize undirected~\cite{orbits_hocevar14} and directed~\cite{sarajlic2016graphlet} graphs. The number of different graphlets grows exponentially with the number of nodes, limiting the practical size of the considered subgraphs. In addition, the set of graphlets has to be selected manually which is meaningful in applications like chemistry where graphs are \emph{known} to be formed of certain building blocks. In contrast, we are interested in roles that do not need to be defined in advance and are \emph{automatically} extracted from data.




\subsection{Node Representations} \label{sec:rw:fe}
Graph neural networks, such as \cite{KipfW17, graphsage_hamilton17, velickovic2018graph, chiang2019cluster}, excel in producing node representations for attributed graphs \cite{hu2020ogb}.
In this work, we consider undirected graphs without attributes for which one finds two categories of node representations, capturing either neighborhood or structural similarity.
The former, e.g.\ \cite{perozzi_deepwalk_2014, grover_node2vec_2016, qiu_network_2018}, are typically learned \emph{per graph}, and can therefore not be used in GraphDCA without modification.
Instead, we focus on node representations which capture structural similarity, i.e.\ node roles, and can be compared across different graphs \cite{henderson_its_2011, henderson_rolx_2012, ribeiro2017struc2vec, donnat_learning_2018, qiu_gcc_2020}.


\subsection{Evaluation of Generative Models for Graphs}

Recently, researchers have proposed several enhancements to the general approach of comparing real and generated data in a representation space, e.g.\ defining precision and recall scores \cite{prec_recall_sajjadi18, improved_prec_recall19} as well as comparing geometric and topological properties of the representation distributions \cite{khrulkov2018geometry, pmlr-v139-poklukar21a, anonymous2022delaunay}.
In this work, we adapt the method proposed by \cite{anonymous2022delaunay}, shown to outperform its predecessors, for comparing graphs.

Generative models for graphs are typically evaluated either by comparing global graph statistics, e.g.\ assortativity coefficients and motif counts, or by comparing graph descriptors consisting of a few scalar properties via Maximum Mean Discrepancy (MMD). 
Global statistics have been used for models trained using a single input graph \cite{bojchevski_netgan_2018, rendsburg_netgan_2020}, while MMD has been adopted in cases when several input graphs are considered~\cite{you2018graphrnn, gran_liao19, obray2021evaluation}. 
In this work, we are mainly concerned with comparing two large graphs rather than two sets of smaller graphs which is why we compare with global statistics in our experiments.

%% file: Sections/graphdca_framework.tex
\section{GraphDCA Evaluation Framework} \label{sec:gdca}

The GraphDCA evaluation framework, shown in Figure~\ref{fig:graph_dca}, compares two graphs by measuring the similarity of their local structural properties and consists of two main components:
\begin{itemize}
    \item a \textit{feature extractor model} $f$ that automatically captures local structural features of the graphs as node representations, thus eliminating the need for manual feature engineering, and
    \item \textit{evaluation scores} that reflect the similarity of node representation sets and can prioritize specific node roles.
\end{itemize}

To analyze the representations we extend the recently proposed Delaunay Component Analysis (DCA)~\cite{anonymous2022delaunay} method 
to graphs such that the comparison can account for node role importance. 
We present rigorous definitions of the weighted evaluation scores in Section~\ref{sec:gdca:evaluation_scores}. GraphDCA can be combined with any feature extractor model $f$ that maps nodes from different graphs to the same representation space.
We consider two recently proposed models described in Section~\ref{sec:gdca:feature_extractors}.
Since the final choice of $f$ depends on data and application in consideration, we additionally contribute with a synthetic dataset of graphs, called \textsc{Groletest}, exhibiting known local structures which can be used as a validation of the GraphDCA framework. We describe the generation details of \textsc{Groletest} in Section~\ref{sec:gdca:groletest}. 
GraphDCA inherits the hyperparameters of its two components for which we use the recommended default choices (see Appendix~\ref{sec:app:hyperparams}).

\subsection{Feature Extractors for Node Representations}
\label{sec:gdca:feature_extractors}

Given two input graphs $G_k = (V_k, E_k)$ with vertex sets $V_k$ and edge sets $E_k$ for $k = 1, 2$, we evaluate their similarity in terms of their node representation sets $f(G_k) = \{z_i = f(v_i) \,|\, v_i \in V_k\}$ obtained with a feature extractor $f: V_k \to  \mathbb{R}^d$. 
Since we are interested in representations which capture node roles in undirected graphs without node attributes, we have selected two of the latest works able to produce such representations, GraphWave \cite{donnat_learning_2018} and Graph Contrastive Coding (GCC) \cite{qiu_gcc_2020}.
We also include a set of manually defined features for comparison.


{\bf GraphWave} produces a node representation from the diffusion of a spectral graph wavelet centered at the node. 
Representations are constructed from these wavelets deterministically by treating them as probability distributions on the graph and characterizing the distributions using the corresponding empirical characteristic functions. 


{\bf GCC} uses contrastive learning to learn representations of node roles. Data instances representing a node are formed by constructing subgraphs using random-walk-based graph sampling initialized at the node. 
A graph neural network encodes these subgraphs into a latent space such that representations of instances originating from the same node are encoded closeby. GCC is pre-trained on a diverse set of graphs to produce general representations which can be used for downstream tasks on other graphs.

\textbf{Manual features.} As a baseline model $f$, we use manually constructed features. 
These are commonly used graph statistics calculated for each node's $\rho$-egonet defined as the induced subgraph of all nodes within shortest path distance $\rho$, as previously used for $\rho=1$ \cite{everett1990ego, akoglu_oddball_2010}. Since different radii likely expose different local structures, we compute and concatenate features for $\rho \in \{1, 2, 3, 4\}$. 
See Appendix~\ref{sec:app:hyperparams} for the list of used statistics.

\subsection{Weighted DCA Evaluation Scores} \label{sec:gdca:evaluation_scores}


We compare two sets of node representations $R_k = f(G_k)$ for $k = 1,2$ given by a feature extractor model $f$ using DCA which we extend to graph data. In the original formulation of DCA, each point from $R_1$ and $R_2$ always contributes equally to the total similarity score between the two sets. While this is a reasonable assumption for the class-balanced image datasets considered by \cite{anonymous2022delaunay}, it is problematic in the context of network analysis where distributions of different node roles are not uniform and there often exist rare roles important for the analysis. To perform the comparison of graphs based on such specific nodes roles, we extend the DCA evaluation scores to weighted representations, which we refer to as \textit{weighted DCA}.

The idea of DCA is to compare the sets $R_k$ by analyzing the connected components of the manifold represented by $R_1 \cup R_2$.
Intuitively, if $R_1$ and $R_2$ contribute to each connected component equally and homogeneously they are considered similar.
The manifold $R_1 \cup R_2$ is approximated by a so-called ``distilled'' Delaunay graph. First, a Delaunay graph $\mathcal{D}$ is built to efficiently capture the neighborhood structure of the representation space. Since $\mathcal{D}$ is always connected by construction, a clustering algorithm is applied to group the points into clusters $\{\mathcal{D}_c\}$ representing the connected components (see~\cite{anonymous2022delaunay} for details). The obtained connected components $\{\mathcal{D}_c\}$ are then analyzed in terms of $R_1$ and $R_2$ points and edges among them expressed in several evaluation scores. 


We denote by $\mathcal{H} = (\mathcal{V}, \mathcal{E})$ a \emph{latent} graph built on a subset of representations $R_1 \cup R_2$ of given \emph{input} graphs $G_1, G_2$. Let $|\mathcal{H}|_{\mathcal{E}}$ denote the cardinality of the edge set $\mathcal{E}$, and let $\mathcal{H}^{R_k} = (\mathcal{V}|_{R_k}, \mathcal{E}_{R_k \times R_k})$ be the subgraph of $\mathcal{H}$ induced by the set $R_k$ for $k=1,2$. Let $w: V_1 \cup V_2 \longrightarrow \mathbb{R}$ denote the weights defined on the input graphs $G_k$ which are further inherited to their representations $R_k$ and consequentially to the vertex set $\mathcal{V}$ of any latent graph $\mathcal{H}$, i.e., $w(\mathcal{H}) = \sum_{z_i \in \mathcal{V}} w(z_i)$.

The weighted DCA consists of three scores: \textit{network quality} which summarizes the total geometric alignment of $R_1$ and $R_2$ across all $\mathcal{D}_c$, and \textit{weighted precision and recall} which reflect the ratio of $R_2$ and $R_1$, respectively, contained in balanced and geometrically well aligned components. These scores are derived from scores assessing the quality of each connected component $\mathcal{D}_c$. Intuitively, if points from $R_1$ and $R_2$ are geometrically well aligned in $\mathcal{D}_c$, then the ratio of latent edges between $R_1$ and $R_2$ contained in $\mathcal{D}_c$ is high compared to the total number of latent edges. This is measured by \textit{component quality} $q(\mathcal{D}_c) \in [0, 1]$ defined as  
\begin{align*}
    q(\mathcal{D}_c) = \begin{cases}
    1 -  \frac{(|\mathcal{D}_c^{R_1}|_\mathcal{E} + |\mathcal{D}_c^{R_2}|_\mathcal{E})}{|\mathcal{D}_c|_\mathcal{E}} & \text{if } |\mathcal{D}_c|_\mathcal{E} \geq 1,\\
    0              & \text{otherwise.}
    \end{cases}
\end{align*}
The definition directly generalizes to the network quality $q(\mathcal{D})$. 
However, $q(\mathcal{D})$ itself does not provide insights into the position of the node representations, for example, to assess how many are contained in 
components of high quality. This is measured by the weighted precision $p_w \in [0, 1]$ and weighted recall $r_w \in [0, 1]$ defined as 
\begin{align*}
   p_w = \frac{\sum_{\mathcal{D}_c \in \mathcal{F}}w(\mathcal{D}_c^{R_2})}{w(\mathcal{D}^{R_2})} \quad \text{and} \quad r_w = \frac{\sum_{\mathcal{D}_c \in \mathcal{F}}w(\mathcal{D}_c^{R_1})}{w(\mathcal{D}^{R_1})},
\end{align*}
where $\mathcal{F}$ denotes the union of all high-quality $\mathcal{D}_c$ here defined as $q(\mathcal{D}_c) > 0$. 
Finally, since all the scores are normalized and increasing with similarity, we define the \textit{weighted DCA score} as their harmonic mean $H$
\begin{align}
   s_{\wDCA}(R_1, R_2) = H (p_w, r_w, q).  \label{eq:dca_similarity}
\end{align}

Note that by using uniform weights, we recover the definitions given by~\cite{anonymous2022delaunay}.

\subsection{\textsc{Groletest} Synthetic Dataset} \label{sec:gdca:groletest}

\label{sec:synthetic-dataset}
Since the choice of the feature extractor model $f$ used in GraphDCA can be flexibly adjusted by the practitioner, we propose a dataset of synthetically generated graphs serving as a validation of the final evaluation framework. The \textsc{Groletest} dataset (derived from \textit{Graph ``Role'' Testing}) is comprised of plain, undirected and unweighted graphs having different local structures described below.

Each synthetic graph consists of a main graph $m$ into which smaller subgraphs of a specific type $h$ are inserted. We consider two types of main graphs: $m = \textit{cycle, tree}$. 
The \textit{cycle} graph consists of a single cycle, while the \textit{tree}
graph is constructed by attaching each new node to the graph with a single edge connecting it to a uniformly sampled node already contained in the graph.  

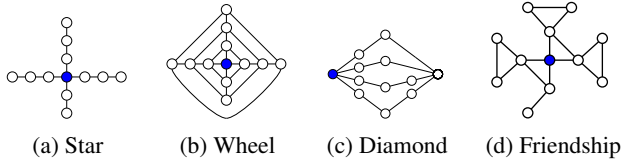
\begin{figure}[t]
\centering
\begin{subfigure}[b]{0.1\textwidth}
    \centering
    \resizebox{\linewidth}{!}{
    \input{figures/graphs/star}
    }
    \caption{Star}
    \label{fig:star}
\end{subfigure}
\hfill
\begin{subfigure}[b]{0.1\textwidth}
    \centering
    \resizebox{\linewidth}{!}{
    \input{figures/graphs/wheel}
    }
    \caption{Wheel}
    \label{fig:wheel}
\end{subfigure}
\hfill
\begin{subfigure}[b]{0.1\textwidth}
    \centering
    \resizebox{\linewidth}{!}{
    \input{figures/graphs/diamond}
    }
    \caption{Diamond}
    \label{fig:diamond}
\end{subfigure}
\hfill
\begin{subfigure}[b]{0.11\textwidth}
    \centering
    \resizebox{\linewidth}{!}{
    \input{figures/graphs/friendship}
    }
    \caption{Friendship}
    \label{fig:friendship}
\end{subfigure}
\caption{Subgraphs used in \textsc{Groletest} dataset with representative central nodes (in blue).}
\label{fig:subgraphs}
\vskip -0.25in
\end{figure}

For the subgraphs $h$, we consider the following five types: $h = \textit{star, wheel, diamond, friendship, random}$. The first four are determined given the total number of nodes and the degree of a representative node, thought of as the central node of the subgraph.
In Figure \ref{fig:subgraphs}, we visualize an example of these subgraphs having $12$ nodes and a central node of degree four (marked in blue). These are all generalizations of graphs that are typically called star, wheel, diamond and friendship in graph theory.
The so-called \textit{random} subgraphs are created by first connecting the central node with new nodes until it has the desired degree. Next, 
the nodes are connected uniformly at random excluding edges to the central node which keeps the desired degree.
To ensure the connectivity of the subgraph, a random edge is added between each connected component and the component containing the central node, again excluding edges to the central node. 
Finally, each subgraph $h$ is inserted to the main graph $m$ by adding a predetermined number of edges between randomly chosen nodes in $h$ and randomly chosen nodes in $m$.

The final \textsc{Groletest} dataset consists of generated graphs $G_h^m$ with a main graph $m$ consisting of $1000$ nodes and $20$ instances of subgraphs $h$, each comprised of $40$ nodes with a central node of degree $4$. Each subgraph is connected to the main graph with four edges. For the \textit{random} subgraphs, $46$ random edges are sampled before all components are connected. In total, each $G_h^m$ contains $1800$ nodes. 

%% file: figures/graphs/star.tex
\begin{tikzpicture}

\def \n {12}
\def \m {4}
\def \radius {0.5}
\def \margin {8}
\node[draw, circle,minimum size=\margin,outer sep=0, inner sep=0, fill=blue] at (0,0) {};
\pgfmathsetmacro\nn{int(\n - 1)};
\foreach \s in {1,...,\nn}
{
  \pgfmathsetmacro\i{floor((\s - 1)/ \m)};
  \node[draw, circle,minimum size=\margin,outer sep=0, inner sep=0] at ({360/\m * (\s - \i*\m - 1)}:{\radius*(\i + 1)} ) {}; 
  \draw[shorten >= \margin/2, shorten <= \margin/2] ({360/\m * (\s - (\i-1)*\m - 1)}:{\radius*((\i-1) + 1)} ) -- ({360/\m * (\s - \i*\m - 1)}:{\radius*(\i + 1)} );
}
\end{tikzpicture}

%% file: figures/graphs/wheel.tex
\begin{tikzpicture}

\def \n {12}
\def \m {4}
\def \radius {0.5}
\def \margin {8}
\node[draw, circle,minimum size=\margin,outer sep=0, inner sep=0, fill=blue] at (0,0) {};
\pgfmathsetmacro\nn{int(ceil((\n - 1)/\m)*\m)};
\pgfmathsetmacro\nout{int((\n - 1))};
\foreach \s in {1,...,\nn}
{
  \pgfmathsetmacro\i{floor((\s - 1)/ \m)};
  \ifthenelse{\s > \nout}{
  \pgfmathsetmacro\xx{360/\m * (\s - (\i-1)*\m - 1)}
  \pgfmathsetmacro\xxx{360/\m * (\s - \i*\m )}
  \draw[shorten >= \margin/2] (\xx:{\radius*(\i + 1)} ) to[in=-90,out=0, distance=10] (\xxx:{\radius*(\i + 1)} );
  \pgfmathsetmacro\xxx{360/\m * (\s - \i*\m -2)}
\draw[shorten >= \margin/2] (\xx:{\radius*(\i + 1)} ) to[in=-90,out=180, distance=10] (\xxx:{\radius*(\i + 1)} );
}{
  \node[draw, circle,minimum size=\margin,outer sep=0, inner sep=0] at ({360/\m * (\s - \i*\m - 1)}:{\radius*(\i + 1)} ) {};
  \draw[shorten >= \margin/2, shorten <= \margin/2] ({360/\m * (\s - (\i-1)*\m - 1)}:{\radius*((\i-1) + 1)} ) -- ({360/\m * (\s - \i*\m - 1)}:{\radius*(\i + 1)} );
  \ifthenelse{ \s = \nout}{}{
  \draw[shorten >= \margin/2, shorten <= \margin/2] ({360/\m * (\s - (\i-1)*\m - 1)}:{\radius*(\i + 1)} ) -- ({360/\m * (\s - \i*\m )}:{\radius*(\i + 1)} );}}

}
\end{tikzpicture}

%% file: figures/graphs/diamond.tex
\begin{tikzpicture}

\def \n {12}
\def \m {4}
\def \radius {0.5}
\def \margin {5} 
\node[draw, circle,minimum size=\margin,outer sep=0, inner sep=0, fill=blue] at (0,0) {};
\pgfmathsetmacro\nn{int(ceil((\n - 1)/\m)*\m)};
\pgfmathsetmacro\nout{int((\n - 2))};
\pgfmathsetmacro\imax{floor((\nn - 1)/ \m)};
\foreach \s in {1,...,\nn}
{
  \pgfmathsetmacro\i{floor((\s - 1)/ \m)};
  \pgfmathsetmacro\ii{min(\i, \imax-\i)};
  \pgfmathsetmacro\iib{min(\i-1, \imax-\i+1)};
  \pgfmathsetmacro\y{\radius*0.5*(1+\ii)*(\s - \i*\m  - \m/2 - 0.5)};
    \pgfmathsetmacro\yb{\radius*0.5*(1+\iib)*(\s - \i*\m  - \m/2 - 0.5)};
\ifthenelse{\s > \nout}{\draw[shorten >= \margin/2, shorten <= -\margin/2] ({\radius*(\i + 1)}, \y )  --({\radius*\i}, \yb );}{
  \node[draw, circle,minimum size=\margin,outer sep=0, inner sep=0] at ({\radius*(\i + 1)}, \y ) {};
   \draw[shorten >= \margin/2, shorten <= \margin/2] ({\radius*(\i + 1)}, \y )  --({\radius*\i}, \yb );}
\node[draw, circle,minimum size=\margin,outer sep=0, inner sep=0] at ({\radius*(\imax+2)},0) {};}
\foreach \s in {1,...,\m}
{
\pgfmathsetmacro\i{floor((\nn + \s - 1)/ \m)};
  \pgfmathsetmacro\ii{min(\i, \imax-\i)};
  \pgfmathsetmacro\iib{min(\i-1, \imax-\i+1)};
    \pgfmathsetmacro\yb{\radius*0.5*(1+\iib)*(\nn + \s - \i*\m  - \m/2 - 0.5)};
       \draw[shorten >= \margin/2, shorten <= \margin/2] ({\radius*(\imax+2)},0)  --({\radius*\i}, \yb );

}
\end{tikzpicture}

%% file: figures/graphs/friendship.tex
\begin{tikzpicture}

\def \n {12}
\def \m {4}
\def \radius {0.3}
\def \margin {3}

\node[draw, circle,minimum size=\margin,outer sep=0, inner sep=0, fill=blue] at (0,0) {};
\pgfmathsetmacro\nn{int(ceil((\n - 1)/\m)*\m)};
\pgfmathsetmacro\nout{int((\n - 1))};
\newcounter{layer}
\setcounter{layer}{0}
\newcounter{node}
\setcounter{node}{0}
\whiledo{{\value{node}<\nn}}{%
\pgfmathsetmacro\nl{\m*pow(2, \thelayer)}
    \stepcounter{layer};
    \foreach \s in {1,...,\nl}
    {
        \stepcounter{node};
        \ifthenelse{\thenode > \nout}{
        }{
            \pgfmathsetmacro\i{\thelayer-1};
            \pgfmathsetmacro\odd{mod(\thenode, 2)};
            \pgfmathsetmacro\angle{360/\nl * (\s - (\i-1)*\nl - 1)};
            \pgfmathsetmacro\newangle{\angle  - (pow(2, \i)-1)*360/(2*\nl)};
            \pgfmathsetmacro\parentangle{\newangle + (2*\odd-1)*360/(2*\nl)}
            \ifthenelse{0 < \i}
            {
                \node[draw, circle,minimum size=\margin,outer sep=0, inner sep=0] at (\newangle:{\radius*(\i + 1)} ) {};}
            {
                \node[draw, circle,minimum size=\margin,outer sep=0, inner sep=0] at (\angle:{\radius*(\i + 1)} ) {};}
            \draw[shorten >= \margin/2, shorten <= \margin/2] (\parentangle:{\radius*((\i-1) + 1)} ) -- (\newangle:{\radius*(\i + 1)} );
            \ifthenelse{0 < \odd}{}
            { 
                \draw[shorten >= \margin/2, shorten <= \margin/2] (\newangle - 360/\nl :{\radius*(\i + 1)} ) -- (\newangle:{\radius*(\i + 1)} );}
        }
    }
}
\end{tikzpicture}

%% file: Sections/experiments.tex
\begin{figure*}[t]
\vskip -0.1in
\newcommand\colscale{0.66}
\begin{subfigure}{\colscale\columnwidth}
    \begin{center}
    \centerline{\resizebox{\linewidth}{!}{
    \includegraphics{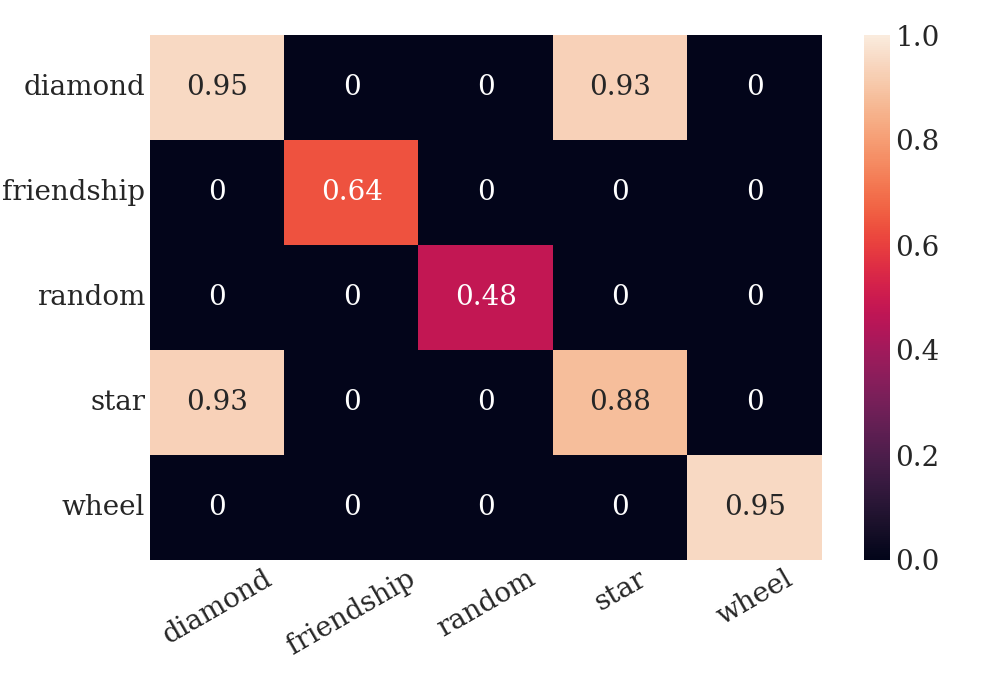}
    }}
    \caption{Manual features}
    \label{fig:synthetic_cm_egonet}
    \end{center}
\end{subfigure}
~
\begin{subfigure}{\colscale\columnwidth}
    \begin{center}
    \centerline{\resizebox{\linewidth}{!}{
    \includegraphics{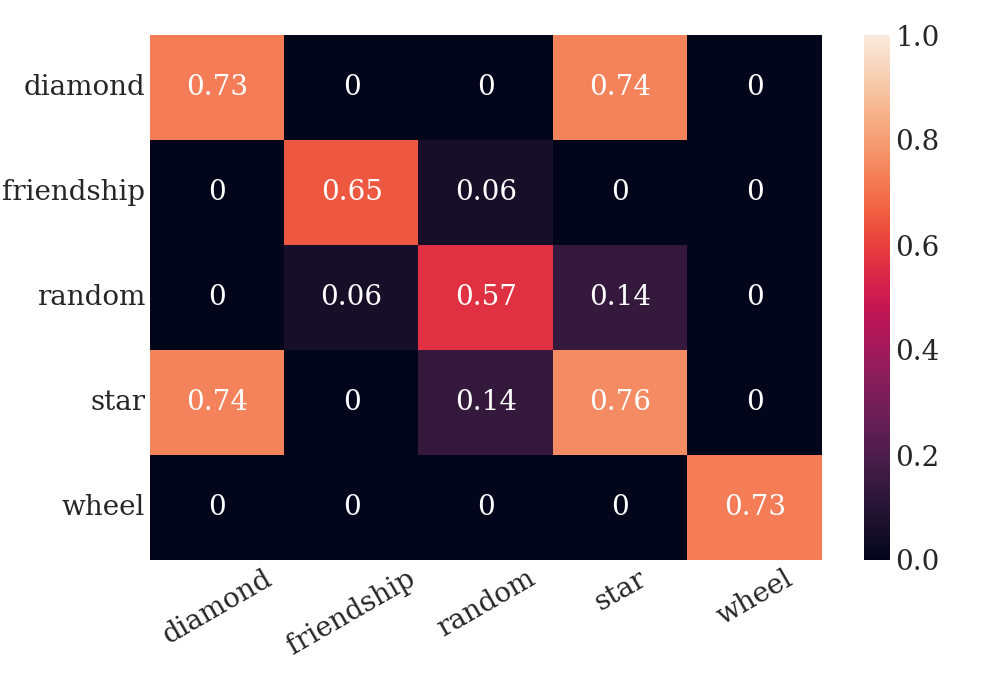}
    }}
    \caption{GCC}
    \label{fig:synthetic_cm_gcc}
    \end{center}
\end{subfigure}
~
\begin{subfigure}{\colscale\columnwidth}
    \begin{center}
    \centerline{\resizebox{\linewidth}{!}{
    \includegraphics{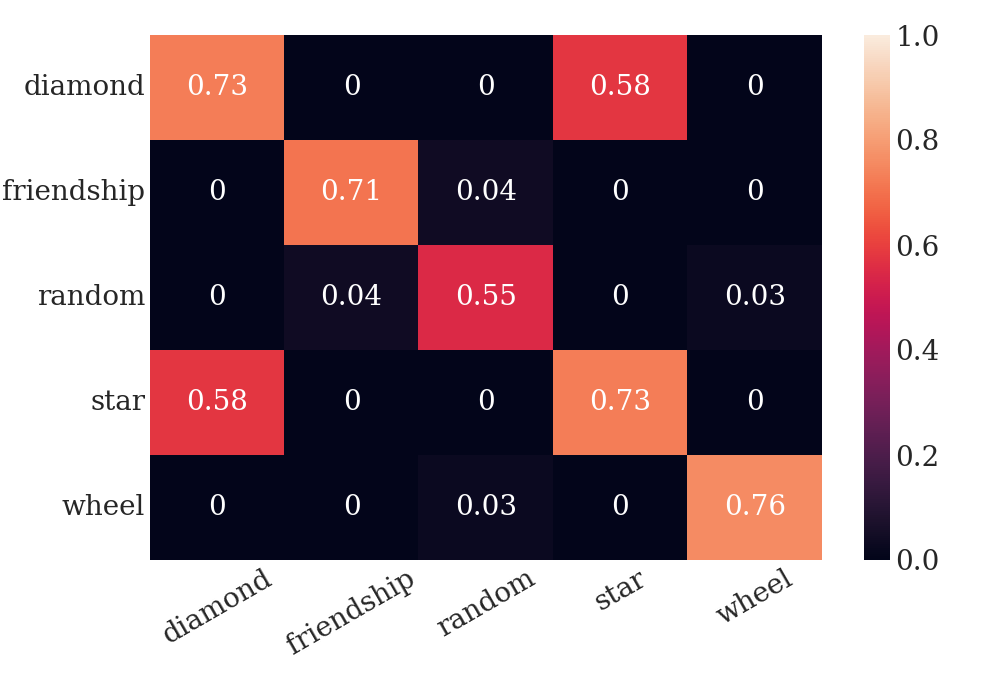}
    }}
    \caption{GraphWave}
    \label{fig:synthetic_cm_graphwave}
    \end{center}
\end{subfigure}
\vskip -0.1in
\caption{Matrices with values representing the average $s_{\wDCA}$ scores obtained for various feature extractors $f$ and $G_1, G_2 \in \{G_h^\textit{cycle} | h = \textit{diamond, friendship, random, star, wheel} \}$. The scores are averaged over 3 randomly generated pairs of graphs.}
\label{fig:synthetic_confusion_matrices}
\end{figure*}

\begin{figure*}[t]
\newcommand\colscale{0.5}
\begin{subfigure}{\colscale\columnwidth}
    \begin{center}
    \Large{\centerline{\resizebox{\linewidth}{!}{
    \input{figures/tex/synthetic_change_main/change_main_graph_egonet_f1_h_marked_weights_mcs2_fix}
    }}}
    \caption{Manual features}
    \label{fig:change_main_egonet}
    \end{center}
\end{subfigure}
~
\begin{subfigure}{\colscale\columnwidth}
    \begin{center}
    \Large{\centerline{\resizebox{\linewidth}{!}{
    \input{figures/tex/synthetic_change_main/change_main_graph_gcc_f1_h_marked_weights_mcs2_fix}
    }}}
    \caption{GCC}
    \label{fig:change_main_gcc}
    \end{center}
\end{subfigure}
~
\begin{subfigure}{\colscale\columnwidth}
    \begin{center}
    \Large{\centerline{\resizebox{\linewidth}{!}{
    \input{figures/tex/synthetic_change_main/change_main_graph_graphwave_f1_h_marked_weights_mcs2_fix}
    }}}
    \caption{GraphWave}
    \label{fig:change_main_graphwave}
    \end{center}
\end{subfigure}
~
\begin{subfigure}{\colscale\columnwidth}
    \begin{center}
    \Large{\centerline{\resizebox{\linewidth}{!}{
    \input{figures/tex/synthetic_change_main/change_main_graph_gstats_rel_hmean_mcs2_fix}
    }}}
    \caption{Global statistics}
    \label{fig:change_main_gstats}
    \end{center}
\end{subfigure}
\vskip -0.1in
\caption{Average $s_{\wDCA}$ score (\subref{fig:change_main_egonet}, \subref{fig:change_main_gcc}, \subref{fig:change_main_graphwave}) obtained between two input graphs with fixed type of subgraphs $h$ having same main graph, i.e., $G_1 = G_h^{\textit{cycle}}, G_2 = G_h^{\textit{cycle}}$ (solid lines), and having different main graphs, i.e., $G_1 = G_h^{\textit{cycle}}, G_2 = G_h^{\textit{tree}}$ (dashed lines). The relative global statistics of input graphs $s_{\gstats}(G_1, G_2)$ are shown in \subref{fig:change_main_gstats}. The scores are averaged over 3 randomly generated pairs of graphs.}
\label{fig:synthetic_varying_main_graph}
\end{figure*} 

\section{Experiments}
We validate GraphDCA through experiments on both \textsc{Groletest} and three well-known real-world graphs described below.
Specifically, using feature extractors $f$ 
outlined in Section \ref{sec:gdca:feature_extractors}, we demonstrate that \textit{i)} GraphDCA can identify \textsc{Groletest} graphs with both similar and dissimilar distributions of local structures, and that \textit{ii)} $s_{\wDCA}$ similarity score is robust when comparing \textsc{Groletest} with different main graphs.
Furthermore, we show that \textit{iii)} $s_{\wDCA}$ decreases consistently with gradually perturbed graph structure both for real graphs, which we perturb globally, and \textsc{Groletest} graphs where only known local structures are changed. 
Finally, we apply GraphDCA with the considered feature extractors to \textit{iv)} evaluate two state-of-the-art generative models, NetGAN \cite{bojchevski_netgan_2018} and CELL \cite{rendsburg_netgan_2020}, trained on both \textsc{Groletest} and real graphs. The resulting $s_{\wDCA}$ scores suggest that these models struggle to capture local structure of the training graph, urging for future improvements. 

\textbf{Similarity scores.}
Since \textsc{Groletest} graphs provide ground truth information regarding node roles, we calculate $s_{\wDCA}$ with respect to the following node weights: we set $w_i = 0$ for all $i$ except for the 20 central nodes (see Figure \ref{fig:subgraphs}) taken as role representatives for each subgraph which we give weight $w_i=1$. We use uniform weighting on real-world and generated graphs where roles are not specified. Results on \textsc{Groletest} obtained with uniform weights are discussed in Appendix~\ref{sec:app:results}. 

In addition to $s_{\wDCA}$, we also compute the same global graphs statistics as in \cite{bojchevski_netgan_2018, rendsburg_netgan_2020}, i.e.\ maximum degree and power law exponent, the global clustering coefficient, assortativity, characteristic path length and motif counts, triangles and squares (see Table \ref{tab:graph_stats} in Appendix \ref{sec:app:gstats} for definitions).
To facilitate comparison with $s_{\wDCA}$, we compute the harmonic mean $H$ of the relative difference of the individual statistics, i.e.,\
\begin{displaymath}
    s_{\gstats}(G_1, G_2) = H \left( \left\{ 1 - \frac{|\zeta_{j,1} - \zeta_{j,2} |}{|\zeta_{j,1}| + |\zeta_{j, 2}|} \right\}_j \right),
\end{displaymath}
where $\zeta_{j,1}$ and $\zeta_{j,2}$ are the values of the same graph statistic calculated on $G_1$ and $G_2$, respectively. Moreover, on \textsc{Groletest}, we report average $s_{\wDCA}$ and $s_{\gstats}$ obtained on $3$ randomly generated pairs of graphs.

\textbf{Datasets.}
We use three well-known real-world graphs, \textsc{Cora-ML}, \textsc{Citeseer}, \textsc{PolBlogs} considered in~\cite{bojchevski_netgan_2018}, in addition to \textsc{Groletest} data. 
Each of the real-world graphs was preprocessed according to the common practice in literature: edge weights were removed, directed edges were turned into undirected, multiple edges and loops were eliminated. Finally, only the largest connected component (LCC) was used for each graph.

\subsection{Local Structural Similarity} \label{sec:struct_sim}


\textbf{Setup.} First, to verify that GraphDCA can identify similar and dissimilar local structures, we calculate $s_{\text{wDCA}}(f(G_h^{\textit{cycle}}), f(G_t^{\textit{cycle}}))$ for each $h,t \in \{ \textit{diamond, friendship, random, star, wheel} \} $ and for each of the three feature extractors $f$ (Section \ref{sec:gdca:feature_extractors}).
Second, to ensure that similarities are robust even when varying the global structure but preserving the local ones, we calculate similarity scores between graphs with the same subgraph types but different main graphs, $s_{\text{wDCA}}(f(G_h^{\textit{cycle}}), f(G_h^{\textit{tree}}))$.



\begin{figure}
\newcommand\colscale{0.49}
\centering
\begin{subfigure}{\colscale\columnwidth}
    \begin{center}
    \Large{\centerline{\resizebox{\linewidth}{!}{
    \input{figures/tex/real_rewire/frac_rewired_counts_F1_qual_rel_score_configuration_cora-ml_mcs_2}
    }}}
    \caption{\textsc{Cora-ML}}
    \label{fig:coraml configuration frac}
    \end{center}
\end{subfigure}
\hfill
\begin{subfigure}{\colscale\columnwidth}
    \begin{center}
    \Large{\centerline{\resizebox{\linewidth}{!}{
    \input{figures/tex/real_rewire/frac_rewired_counts_F1_qual_rel_score_configuration_citeseer_mcs_2}
    }}}
    \caption{\textsc{Citeseer}}
    \label{fig:citeseer configuration frac}
    \end{center}
\end{subfigure}
\\
\begin{subfigure}{\colscale\columnwidth}
    \begin{center}
    \Large{\centerline{\resizebox{\linewidth}{!}{
    \input{figures/tex/real_rewire/frac_rewired_counts_F1_qual_rel_score_configuration_polblogs_mcs_2}
    }}}
    \caption{\textsc{PolBlogs}}
    \label{fig:polblogs configuration frac}
    \end{center}
\end{subfigure}
\hfill
\begin{minipage}[b][0.3\columnwidth][t]{\colscale\columnwidth}
\centering
\caption{Rewiring of real-world graphs using the configuration rewiring model. The y-axis shows $s_{\gstats}$ (in black) 
and $s_{\wDCA}$ scores for different feature extractors (in color), averaged over $5$ different rewirings. \label{fig:configuration frac}}
\end{minipage}
\vskip -0.3in
\end{figure}
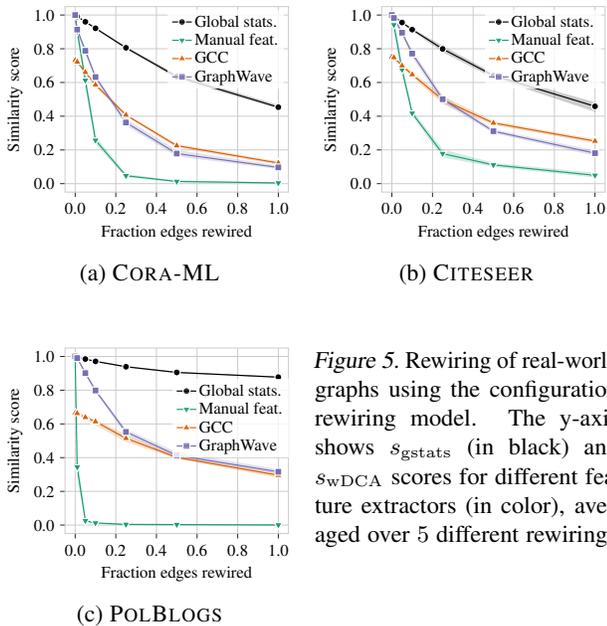

\begin{figure*}[t]
\newcommand\colscale{0.5}
\begin{subfigure}{\colscale\columnwidth}
    \begin{center}
    \Large{\centerline{\resizebox{\linewidth}{!}{
    \input{figures/tex/synthetic_rewire/frac_rewired_diamond_marked_F1_qual_rel_score_mcs_2}
    }}}
    \caption{Diamond}
    \label{fig:rewire diamond}
    \end{center}
\end{subfigure}
~
\begin{subfigure}{\colscale\columnwidth}
    \begin{center}
    \Large{\centerline{\resizebox{\linewidth}{!}{
    \input{figures/tex/synthetic_rewire/frac_rewired_friendship_marked_F1_qual_rel_score_mcs_2}
    }}}
    \caption{Friendship}
    \label{fig:rewire friendship}
    \end{center}
\end{subfigure}
~
\begin{subfigure}{\colscale\columnwidth}
    \begin{center}
    \Large{\centerline{\resizebox{\linewidth}{!}{
    \input{figures/tex/synthetic_rewire/frac_rewired_star_marked_F1_qual_rel_score_mcs_2}
    }}}
    \caption{Star}
    \label{fig:rewire star}
    \end{center}
\end{subfigure}
~
\begin{subfigure}{\colscale\columnwidth}
    \begin{center}
    \Large{\centerline{\resizebox{\linewidth}{!}{
    \input{figures/tex/synthetic_rewire/frac_rewired_wheel_marked_F1_qual_rel_score_mcs_2}
    }}}
    \caption{Wheel}
    \label{fig:rewire wheel}
    \end{center}
\end{subfigure}
\vskip -0.1in
\caption{Rewiring of \textsc{Groletest} subgraphs via the addition and removal of edges. The x-axis specifies the fraction of changed edges. The y-axis shows $s_{\gstats}$ scores (in black) and $s_{\wDCA}$  scores for different feature extractors (in color), averaged over $5$ different rewirings.}
\label{fig:synthetic rewire}
\vskip -0.2in
\end{figure*}
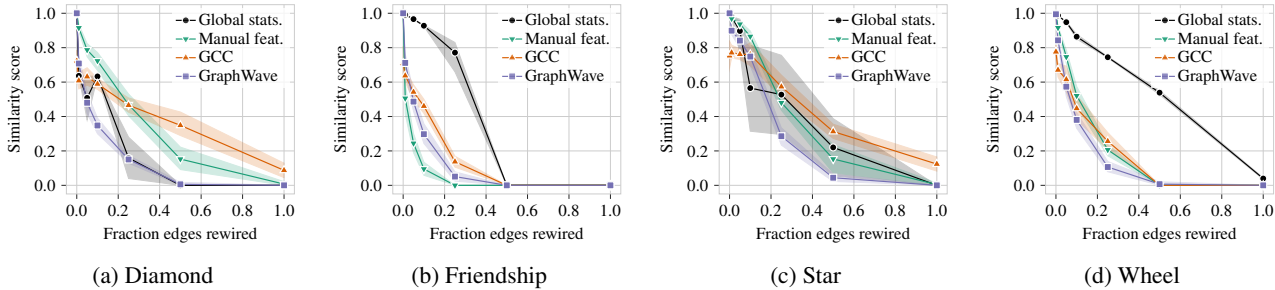



\textbf{Results and discussion.}
We report the average $s_{\wDCA}$ scores in the form of matrices in Figure~\ref{fig:synthetic_confusion_matrices}. 
Firstly, we observe that the scores obtained with all $f$ are prevailing on the diagonal which is the desired behavior. 
Secondly, all models detect similarity between $G_{\textit{diamond}}^{\textit{cycle}}$ and $G_{\textit{star}}^{\textit{cycle}}$ and a lower similarity score between $G_{\textit{random}}^{\textit{cycle}}$ graphs. 
The similarity between \textit{diamond} and \textit{star} subgraphs is expected as the only difference between these subgraphs are $3$ additional edges in the \textit{diamond} which are all far away from the central node. These additional edges are not included in the manual features as these only consider nodes within shortest path distance of $4$. Similarly, it is highly unlikely for GCC to include these edges due to the restart probability of the random walks.
Consequently, GraphWave is the only feature extractor yielding noticeably lower scores between \textit{diamond} and \textit{star} than between \textit{diamond} and \textit{diamond}.
The lower similarity score between \textit{random} subgraphs is also expected since these subgraphs, unlike the other subgraph types, are only similar and not identical. Due to their random structure, $G_{\textit{random}}^{\textit{cycle}}$ are in some cases slightly similar to other subgraphs (Figure~\ref{fig:synthetic_cm_gcc}). 

In Figure~\ref{fig:synthetic_varying_main_graph}, we show $s_{\wDCA}$ for graph pairs $(G_h^{\textit{cycle}}, G_h^{\textit{tree}})$ having similar local $h$ but different global structure $m$ (dashed lines), and graph pairs  $(G_h^{\textit{cycle}}, G_h^{\textit{cycle}})$ with both similar local and global structure (solid lines).
We observe that global statistics similarity $s_{\gstats}$ (Figure~\ref{fig:change_main_gstats}) varies significantly across different input combinations. 
For example, $s_{\gstats}$ changes drastically for \textit{random} subgraphs, exhibits large variance for \textit{friendship} and only changes slightly for \textit{star} subgraphs. 
On the other hand, $s_{\wDCA}$ scores are largely robust to changes in the main graph, especially for representations obtained with GCC (Figure~\ref{fig:change_main_gcc}). 
The $s_{\wDCA}$ scores do change more when using manual features (Figure~\ref{fig:change_main_egonet}) possibly since these are more sensitive to minor changes in the graph given that single edges can significantly change the $\rho$-egonet structure.



\comment{
\begin{figure*}[t]
\vskip -0.1in
\newcommand\colscale{0.5}
\begin{subfigure}{\colscale\columnwidth}
    \begin{center}
    \centerline{\resizebox{\linewidth}{!}{
    \input{figures/tex/real_rewire/frac_rewired_counts_F1_qual_rel_score_configuration_cora-ml_mcs_2}
    }}
    \caption{\textsc{Cora-ML}}
    \label{fig:coraml configuration frac}
    \end{center}
\end{subfigure}
\hfill
\begin{subfigure}{\colscale\columnwidth}
    \begin{center}
    \centerline{\resizebox{\linewidth}{!}{
    \input{figures/tex/real_rewire/frac_rewired_counts_F1_qual_rel_score_configuration_citeseer_mcs_2}
    }}
    \caption{\textsc{Citeseer}}
    \label{fig:citeseer configuration frac}
    \end{center}
\end{subfigure}
\hfill
\begin{subfigure}{\colscale\columnwidth}
    \begin{center}
    \centerline{\resizebox{\linewidth}{!}{
    \input{figures/tex/real_rewire/frac_rewired_counts_F1_qual_rel_score_configuration_polblogs_mcs_2}
    }}
    \caption{\textsc{PolBlogs}}
    \label{fig:polblogs configuration frac}
    \end{center}
\end{subfigure}
\vskip -0.1in
\caption{Rewiring of real graphs using the configuration rewiring model. The y-axes shows $s_{\gstats}$ scores in black and $s_{\wDCA}$  scores for three different feature extractors in color, averaged over five different rewirings.}
\label{fig:configuration frac}
\end{figure*}
}

\subsection{Gradual Structure Perturbation}
\label{sec:rewiring}
To show that $s_{\wDCA}$ can capture gradually decreasing similarity, we preform structure perturbation experiments on the three real-world graphs and the \textsc{Groletest} graphs.

{\bf Setup.} For the real-world graphs, we use the configuration rewiring model where edges are rewired such that the degree distributions of the graphs remain unchanged 
but not local structures.
For \textsc{Groletest}, we only perturb the subgraph structures. Instead of configuration rewiring, we first find a random minimum spanning tree of the subgraph, randomly remove edges from the subgraph not present in the tree, and add random previously non-existing edges in the subgraph.
In both cases, we measure the perturbation as the fraction of rewired, added or removed edges to the number of edges in the original graph. We report the similarity scores averaged over $5$ different rewirings per fraction.
See Appendix~\ref{sec:app:rewire} for additional details on the perturbation procedure.


\textbf{Results and discussion.} In Figure \ref{fig:configuration frac}, we visualize the scores obtained on real-world graphs. We observe that $s_{\wDCA}$ with all $f$ and $s_{\gstats}$ drop consistently as the edges are rewired.
\comment{
    \begin{wrapfigure}{r}{0.5\columnwidth}\centering
    \newcommand\colscale{0.5}
    \begin{subfigure}{\colscale\columnwidth}
        \begin{center}
        \centerline{\resizebox{\linewidth}{!}{
        \input{figures/tex/real_rewire/frac_rewired_counts_F1_qual_rel_score_configuration_cora-ml_mcs_2}
        }}
        \caption{\textsc{Cora-ML}}
        \label{fig:coraml configuration frac}
        \end{center}
    \end{subfigure}
    \begin{subfigure}{\colscale\columnwidth}
        \begin{center}
        \centerline{\resizebox{\linewidth}{!}{
        \input{figures/tex/real_rewire/frac_rewired_counts_F1_qual_rel_score_configuration_citeseer_mcs_2}
        }}
        \caption{\textsc{Citeseer}}
        \label{fig:citeseer configuration frac}
        \end{center}
    \end{subfigure}
    \begin{subfigure}{\colscale\columnwidth}
        \begin{center}
        \centerline{\resizebox{\linewidth}{!}{
        \input{figures/tex/real_rewire/frac_rewired_counts_F1_qual_rel_score_configuration_polblogs_mcs_2}
        }}
        \caption{\textsc{PolBlogs}}
        \label{fig:polblogs configuration frac}
        \end{center}
    \end{subfigure}
    \caption{Rewiring of real-world graphs using the configuration rewiring model. The $s_{\gstats}$ (black) 
    and $s_{\wDCA}$ scores for different $f$ (color) are averaged over $5$ different rewirings.
    }
    \label{fig:configuration frac}
    \end{wrapfigure}
}
The changes in global statistics are mainly attributed to changes in the global clustering coefficient and number of triangle/squares which may differ by up to a factor 7 between the original and rewired graphs.
Since configuration rewiring is used, max degree and power law exponent are unaffected, while we observe that the assortativity and characteristic path length change at most by a factor of 2.
For \textsc{PolBlogs}, $s_{\gstats}$ remains high even after all edges have been rewired since this graph is much denser than the two citation graphs, \textsc{Cora-ML} and \textsc{Citeseer}, with average degree $27$ compared to $5.6$ and $3.5$, respectively.
Consequently, the global clustering coefficient and motif counts do not drop as much during rewiring of \textsc{PolBlogs} since new triangles and squares are more likely to form during the rewiring.

While the behavior of the GraphWave and GCC representations is consistent across all datasets, $s_{\wDCA}$ obtained with manual features drops rapidly for \textsc{PolBlogs}.
We conjecture the cause to be the presence of two distinct communities, corresponding to the two major US political parties \cite{divided_blog}.
Rewiring of single edges may result in significant changes in the $\rho$-egonet structure as community bridges are formed or broken.
This in turn induces larger changes in the distribution of the manual features compared to the GraphWave and GCC which are not as sensitive to single edge perturbations.
A final observation is that $s_{\wDCA}$ obtained with GCC never reaches $1.0$ even when two isomorphic graphs are compared which is expected as GCC feature extraction is not deterministic.


The similarity scores obtained on \textsc{Groletest} are shown in Figure \ref{fig:synthetic rewire}. We observe a largely consistent gradual drop of $s_{\wDCA}$ obtained using different feature extractors $f$, where GraphWave representations exhibit particularly low variance.
On the contrary, the behavior of $s_{\gstats}$ differs significantly between the subgraph types as edge perturbation affects the global statistics differently depending on the subgraph structure.
For the \textit{diamond} and \textit{star} subgraphs, the characteristic path length changes consistently with the perturbation, while the power law exponential drops from $20.7$ to $2.4$ with just 1\% of perturbed edges.
This sudden change is not observed for \textit{friendship} and \textit{wheel} which instead exhibit larger changes in assortativity, global clustering coefficient and number of triangles. 
This highlights the stability of $s_{\wDCA}$ compared to the sensitivity of the choice of global statistics.

\subsection{Evaluation of Graph Generative Models}
We demonstrate how GraphDCA can provide valuable insights for evaluation of graph generative models which is an important open problem. We consider the recently proposed NetGAN and CELL generative models trained on both \textsc{Groletest} and real-world datasets. Comparing these models is particularly interesting as CELL is essentially a NetGAN model stripped of the GAN component, while arguably retaining its performance.

\textbf{Setup.} 
Prior to the training of the models, the edges of each preprocessed graph were split into 85\% train, 10\% validation and 5\% test sets, following the procedure from~\cite{bojchevski_netgan_2018}. NetGAN was separately trained with 50\% edge overlap (EO) and link prediction validation (Val) stopping criteria. CELL model was trained using the same splits with 50\% EO criterion. Hyperparameters and experimental details on training and generation can be found in Appendix~\ref{sec:app:generative_hyper}.
We run GraphDCA on three randomly generated graphs $G_2$ for each considered training graph $G_1$. 

\begin{table*}[t]
\centering
\vspace{-1ex} 
\caption{Global graph statistics of the real-world training graphs $G_1$ and $s_{\wDCA}$ scores obtained on representations extracted from the considered feature extractors $f$ of $G_1$ and graphs $G_2$ generated by NetGAN and CELL (50\% EO stopping criterion). The scores are averaged over 3 independently generated graphs with the same training split.
}
\vspace{2ex}
\begin{adjustbox}{width=\linewidth,center}
\input{tables/generative_models_with_ground_truth}
\end{adjustbox}
\label{table:real_gen_models:dca}
\end{table*}

\begin{table*}[t]
\centering
\caption{Global graph statistics of \textsc{Groletest} training graphs $G_1$ and $s_{\wDCA}$ scores obtained on representations extracted from the considered feature extractors $f$ of $G_1$ and graphs $G_2$ generated by NetGAN and CELL (50\% EO stopping criterion). The scores are averaged over 3 independently generated graphs with the same training split.}
\vspace{2ex}
\begin{adjustbox}{width=\linewidth,center}
\input{tables/mainpaper_synthetic_graphwave_gm}
\end{adjustbox}
\label{table:synthetic_gen_models:dca}
\end{table*}

\textbf{Results and discussion.} 
The $s_{\wDCA}$ scores and global statistics obtained on real-world graphs are shown in Table~\ref{table:real_gen_models:dca}. Both NetGAN and CELL capture most of the global properties well which is expected
since global statistics likely served as a methodological validation during the model development. However, they struggle to reproduce triangles and squares -- local statistics that is often overlooked despite consistently being non reproducible~\cite{rendsburg_netgan_2020}. With such large differences in triangle and square counts (often over 50\%), evaluation of the preservation of local structures in the generated graphs is challenging.

On the other hand, $s_{\wDCA}$ with different feature extractors exhibit similar ratios as in the rewiring experiments in Section~\ref{sec:rewiring}. Interestingly, the scores obtained for all three types of graphs are close to the values observed after the rewiring of $\sim20\%$ edges (see Figure~\ref{fig:configuration frac}). Since approximately 50\% of the edges are different in the generated and the training graphs, this indicates a certain degree of generalization achieved by the models and confirms that model performance is not defined by high edge overlap criterion.

When comparing the performance of two models via  $s_{\wDCA}$, the ratio between NetGAN and CELL scores is consistent across different $f$. For \textsc{Cora-ML} and \textsc{Citeseer}, this ratio is $\sim1.2$, indicating that NetGAN outperforms CELL to a small extent. The similarity between the scores for the two datasets can be due to the semantic similarity of the graphs which are both citation networks. In contrast, CELL ourperforms NetGAN on \textsc{PolBlogs} (with score ratio $\sim0.75$), possibly due to a larger density of the graph which might be more sensitive to the number of random walks sampled in NetGAN. Overall, both models produce medium $s_{\wDCA}$ scores on the real-world graphs indicating poor reproducibility of the local patterns. The bias towards reproduction of global statistics is particularly visible in $s_{\wDCA}$ obtained with the manual features which are more sensitive to changes in the local structure.

In Table~\ref{table:synthetic_gen_models:dca} we report the results obtained on \textsc{Groletest} for $h = \textit{diamond, friendship, wheel}$ (for $h = \textit{random, star}$ see Table~\ref{table:app_synthetic_gen_models_random_star:dca} in Appendix~\ref{sec:app:gen_models}). We observe that the models struggle to produce graphs that would well reflect all the global statistics. This is possibly due to a combination of low density of the graphs (average degree is not higher than 2) and high characteristic path length (ca. 40-80). Short random walks utilized by NetGAN and low-rank logit space in CELL have lower capability of capturing the sparsity of the graph as models have to store progressively more information about the local structure to avoid shortcuts. Moreover, we observe that global statistics exhibit larger variations across different generated graphs of the same type than in real-world graphs. In contrast, $s_{\wDCA}$ values still have low variation compared to the unstable global statistics, and higher confidence in the model evaluation.
Across all $5$ subgraph types, CELL outperforms NetGAN in terms of $s_{\wDCA}$ scores, indicating its higher ability in capturing the local structure of low-density graphs. This effect can also be illustrated by the fact that CELL produces graphs with 0 triangle count when trained on graphs without triangles (\textit{diamond, star, random}), while NetGAN pollutes local properties of the nodes. The CELL $s_{\wDCA}$ scores obtained with manual features are accordingly relatively high for these graphs. The overall low $s_{\wDCA}$ scores show that neither NetGAN nor CELL can generate graphs that resemble well the training graphs, although CELL is more efficient in replicating the egocentric networks in \textsc{Groletest} graphs.

Finally, we compare graphs generated by NetGAN models trained with EO and Val stopping criteria. The global statistics and $s_{\wDCA}$ scores can be found in Table~\ref{table:app_real_netgan_val_vs_eo:dca} in Appendix~\ref{sec:app:gen_models}. The results support previous conclusions about higher efficiency of the edge overlap criterion in generating new graphs. It is worth noting that in cases when all global statistics are either both similar or both different from the training values (e.g., on \textsc{Cora-ML}), $s_{\wDCA}$ scores provide a more clear comparison between two models, in this case in favor of EO criterion.

%% file: figures/tex/synthetic_change_main/change_main_graph_egonet_f1_h_marked_weights_mcs2_fix.tex
\begin{tikzpicture}

\definecolor{color0}{rgb}{0.105882352941176,0.619607843137255,0.466666666666667}
\definecolor{color1}{rgb}{0.0705882352941176,0.431372549019608,0.325490196078431}
\begin{axis}[
axis line style={white!80!black},
legend cell align={left},
legend style={
  fill opacity=0.8,
  draw opacity=1,
  text opacity=1,
  at={(1.0,0.05)},
  anchor=south east,
  draw=none
},
tick align=outside,
tick pos=left,
x grid style={white!80!black},
xmajorgrids,
xmin=-0.2, xmax=4.2,
xtick style={color=white!15!black},
xtick={0,1,2,3,4},
xtick={0,1,2,3,4},
xtick={0,1,2,3,4},
xticklabel style={rotate=30.0},
xticklabels={diamond,friendship,random,star,wheel},
xticklabels={diamond,friendship,random,star,wheel},
xticklabels={diamond,friendship,random,star,wheel},
y grid style={white!80!black},
ylabel={Similarity score},
ymajorgrids,
ymin=0, ymax=1,
ytick style={color=white!15!black},
ytick={0,0.2,0.4,0.6,0.8,1},
yticklabels={0.0,0.2,0.4,0.6,0.8,1.0}
]
\path [draw=color0, line width=1pt]
(axis cs:0,0.913)
--(axis cs:0,0.979);

\path [draw=color0, line width=1pt]
(axis cs:1,0.532)
--(axis cs:1,0.734);

\path [draw=color0, line width=1pt]
(axis cs:2,0.366)
--(axis cs:2,0.562);

\path [draw=color0, line width=1pt]
(axis cs:3,0.858)
--(axis cs:3,0.902);

\path [draw=color0, line width=1pt]
(axis cs:4,0.941)
--(axis cs:4,0.963);

\path [draw=color1, line width=1pt]
(axis cs:0,0.41)
--(axis cs:0,0.594);

\path [draw=color1, line width=1pt]
(axis cs:1,0.423)
--(axis cs:1,0.507);

\path [draw=color1, line width=1pt]
(axis cs:2,0.36)
--(axis cs:2,0.426);

\path [draw=color1, line width=1pt]
(axis cs:3,0.53)
--(axis cs:3,0.6);

\path [draw=color1, line width=1pt]
(axis cs:4,0.583)
--(axis cs:4,0.753);

\addplot [line width=1pt, color0, mark=triangle*, mark size=3, mark options={solid,rotate=180}]
table {%
0 0.946
1 0.633
2 0.464
3 0.88
4 0.952
};
\addlegendentry{cycle-cycle}
\addplot [line width=1pt, color1, dashed, mark=diamond*, mark size=3, mark options={solid}]
table {%
0 0.502
1 0.465
2 0.393
3 0.565
4 0.668
};
\addlegendentry{cycle-tree}
\end{axis}
\end{tikzpicture}

%% file: figures/tex/synthetic_change_main/change_main_graph_gcc_f1_h_marked_weights_mcs2_fix.tex
\begin{tikzpicture}

\definecolor{color0}{rgb}{0.850980392156863,0.372549019607843,0.00784313725490196}
\definecolor{color1}{rgb}{0.592156862745098,0.258823529411765,0.00392156862745098}

\begin{axis}[
axis line style={white!80!black},
legend cell align={left},
legend style={
  fill opacity=0.8,
  draw opacity=1,
  text opacity=1,
  at={(1.0,0.05)},
  anchor=south east,
  draw=none
},
tick align=outside,
tick pos=left,
x grid style={white!80!black},
xmajorgrids,
xmin=-0.2, xmax=4.2,
xtick style={color=white!15!black},
xtick={0,1,2,3,4},
xtick={0,1,2,3,4},
xtick={0,1,2,3,4},
xticklabel style={rotate=30.0},
xticklabels={diamond,friendship,random,star,wheel},
xticklabels={diamond,friendship,random,star,wheel},
xticklabels={diamond,friendship,random,star,wheel},
y grid style={white!80!black},
ylabel={Similarity score},
ymajorgrids,
ymin=0, ymax=1,
ytick style={color=white!15!black},
ytick={0,0.2,0.4,0.6,0.8,1},
yticklabels={0.0,0.2,0.4,0.6,0.8,1.0}
]
\path [draw=color0, line width=1pt]
(axis cs:0,0.676)
--(axis cs:0,0.77);

\path [draw=color0, line width=1pt]
(axis cs:1,0.554)
--(axis cs:1,0.718);

\path [draw=color0, line width=1pt]
(axis cs:2,0.423)
--(axis cs:2,0.699);

\path [draw=color0, line width=1pt]
(axis cs:3,0.607)
--(axis cs:3,0.889);

\path [draw=color0, line width=1pt]
(axis cs:4,0.651)
--(axis cs:4,0.801);

\path [draw=color1, line width=1pt]
(axis cs:0,0.612)
--(axis cs:0,0.702);

\path [draw=color1, line width=1pt]
(axis cs:1,0.481)
--(axis cs:1,0.611);

\path [draw=color1, line width=1pt]
(axis cs:2,0.434)
--(axis cs:2,0.55);

\path [draw=color1, line width=1pt]
(axis cs:3,0.4)
--(axis cs:3,0.804);

\path [draw=color1, line width=1pt]
(axis cs:4,0.514)
--(axis cs:4,0.686);

\addplot [line width=1pt, color0, mark=triangle*, mark size=3, mark options={solid}]
table {%
0 0.723
1 0.636
2 0.561
3 0.748
4 0.726
};
\addlegendentry{cycle-cycle}
\addplot [line width=1pt, color1, dashed, mark=diamond*, mark size=3, mark options={solid}]
table {%
0 0.657
1 0.546
2 0.492
3 0.602
4 0.6
};
\addlegendentry{cycle-tree}
\end{axis}

\end{tikzpicture}

%% file: figures/tex/synthetic_change_main/change_main_graph_graphwave_f1_h_marked_weights_mcs2_fix.tex
\begin{tikzpicture}

\definecolor{color0}{rgb}{0.458823529411765,0.43921568627451,0.701960784313725}
\definecolor{color1}{rgb}{0.317647058823529,0.305882352941176,0.490196078431373}

\begin{axis}[
axis line style={white!80!black},
legend cell align={left},
legend style={
  fill opacity=0.8,
  draw opacity=1,
  text opacity=1,
  at={(1.0,0.05)},
  anchor=south east,
  draw=none
},
tick align=outside,
tick pos=left,
x grid style={white!80!black},
xmajorgrids,
xmin=-0.2, xmax=4.2,
xtick style={color=white!15!black},
xtick={0,1,2,3,4},
xtick={0,1,2,3,4},
xtick={0,1,2,3,4},
xticklabel style={rotate=30.0},
xticklabels={diamond,friendship,random,star,wheel},
xticklabels={diamond,friendship,random,star,wheel},
xticklabels={diamond,friendship,random,star,wheel},
y grid style={white!80!black},
ylabel={Similarity score},
ymajorgrids,
ymin=0, ymax=1,
ytick style={color=white!15!black},
ytick={0,0.2,0.4,0.6,0.8,1},
yticklabels={0.0,0.2,0.4,0.6,0.8,1.0}
]
\path [draw=color0, line width=1pt]
(axis cs:0,0.67)
--(axis cs:0,0.782);

\path [draw=color0, line width=1pt]
(axis cs:1,0.536)
--(axis cs:1,0.854);

\path [draw=color0, line width=1pt]
(axis cs:2,0.494)
--(axis cs:2,0.6);

\path [draw=color0, line width=1pt]
(axis cs:3,0.613)
--(axis cs:3,0.833);

\path [draw=color0, line width=1pt]
(axis cs:4,0.707)
--(axis cs:4,0.801);

\path [draw=color1, line width=1pt]
(axis cs:0,0.351)
--(axis cs:0,0.509);

\path [draw=color1, line width=1pt]
(axis cs:1,0.503)
--(axis cs:1,0.577);

\path [draw=color1, line width=1pt]
(axis cs:2,0.472)
--(axis cs:2,0.502);

\path [draw=color1, line width=1pt]
(axis cs:3,0.458)
--(axis cs:3,0.554);

\path [draw=color1, line width=1pt]
(axis cs:4,0.347)
--(axis cs:4,0.585);

\addplot [line width=1pt, color0, mark=square*, mark size=3, mark options={solid}]
table {%
0 0.726
1 0.695
2 0.547
3 0.723
4 0.754
};
\addlegendentry{cycle-cycle}
\addplot [line width=1pt, color1, dashed, mark=diamond*, mark size=3, mark options={solid}]
table {%
0 0.43
1 0.54
2 0.487
3 0.506
4 0.466
};
\addlegendentry{cycle-tree}
\end{axis}

\end{tikzpicture}

%% file: figures/tex/synthetic_change_main/change_main_graph_gstats_rel_hmean_mcs2_fix.tex
\begin{tikzpicture}

\begin{axis}[
axis line style={white!80!black},
legend cell align={left},
legend style={
  fill opacity=0.8,
  draw opacity=1,
  text opacity=1,
  at={(1.0,0.05)},
  anchor=south east,
  draw=none
},
tick align=outside,
tick pos=left,
x grid style={white!80!black},
xmajorgrids,
xmin=-0.2, xmax=4.2,
xtick style={color=white!15!black},
xtick={0,1,2,3,4},
xtick={0,1,2,3,4},
xtick={0,1,2,3,4},
xticklabel style={rotate=30.0},
xticklabels={diamond,friendship,random,star,wheel},
xticklabels={diamond,friendship,random,star,wheel},
xticklabels={diamond,friendship,random,star,wheel},
y grid style={white!80!black},
ylabel={Similarity score},
ymajorgrids,
ymin=0, ymax=1,
ytick style={color=white!15!black},
ytick={0,0.2,0.4,0.6,0.8,1},
yticklabels={0.0,0.2,0.4,0.6,0.8,1.0}
]
\path [draw=black, line width=1pt]
(axis cs:0,0.962)
--(axis cs:0,0.992);

\path [draw=black, line width=1pt]
(axis cs:1,0.985)
--(axis cs:1,1.001);

\path [draw=black, line width=1pt]
(axis cs:2,0.905)
--(axis cs:2,0.969);

\path [draw=black, line width=1pt]
(axis cs:3,0.968)
--(axis cs:3,0.998);

\path [draw=black, line width=1pt]
(axis cs:4,0.996)
--(axis cs:4,1);

\path [draw=white!29.8039215686275!black, line width=1pt]
(axis cs:0,0.443)
--(axis cs:0,0.581);

\path [draw=white!29.8039215686275!black, line width=1pt]
(axis cs:1,0.093)
--(axis cs:1,0.405);

\path [draw=white!29.8039215686275!black, line width=1pt]
(axis cs:2,0)
--(axis cs:2,0);

\path [draw=white!29.8039215686275!black, line width=1pt]
(axis cs:3,0.74)
--(axis cs:3,0.82);

\path [draw=white!29.8039215686275!black, line width=1pt]
(axis cs:4,0.565)
--(axis cs:4,0.589);

\addplot [line width=1pt, black, mark=asterisk, mark size=3, mark options={solid}]
table {%
0 0.977
1 0.993
2 0.937
3 0.983
4 0.998
};
\addlegendentry{cycle-cycle}
\addplot [line width=1pt, white!29.8039215686275!black, dashed, mark=diamond*, mark size=3, mark options={solid}]
table {%
0 0.512
1 0.249
2 0
3 0.78
4 0.577
};
\addlegendentry{cycle-tree}
\end{axis}

\end{tikzpicture}

%% file: figures/tex/real_rewire/frac_rewired_counts_F1_qual_rel_score_configuration_cora-ml_mcs_2.tex
\begin{tikzpicture}

\definecolor{color0}{rgb}{0.105882352941176,0.619607843137255,0.466666666666667}
\definecolor{color1}{rgb}{0.850980392156863,0.372549019607843,0.00784313725490196}
\definecolor{color2}{rgb}{0.458823529411765,0.43921568627451,0.701960784313725}

\begin{axis}[
axis line style={white!80!black},
legend cell align={left},
legend style={fill opacity=0.8, draw opacity=1, text opacity=1, draw=none},
tick align=outside,
tick pos=left,
x grid style={white!80!black},
xlabel={Fraction edges rewired},
xmajorgrids,
xmin=-0.05, xmax=1.05,
xtick style={color=white!15!black},
xtick={-0.2,0,0.2,0.4,0.6,0.8,1,1.2},
xticklabels={−0.2,0.0,0.2,0.4,0.6,0.8,1.0,1.2},
y grid style={white!80!black},
ylabel={Similarity score},
ymajorgrids,
ymin=-0.05, ymax=1.05,
ytick style={color=white!15!black},
ytick={-0.2,0,0.2,0.4,0.6,0.8,1,1.2},
yticklabels={−0.2,0.0,0.2,0.4,0.6,0.8,1.0,1.2}
]
\path [draw=black, fill=black, opacity=0.2]
(axis cs:0.01,0.993216272083227)
--(axis cs:0.01,0.989792203083711)
--(axis cs:0.05,0.958490399365209)
--(axis cs:0.1,0.91982980530032)
--(axis cs:0.25,0.801117196918572)
--(axis cs:0.5,0.625755749019761)
--(axis cs:1,0.446931271737777)
--(axis cs:1,0.463562257008967)
--(axis cs:1,0.463562257008967)
--(axis cs:0.5,0.644681457422169)
--(axis cs:0.25,0.809222002920867)
--(axis cs:0.1,0.923328646955506)
--(axis cs:0.05,0.961515715740206)
--(axis cs:0.01,0.993216272083227)
--cycle;

\path [draw=color0, fill=color0, opacity=0.2]
(axis cs:0.01,0.925685839939277)
--(axis cs:0.01,0.885060907686678)
--(axis cs:0.05,0.581622809855234)
--(axis cs:0.1,0.238839332992456)
--(axis cs:0.25,0.0415595387039647)
--(axis cs:0.5,0.0108086104150625)
--(axis cs:1,0.00385829749939577)
--(axis cs:1,0.00464485525748689)
--(axis cs:1,0.00464485525748689)
--(axis cs:0.5,0.0156941433871793)
--(axis cs:0.25,0.0535780424355363)
--(axis cs:0.1,0.275462101495476)
--(axis cs:0.05,0.641521625680662)
--(axis cs:0.01,0.925685839939277)
--cycle;

\path [draw=color1, fill=color1, opacity=0.2]
(axis cs:0.01,0.724436820165328)
--(axis cs:0.01,0.71936562219674)
--(axis cs:0.05,0.649426327566468)
--(axis cs:0.1,0.578342138896261)
--(axis cs:0.25,0.402950622841697)
--(axis cs:0.5,0.217287091861786)
--(axis cs:1,0.115533597358744)
--(axis cs:1,0.128218098506446)
--(axis cs:1,0.128218098506446)
--(axis cs:0.5,0.232721584214845)
--(axis cs:0.25,0.411353871626444)
--(axis cs:0.1,0.593972778068704)
--(axis cs:0.05,0.671809480021305)
--(axis cs:0.01,0.724436820165328)
--cycle;

\path [draw=color2, fill=color2, opacity=0.2]
(axis cs:0.01,0.920309196134855)
--(axis cs:0.01,0.904665361684864)
--(axis cs:0.05,0.772643791388788)
--(axis cs:0.1,0.624908961030332)
--(axis cs:0.25,0.347605464820782)
--(axis cs:0.5,0.164241421940454)
--(axis cs:1,0.0894044241602197)
--(axis cs:1,0.104225402338139)
--(axis cs:1,0.104225402338139)
--(axis cs:0.5,0.196452312100728)
--(axis cs:0.25,0.379331451346026)
--(axis cs:0.1,0.63977688745644)
--(axis cs:0.05,0.799016269384438)
--(axis cs:0.01,0.920309196134855)
--cycle;

\addplot [line width=1pt, black, mark=*, mark size=3, mark options={solid,draw=white}]
table {%
0 1
0.01 0.991518779745674
0.05 0.960001794917895
0.1 0.921433885466714
0.25 0.805723458225408
0.5 0.635218603220965
1 0.453486658459988
};
\addlegendentry{Global stats.}
\addplot [line width=1pt, color0, mark=triangle*, mark size=4, mark options={solid,rotate=180,draw=white}]
table {%
0 1
0.01 0.903132923990889
0.05 0.612437731523463
0.1 0.257150717243966
0.25 0.0468720866266648
0.5 0.0132332936934878
1 0.00425486868892209
};
\addlegendentry{Manual feat.}
\addplot [line width=1pt, color1, mark=triangle*, mark size=4, mark options={solid,draw=white}]
table {%
0 0.731825440927389
0.01 0.721878054727028
0.05 0.660548792751656
0.1 0.586541202735346
0.25 0.407079859450044
0.5 0.225264006934133
1 0.122725086823056
};
\addlegendentry{GCC}
\addplot [line width=1pt, color2, mark=square*, mark size=3, mark options={solid,draw=white}]
table {%
0 1
0.01 0.91366391235852
0.05 0.788109102449488
0.1 0.632342924243386
0.25 0.362568236151785
0.5 0.178428025480128
1 0.0962466845474052
};
\addlegendentry{GraphWave}
\end{axis}

\end{tikzpicture}

%% file: figures/tex/real_rewire/frac_rewired_counts_F1_qual_rel_score_configuration_citeseer_mcs_2.tex
\begin{tikzpicture}

\definecolor{color0}{rgb}{0.105882352941176,0.619607843137255,0.466666666666667}
\definecolor{color1}{rgb}{0.850980392156863,0.372549019607843,0.00784313725490196}
\definecolor{color2}{rgb}{0.458823529411765,0.43921568627451,0.701960784313725}

\begin{axis}[
axis line style={white!80!black},
legend cell align={left},
legend style={fill opacity=0.8, draw opacity=1, text opacity=1, draw=none},
tick align=outside,
tick pos=left,
x grid style={white!80!black},
xlabel={Fraction edges rewired},
xmajorgrids,
xmin=-0.05, xmax=1.05,
xtick style={color=white!15!black},
xtick={-0.2,0,0.2,0.4,0.6,0.8,1,1.2},
xticklabels={−0.2,0.0,0.2,0.4,0.6,0.8,1.0,1.2},
y grid style={white!80!black},
ylabel={Similarity score},
ymajorgrids,
ymin=-0.05, ymax=1.05,
ytick style={color=white!15!black},
ytick={-0.2,0,0.2,0.4,0.6,0.8,1,1.2},
yticklabels={−0.2,0.0,0.2,0.4,0.6,0.8,1.0,1.2}
]
\path [draw=black, fill=black, opacity=0.2]
(axis cs:0.01,0.99137621611837)
--(axis cs:0.01,0.988942900504314)
--(axis cs:0.05,0.951493517981453)
--(axis cs:0.1,0.909305597209986)
--(axis cs:0.25,0.783859574791725)
--(axis cs:0.5,0.629856372407409)
--(axis cs:1,0.429565963046598)
--(axis cs:1,0.47903502763123)
--(axis cs:1,0.47903502763123)
--(axis cs:0.5,0.651595825978044)
--(axis cs:0.25,0.814237729649709)
--(axis cs:0.1,0.917706565172663)
--(axis cs:0.05,0.95892561947594)
--(axis cs:0.01,0.99137621611837)
--cycle;

\path [draw=color0, fill=color0, opacity=0.2]
(axis cs:0.01,0.966606244983908)
--(axis cs:0.01,0.912693319175715)
--(axis cs:0.05,0.66198647188903)
--(axis cs:0.1,0.406634158902807)
--(axis cs:0.25,0.158418610834713)
--(axis cs:0.5,0.102512062957484)
--(axis cs:1,0.0396613835757496)
--(axis cs:1,0.0607690147862877)
--(axis cs:1,0.0607690147862877)
--(axis cs:0.5,0.120502673979273)
--(axis cs:0.25,0.200358378579168)
--(axis cs:0.1,0.427479761549382)
--(axis cs:0.05,0.696163913248968)
--(axis cs:0.01,0.966606244983908)
--cycle;

\path [draw=color1, fill=color1, opacity=0.2]
(axis cs:0.01,0.753703285494087)
--(axis cs:0.01,0.742442795577951)
--(axis cs:0.05,0.69339407028714)
--(axis cs:0.1,0.638614212728541)
--(axis cs:0.25,0.476219557173287)
--(axis cs:0.5,0.34979292793192)
--(axis cs:1,0.243149056110672)
--(axis cs:1,0.261266579927758)
--(axis cs:1,0.261266579927758)
--(axis cs:0.5,0.367950650813921)
--(axis cs:0.25,0.517756892360872)
--(axis cs:0.1,0.659074765114578)
--(axis cs:0.05,0.705301841342028)
--(axis cs:0.01,0.753703285494087)
--cycle;

\path [draw=color2, fill=color2, opacity=0.2]
(axis cs:0.01,0.984439821882196)
--(axis cs:0.01,0.980926895894636)
--(axis cs:0.05,0.888509409922577)
--(axis cs:0.1,0.756927158320214)
--(axis cs:0.25,0.483538864325164)
--(axis cs:0.5,0.302261229200755)
--(axis cs:1,0.168146376567369)
--(axis cs:1,0.193816263050757)
--(axis cs:1,0.193816263050757)
--(axis cs:0.5,0.319399265298487)
--(axis cs:0.25,0.517859670122161)
--(axis cs:0.1,0.786030936638227)
--(axis cs:0.05,0.89937390880194)
--(axis cs:0.01,0.984439821882196)
--cycle;

\addplot [line width=1pt, black, mark=*, mark size=3, mark options={solid,draw=white}]
table {%
0 1
0.01 0.990382864900061
0.05 0.955738533003817
0.1 0.913508202252025
0.25 0.798953046192778
0.5 0.641177530986098
1 0.458553954124123
};
\addlegendentry{Global stats.}
\addplot [line width=1pt, color0, mark=triangle*, mark size=4, mark options={solid,rotate=180,draw=white}]
table {%
0 1
0.01 0.941789224817939
0.05 0.676042984252608
0.1 0.417713869550967
0.25 0.17817706695503
0.5 0.111234125609697
1 0.0494216600432039
};
\addlegendentry{Manual feat.}
\addplot [line width=1pt, color1, mark=triangle*, mark size=4, mark options={solid,draw=white}]
table {%
0 0.751919648951262
0.01 0.748170977557019
0.05 0.699422480020017
0.1 0.647233847281183
0.25 0.497303800348159
0.5 0.359865111523377
1 0.252322388793745
};
\addlegendentry{GCC}
\addplot [line width=1pt, color2, mark=square*, mark size=3, mark options={solid,draw=white}]
table {%
0 1
0.01 0.982626712875177
0.05 0.895152692343511
0.1 0.77147904747922
0.25 0.500699267223662
0.5 0.310901415308935
1 0.180748312639555
};
\addlegendentry{GraphWave}
\end{axis}

\end{tikzpicture}

%% file: figures/tex/real_rewire/frac_rewired_counts_F1_qual_rel_score_configuration_polblogs_mcs_2.tex
\begin{tikzpicture}

\definecolor{color0}{rgb}{0.105882352941176,0.619607843137255,0.466666666666667}
\definecolor{color1}{rgb}{0.850980392156863,0.372549019607843,0.00784313725490196}
\definecolor{color2}{rgb}{0.458823529411765,0.43921568627451,0.701960784313725}

\begin{axis}[
axis line style={white!80!black},
legend cell align={left},
legend style={fill opacity=0.8, draw opacity=1, text opacity=1, draw=none, at={(1,0.62)}, anchor=east},
tick align=outside,
tick pos=left,
x grid style={white!80!black},
xlabel={Fraction edges rewired},
xmajorgrids,
xmin=-0.05, xmax=1.05,
xtick style={color=white!15!black},
xtick={-0.2,0,0.2,0.4,0.6,0.8,1,1.2},
xticklabels={−0.2,0.0,0.2,0.4,0.6,0.8,1.0,1.2},
y grid style={white!80!black},
ylabel={Similarity score},
ymajorgrids,
ymin=-0.05, ymax=1.05,
ytick style={color=white!15!black},
ytick={-0.2,0,0.2,0.4,0.6,0.8,1,1.2},
yticklabels={−0.2,0.0,0.2,0.4,0.6,0.8,1.0,1.2}
]
\path [draw=black, fill=black, opacity=0.2]
(axis cs:0.01,0.996862449911294)
--(axis cs:0.01,0.996403047208102)
--(axis cs:0.05,0.984289945068464)
--(axis cs:0.1,0.970334155640977)
--(axis cs:0.25,0.937159375870146)
--(axis cs:0.5,0.904252874685319)
--(axis cs:1,0.876304229159948)
--(axis cs:1,0.878362538143107)
--(axis cs:1,0.878362538143107)
--(axis cs:0.5,0.906237034956287)
--(axis cs:0.25,0.940844822384785)
--(axis cs:0.1,0.971598787450686)
--(axis cs:0.05,0.985283148471978)
--(axis cs:0.01,0.996862449911294)
--cycle;

\path [draw=color0, fill=color0, opacity=0.2]
(axis cs:0.01,0.397936100879951)
--(axis cs:0.01,0.295111552907347)
--(axis cs:0.05,0.0210507474611835)
--(axis cs:0.1,0.0108825853125745)
--(axis cs:0.25,0.0029913187569302)
--(axis cs:0.5,0.0018109061557846)
--(axis cs:1,0.000166074033203887)
--(axis cs:1,0.00194412314117591)
--(axis cs:1,0.00194412314117591)
--(axis cs:0.5,0.00425213056910172)
--(axis cs:0.25,0.00537372938838278)
--(axis cs:0.1,0.0139830549146474)
--(axis cs:0.05,0.0293645403603607)
--(axis cs:0.01,0.397936100879951)
--cycle;

\path [draw=color1, fill=color1, opacity=0.2]
(axis cs:0.01,0.66979056380473)
--(axis cs:0.01,0.653893374953245)
--(axis cs:0.05,0.628581154135426)
--(axis cs:0.1,0.596321461925874)
--(axis cs:0.25,0.496471436562096)
--(axis cs:0.5,0.396050376611553)
--(axis cs:1,0.282437262589318)
--(axis cs:1,0.305881660936038)
--(axis cs:1,0.305881660936038)
--(axis cs:0.5,0.406911424018761)
--(axis cs:0.25,0.532901153085231)
--(axis cs:0.1,0.622664279532582)
--(axis cs:0.05,0.647531888004717)
--(axis cs:0.01,0.66979056380473)
--cycle;

\path [draw=color2, fill=color2, opacity=0.2]
(axis cs:0.01,0.99128605538481)
--(axis cs:0.01,0.989300978781968)
--(axis cs:0.05,0.893910802349973)
--(axis cs:0.1,0.792862751162196)
--(axis cs:0.25,0.539209104094387)
--(axis cs:0.5,0.398457180366912)
--(axis cs:1,0.306907429446248)
--(axis cs:1,0.328455854683033)
--(axis cs:1,0.328455854683033)
--(axis cs:0.5,0.429661927833618)
--(axis cs:0.25,0.564492178613377)
--(axis cs:0.1,0.80373692379367)
--(axis cs:0.05,0.90900290123607)
--(axis cs:0.01,0.99128605538481)
--cycle;

\addplot [line width=1pt, black, mark=*, mark size=3, mark options={solid,draw=white}]
table {%
0 1
0.01 0.996644054482233
0.05 0.984798779642833
0.1 0.970966471545831
0.25 0.939072457482297
0.5 0.905292199562874
1 0.877317065121628
};
\addlegendentry{Global stats.}
\addplot [line width=1pt, color0, mark=triangle*, mark size=4, mark options={solid,rotate=180,draw=white}]
table {%
0 1
0.01 0.343947572144952
0.05 0.0255332768736963
0.1 0.0125720701035823
0.25 0.00403405559979814
0.5 0.00303151836244316
1 0.000870726183501079
};
\addlegendentry{Manual feat.}
\addplot [line width=1pt, color1, mark=triangle*, mark size=4, mark options={solid,draw=white}]
table {%
0 0.664659330203334
0.01 0.6626218051058
0.05 0.638877570919557
0.1 0.613423019921906
0.25 0.514686294823664
0.5 0.401691534098691
1 0.295450070605628
};
\addlegendentry{GCC}
\addplot [line width=1pt, color2, mark=square*, mark size=3, mark options={solid,draw=white}]
table {%
0 1
0.01 0.990459350229733
0.05 0.901512582564614
0.1 0.798299837477933
0.25 0.552470466225211
0.5 0.413741426332187
1 0.316271878738101
};
\addlegendentry{GraphWave}
\end{axis}

\end{tikzpicture}

%% file: figures/tex/synthetic_rewire/frac_rewired_diamond_marked_F1_qual_rel_score_mcs_2.tex
\begin{tikzpicture}

\definecolor{color0}{rgb}{0.105882352941176,0.619607843137255,0.466666666666667}
\definecolor{color1}{rgb}{0.850980392156863,0.372549019607843,0.00784313725490196}
\definecolor{color2}{rgb}{0.458823529411765,0.43921568627451,0.701960784313725}

\begin{axis}[
axis line style={white!80!black},
legend cell align={left},
legend style={fill opacity=0.8, draw opacity=1, text opacity=1, draw=none},
tick align=outside,
tick pos=left,
x grid style={white!80!black},
xlabel={Fraction edges rewired},
xmajorgrids,
xmin=-0.05, xmax=1.05,
xtick style={color=white!15!black},
xtick={-0.2,0,0.2,0.4,0.6,0.8,1,1.2},
xticklabels={−0.2,0.0,0.2,0.4,0.6,0.8,1.0,1.2},
y grid style={white!80!black},
ylabel={Similarity score},
ymajorgrids,
ymin=-0.05, ymax=1.05,
ytick style={color=white!15!black},
ytick={-0.2,0,0.2,0.4,0.6,0.8,1,1.2},
yticklabels={−0.2,0.0,0.2,0.4,0.6,0.8,1.0,1.2}
]
\path [draw=black, fill=black, opacity=0.2]
(axis cs:0,1)
--(axis cs:0,1)
--(axis cs:0.01,0.634795328387451)
--(axis cs:0.05,0.37979801005485)
--(axis cs:0.1,0.630867527476444)
--(axis cs:0.25,0.0395746438507714)
--(axis cs:0.5,0)
--(axis cs:1,0)
--(axis cs:1,0)
--(axis cs:1,0)
--(axis cs:0.5,0)
--(axis cs:0.25,0.279741166605099)
--(axis cs:0.1,0.634918734905901)
--(axis cs:0.05,0.633734422144804)
--(axis cs:0.01,0.637762260893024)
--(axis cs:0,1)
--cycle;

\path [draw=color0, fill=color0, opacity=0.2]
(axis cs:0,1)
--(axis cs:0,1)
--(axis cs:0.01,0.89510197096532)
--(axis cs:0.05,0.746924750814952)
--(axis cs:0.1,0.668274796938038)
--(axis cs:0.25,0.404394743020316)
--(axis cs:0.5,0.0915192426929638)
--(axis cs:1,0)
--(axis cs:1,0.0217178197149355)
--(axis cs:1,0.0217178197149355)
--(axis cs:0.5,0.222214625326783)
--(axis cs:0.25,0.530594632701068)
--(axis cs:0.1,0.772358264761044)
--(axis cs:0.05,0.823911231010569)
--(axis cs:0.01,0.936429287700276)
--(axis cs:0,1)
--cycle;

\path [draw=color1, fill=color1, opacity=0.2]
(axis cs:0,0.785339451491743)
--(axis cs:0,0.63016157989228)
--(axis cs:0.01,0.560573593754094)
--(axis cs:0.05,0.581557988743475)
--(axis cs:0.1,0.557766753983192)
--(axis cs:0.25,0.424932427448881)
--(axis cs:0.5,0.289363137510474)
--(axis cs:1,0.0449242607152453)
--(axis cs:1,0.128845367209172)
--(axis cs:1,0.128845367209172)
--(axis cs:0.5,0.426218166251553)
--(axis cs:0.25,0.507606220830036)
--(axis cs:0.1,0.621269299900994)
--(axis cs:0.05,0.683405988473932)
--(axis cs:0.01,0.656336025612211)
--(axis cs:0,0.785339451491743)
--cycle;

\path [draw=color2, fill=color2, opacity=0.2]
(axis cs:0,1)
--(axis cs:0,1)
--(axis cs:0.01,0.663402442890261)
--(axis cs:0.05,0.417272972863572)
--(axis cs:0.1,0.299615236742926)
--(axis cs:0.25,0.128865450844526)
--(axis cs:0.5,0)
--(axis cs:1,0)
--(axis cs:1,0)
--(axis cs:1,0)
--(axis cs:0.5,0.0176994943583456)
--(axis cs:0.25,0.172843111550801)
--(axis cs:0.1,0.391189030894669)
--(axis cs:0.05,0.539331506866974)
--(axis cs:0.01,0.752906887476588)
--(axis cs:0,1)
--cycle;

\addplot [line width=1pt, black, mark=*, mark size=3, mark options={solid,draw=white}]
table {%
0 1
0.01 0.636260571940902
0.05 0.508188089368127
0.1 0.632957289490248
0.25 0.158946080369185
0.5 0
1 0
};
\addlegendentry{Global stats.}
\addplot [line width=1pt, color0, mark=triangle*, mark size=4, mark options={solid,rotate=180,draw=white}]
table {%
0 1
0.01 0.915322075846791
0.05 0.786994751331238
0.1 0.723053023591957
0.25 0.46848822223131
0.5 0.152644851939712
1 0.00723927323831184
};
\addlegendentry{Manual feat.}
\addplot [line width=1pt, color1, mark=triangle*, mark size=4, mark options={solid,draw=white}]
table {%
0 0.722774432567027
0.01 0.608311749209922
0.05 0.631444064373138
0.1 0.588965262395019
0.25 0.465642979639123
0.5 0.348653472318069
1 0.0871868206201822
};
\addlegendentry{GCC}
\addplot [line width=1pt, color2, mark=square*, mark size=3, mark options={solid,draw=white}]
table {%
0 1
0.01 0.70799379039512
0.05 0.479894324780788
0.1 0.347870774604011
0.25 0.150309875653888
0.5 0.00589983145278188
1 0
};
\addlegendentry{GraphWave}
\end{axis}

\end{tikzpicture}

%% file: figures/tex/synthetic_rewire/frac_rewired_friendship_marked_F1_qual_rel_score_mcs_2.tex
\begin{tikzpicture}

\definecolor{color0}{rgb}{0.105882352941176,0.619607843137255,0.466666666666667}
\definecolor{color1}{rgb}{0.850980392156863,0.372549019607843,0.00784313725490196}
\definecolor{color2}{rgb}{0.458823529411765,0.43921568627451,0.701960784313725}

\begin{axis}[
axis line style={white!80!black},
legend cell align={left},
legend style={fill opacity=0.8, draw opacity=1, text opacity=1, draw=none},
tick align=outside,
tick pos=left,
x grid style={white!80!black},
xlabel={Fraction edges rewired},
xmajorgrids,
xmin=-0.05, xmax=1.05,
xtick style={color=white!15!black},
xtick={-0.2,0,0.2,0.4,0.6,0.8,1,1.2},
xticklabels={−0.2,0.0,0.2,0.4,0.6,0.8,1.0,1.2},
y grid style={white!80!black},
ylabel={Similarity score},
ymajorgrids,
ymin=-0.05, ymax=1.05,
ytick style={color=white!15!black},
ytick={-0.2,0,0.2,0.4,0.6,0.8,1,1.2},
yticklabels={−0.2,0.0,0.2,0.4,0.6,0.8,1.0,1.2}
]
\path [draw=black, fill=black, opacity=0.2]
(axis cs:0,1)
--(axis cs:0,1)
--(axis cs:0.01,0.990605508627546)
--(axis cs:0.05,0.964030078142453)
--(axis cs:0.1,0.924188943288123)
--(axis cs:0.25,0.659215820099352)
--(axis cs:0.5,0)
--(axis cs:1,0)
--(axis cs:1,0)
--(axis cs:1,0)
--(axis cs:0.5,0)
--(axis cs:0.25,0.830155788530294)
--(axis cs:0.1,0.930681753875514)
--(axis cs:0.05,0.967858704539715)
--(axis cs:0.01,0.9915266668866)
--(axis cs:0,1)
--cycle;

\path [draw=color0, fill=color0, opacity=0.2]
(axis cs:0,1)
--(axis cs:0,1)
--(axis cs:0.01,0.414558087707144)
--(axis cs:0.05,0.189276054057541)
--(axis cs:0.1,0.0583687045586547)
--(axis cs:0.25,0)
--(axis cs:0.5,0)
--(axis cs:1,0)
--(axis cs:1,0)
--(axis cs:1,0)
--(axis cs:0.5,0)
--(axis cs:0.25,0)
--(axis cs:0.1,0.134681956409125)
--(axis cs:0.05,0.300812651760908)
--(axis cs:0.01,0.587591176125952)
--(axis cs:0,1)
--cycle;

\path [draw=color1, fill=color1, opacity=0.2]
(axis cs:0,0.807551766138855)
--(axis cs:0,0.637669499583108)
--(axis cs:0.01,0.599253911371688)
--(axis cs:0.05,0.496969780011031)
--(axis cs:0.1,0.418439010959913)
--(axis cs:0.25,0.105549014796986)
--(axis cs:0.5,0)
--(axis cs:1,0)
--(axis cs:1,0)
--(axis cs:1,0)
--(axis cs:0.5,0)
--(axis cs:0.25,0.169541845317762)
--(axis cs:0.1,0.500709199174866)
--(axis cs:0.05,0.584269239484672)
--(axis cs:0.01,0.675541069779458)
--(axis cs:0,0.807551766138855)
--cycle;

\path [draw=color2, fill=color2, opacity=0.2]
(axis cs:0,1)
--(axis cs:0,1)
--(axis cs:0.01,0.657677611068171)
--(axis cs:0.05,0.429985550042627)
--(axis cs:0.1,0.243277607809018)
--(axis cs:0.25,0.024462439111914)
--(axis cs:0.5,0)
--(axis cs:1,0)
--(axis cs:1,0)
--(axis cs:1,0)
--(axis cs:0.5,0)
--(axis cs:0.25,0.0785122053187186)
--(axis cs:0.1,0.35800943548751)
--(axis cs:0.05,0.538519207473418)
--(axis cs:0.01,0.764051367399262)
--(axis cs:0,1)
--cycle;

\addplot [line width=1pt, black, mark=*, mark size=3, mark options={solid,draw=white}]
table {%
0 1
0.01 0.991073521500448
0.05 0.96600000648364
0.1 0.927591009935351
0.25 0.771339257316674
0.5 0
1 0
};
\addlegendentry{Global stats.}
\addplot [line width=1pt, color0, mark=triangle*, mark size=4, mark options={solid,rotate=180,draw=white}]
table {%
0 1
0.01 0.505756950653876
0.05 0.244337217915859
0.1 0.0950961912733488
0.25 0
0.5 0
1 0
};
\addlegendentry{Manual feat.}
\addplot [line width=1pt, color1, mark=triangle*, mark size=4, mark options={solid,draw=white}]
table {%
0 0.708456628042219
0.01 0.637231649857901
0.05 0.54247574870644
0.1 0.460219574170449
0.25 0.137210918052882
0.5 0
1 0
};
\addlegendentry{GCC}
\addplot [line width=1pt, color2, mark=square*, mark size=3, mark options={solid,draw=white}]
table {%
0 1
0.01 0.711507530883713
0.05 0.486539175610388
0.1 0.297230470491685
0.25 0.0508705692060587
0.5 0
1 0
};
\addlegendentry{GraphWave}
\end{axis}

\end{tikzpicture}

%% file: figures/tex/synthetic_rewire/frac_rewired_star_marked_F1_qual_rel_score_mcs_2.tex
\begin{tikzpicture}

\definecolor{color0}{rgb}{0.105882352941176,0.619607843137255,0.466666666666667}
\definecolor{color1}{rgb}{0.850980392156863,0.372549019607843,0.00784313725490196}
\definecolor{color2}{rgb}{0.458823529411765,0.43921568627451,0.701960784313725}

\begin{axis}[
axis line style={white!80!black},
legend cell align={left},
legend style={fill opacity=0.8, draw opacity=1, text opacity=1, draw=none},
tick align=outside,
tick pos=left,
x grid style={white!80!black},
xlabel={Fraction edges rewired},
xmajorgrids,
xmin=-0.05, xmax=1.05,
xtick style={color=white!15!black},
xtick={-0.2,0,0.2,0.4,0.6,0.8,1,1.2},
xticklabels={−0.2,0.0,0.2,0.4,0.6,0.8,1.0,1.2},
y grid style={white!80!black},
ylabel={Similarity score},
ymajorgrids,
ymin=-0.05, ymax=1.05,
ytick style={color=white!15!black},
ytick={-0.2,0,0.2,0.4,0.6,0.8,1,1.2},
yticklabels={−0.2,0.0,0.2,0.4,0.6,0.8,1.0,1.2}
]
\path [draw=black, fill=black, opacity=0.2]
(axis cs:0,1)
--(axis cs:0,1)
--(axis cs:0.01,0.975491150569987)
--(axis cs:0.05,0.767982436477459)
--(axis cs:0.1,0.314379747626482)
--(axis cs:0.25,0.295354788817845)
--(axis cs:0.5,0.0549698081676743)
--(axis cs:1,0)
--(axis cs:1,0)
--(axis cs:1,0)
--(axis cs:0.5,0.387658350579627)
--(axis cs:0.25,0.755928645643245)
--(axis cs:0.1,0.813633546142931)
--(axis cs:0.05,0.965211493097701)
--(axis cs:0.01,0.982247982199749)
--(axis cs:0,1)
--cycle;

\path [draw=color0, fill=color0, opacity=0.2]
(axis cs:0,1)
--(axis cs:0,1)
--(axis cs:0.01,0.955355889581235)
--(axis cs:0.05,0.919902481762549)
--(axis cs:0.1,0.828407290585559)
--(axis cs:0.25,0.436318184608313)
--(axis cs:0.5,0.0725535941751976)
--(axis cs:1,0)
--(axis cs:1,0.0136453476557361)
--(axis cs:1,0.0136453476557361)
--(axis cs:0.5,0.236422031345903)
--(axis cs:0.25,0.524057422494323)
--(axis cs:0.1,0.897157638915998)
--(axis cs:0.05,0.954573935481228)
--(axis cs:0.01,0.981292003345851)
--(axis cs:0,1)
--cycle;

\path [draw=color1, fill=color1, opacity=0.2]
(axis cs:0,0.837209302325581)
--(axis cs:0,0.682020802377414)
--(axis cs:0.01,0.740006882172761)
--(axis cs:0.05,0.731486876138624)
--(axis cs:0.1,0.694971524523953)
--(axis cs:0.25,0.515209033705849)
--(axis cs:0.5,0.272482397433695)
--(axis cs:1,0.0826952418521511)
--(axis cs:1,0.164867587292723)
--(axis cs:1,0.164867587292723)
--(axis cs:0.5,0.35516926310409)
--(axis cs:0.25,0.640696367237152)
--(axis cs:0.1,0.828235308232254)
--(axis cs:0.05,0.793090635194565)
--(axis cs:0.01,0.814518944575429)
--(axis cs:0,0.837209302325581)
--cycle;

\path [draw=color2, fill=color2, opacity=0.2]
(axis cs:0,1)
--(axis cs:0,1)
--(axis cs:0.01,0.875884496237705)
--(axis cs:0.05,0.796963223622584)
--(axis cs:0.1,0.694012093905677)
--(axis cs:0.25,0.233427110980724)
--(axis cs:0.5,0.0225619681696305)
--(axis cs:1,0)
--(axis cs:1,0)
--(axis cs:1,0)
--(axis cs:0.5,0.0651205110724144)
--(axis cs:0.25,0.334380453636867)
--(axis cs:0.1,0.799063154074675)
--(axis cs:0.05,0.874027998094115)
--(axis cs:0.01,0.921386492732534)
--(axis cs:0,1)
--cycle;

\addplot [line width=1pt, black, mark=*, mark size=3, mark options={solid,draw=white}]
table {%
0 1
0.01 0.979074593048863
0.05 0.896963026075371
0.1 0.565215987162407
0.25 0.527217952775744
0.5 0.220334085620527
1 0
};
\addlegendentry{Global stats.}
\addplot [line width=1pt, color0, mark=triangle*, mark size=4, mark options={solid,rotate=180,draw=white}]
table {%
0 1
0.01 0.96882968187784
0.05 0.936583818097228
0.1 0.865357146476578
0.25 0.480341141083228
0.5 0.153311770739749
1 0.00454844921857871
};
\addlegendentry{Manual feat.}
\addplot [line width=1pt, color1, mark=triangle*, mark size=4, mark options={solid,draw=white}]
table {%
0 0.753163281654245
0.01 0.770732401549306
0.05 0.761797443715283
0.1 0.760329355925192
0.25 0.573134508184283
0.5 0.312593500392843
1 0.123538126774163
};
\addlegendentry{GCC}
\addplot [line width=1pt, color2, mark=square*, mark size=3, mark options={solid,draw=white}]
table {%
0 1
0.01 0.898859925961384
0.05 0.841276187318668
0.1 0.748284033372579
0.25 0.285397310261753
0.5 0.0441912072897159
1 0
};
\addlegendentry{GraphWave}
\end{axis}

\end{tikzpicture}

%% file: figures/tex/synthetic_rewire/frac_rewired_wheel_marked_F1_qual_rel_score_mcs_2.tex
\begin{tikzpicture}

\definecolor{color0}{rgb}{0.105882352941176,0.619607843137255,0.466666666666667}
\definecolor{color1}{rgb}{0.850980392156863,0.372549019607843,0.00784313725490196}
\definecolor{color2}{rgb}{0.458823529411765,0.43921568627451,0.701960784313725}

\begin{axis}[
axis line style={white!80!black},
legend cell align={left},
legend style={fill opacity=0.8, draw opacity=1, text opacity=1, draw=none},
tick align=outside,
tick pos=left,
x grid style={white!80!black},
xlabel={Fraction edges rewired},
xmajorgrids,
xmin=-0.05, xmax=1.05,
xtick style={color=white!15!black},
xtick={-0.2,0,0.2,0.4,0.6,0.8,1,1.2},
xticklabels={−0.2,0.0,0.2,0.4,0.6,0.8,1.0,1.2},
y grid style={white!80!black},
ylabel={Similarity score},
ymajorgrids,
ymin=-0.05, ymax=1.05,
ytick style={color=white!15!black},
ytick={-0.2,0,0.2,0.4,0.6,0.8,1,1.2},
yticklabels={−0.2,0.0,0.2,0.4,0.6,0.8,1.0,1.2}
]
\path [draw=black, fill=black, opacity=0.2]
(axis cs:0,1)
--(axis cs:0,1)
--(axis cs:0.01,0.986448515939823)
--(axis cs:0.05,0.944359868463629)
--(axis cs:0.1,0.85008639024743)
--(axis cs:0.25,0.74066087986255)
--(axis cs:0.5,0.527196984909104)
--(axis cs:1,0.0349129504353405)
--(axis cs:1,0.0440211138153941)
--(axis cs:1,0.0440211138153941)
--(axis cs:0.5,0.550965141851536)
--(axis cs:0.25,0.748252634711413)
--(axis cs:0.1,0.878303426273199)
--(axis cs:0.05,0.951697118292087)
--(axis cs:0.01,0.988189898411474)
--(axis cs:0,1)
--cycle;

\path [draw=color0, fill=color0, opacity=0.2]
(axis cs:0,1)
--(axis cs:0,1)
--(axis cs:0.01,0.902316718509339)
--(axis cs:0.05,0.707985296886453)
--(axis cs:0.1,0.460486459953031)
--(axis cs:0.25,0.171909108424902)
--(axis cs:0.5,0)
--(axis cs:1,0)
--(axis cs:1,0)
--(axis cs:1,0)
--(axis cs:0.5,0)
--(axis cs:0.25,0.236041000670131)
--(axis cs:0.1,0.575005141951538)
--(axis cs:0.05,0.786037466401198)
--(axis cs:0.01,0.92538904212211)
--(axis cs:0,1)
--cycle;

\path [draw=color1, fill=color1, opacity=0.2]
(axis cs:0,0.815533980582524)
--(axis cs:0,0.737430167597765)
--(axis cs:0.01,0.629192943933504)
--(axis cs:0.05,0.543330454087447)
--(axis cs:0.1,0.380285591576168)
--(axis cs:0.25,0.208081935886565)
--(axis cs:0.5,0)
--(axis cs:1,0)
--(axis cs:1,0)
--(axis cs:1,0)
--(axis cs:0.5,0)
--(axis cs:0.25,0.306646660012022)
--(axis cs:0.1,0.523080944547576)
--(axis cs:0.05,0.69262818821694)
--(axis cs:0.01,0.70855929900769)
--(axis cs:0,0.815533980582524)
--cycle;

\path [draw=color2, fill=color2, opacity=0.2]
(axis cs:0,1)
--(axis cs:0,0.982758620689655)
--(axis cs:0.01,0.812928671555913)
--(axis cs:0.05,0.515870684502455)
--(axis cs:0.1,0.331147016560586)
--(axis cs:0.25,0.0789584580804537)
--(axis cs:0.5,0)
--(axis cs:1,0)
--(axis cs:1,0)
--(axis cs:1,0)
--(axis cs:0.5,0.0207483020441772)
--(axis cs:0.25,0.138729595838075)
--(axis cs:0.1,0.422710759067652)
--(axis cs:0.05,0.625020501470696)
--(axis cs:0.01,0.867747786437616)
--(axis cs:0,1)
--cycle;

\addplot [line width=1pt, black, mark=*, mark size=3, mark options={solid,draw=white}]
table {%
0 1
0.01 0.987314186972972
0.05 0.948085154632807
0.1 0.863974719831894
0.25 0.744471578093861
0.5 0.538575694715956
1 0.0395868609001286
};
\addlegendentry{Global stats.}
\addplot [line width=1pt, color0, mark=triangle*, mark size=4, mark options={solid,rotate=180,draw=white}]
table {%
0 1
0.01 0.915870895212659
0.05 0.746566404907039
0.1 0.520814228983697
0.25 0.206400050171548
0.5 0
1 0
};
\addlegendentry{Manual feat.}
\addplot [line width=1pt, color1, mark=triangle*, mark size=4, mark options={solid,draw=white}]
table {%
0 0.775932861755461
0.01 0.670322575285596
0.05 0.615501020321743
0.1 0.447036161824295
0.25 0.25631898410264
0.5 0
1 0
};
\addlegendentry{GCC}
\addplot [line width=1pt, color2, mark=square*, mark size=3, mark options={solid,draw=white}]
table {%
0 0.994252873563218
0.01 0.842639384482825
0.05 0.573263414730558
0.1 0.380334403416833
0.25 0.106474015275125
0.5 0.00691610068139241
1 0
};
\addlegendentry{GraphWave}
\end{axis}

\end{tikzpicture}

%% file: tables/generative_models_with_ground_truth.tex
\begin{small}
\begin{sc}
\begin{tabular}{@{}r|ccc|ccc|ccc@{}}
\toprule    
  & \multicolumn{3}{c|}{Cora-ML}   & \multicolumn{3}{c|}{Citeseer} & \multicolumn{3}{c}{PolBlogs}      \\
\midrule
 &  \thead{Ground \\ truth} & \thead{NetGAN} & \thead{CELL} & \thead{Ground \\ truth} & NetGAN & CELL & \thead{Ground \\ truth} & NetGAN & CELL \\
\midrule
Max degree &  $238$  &  $216.00 \pm 11.14$  &  $191.33 \pm 6.51$  &  $76$  &  $88.00 \pm 1.73$  &  $67 \pm 1.73$  &  $303$  &  $279.33 \pm 14.57$  &  $250.00 \pm 6.24$  \\ 
Assortativity &  $-0.08$  &  $-0.08 \pm 0.00$  &  $-0.07 \pm 0.00$  &  $-0.19$  &  $-0.16 \pm 0.00$  &  $-0.2 \pm 0.01$  &  $-0.22$  &  $-0.25 \pm 0.01$  &  $-0.25 \pm 0.00$  \\ 
Triangle count &  $2,802$  &  $1,772.33 \pm 17.39$  &  $1,409.67 \pm 50.85$  &  $304$  &  $195.33 \pm 5.86$  &  $88.33 \pm 10.69$  &  $61,108$  &  $36,294.33 \pm 348.62$  &  $44,862.33 \pm 275.61$  \\ 
Square count &  $14,268$  &  $6,741.33 \pm 229.38$  &  $7,035.00 \pm 424.89$  &  $1,441$  &  $371.33 \pm 35.22$  &  $420.33 \pm 16.26$  &  $2,654,319$  &  $1,350,075.67 \pm 17073.43$  &  $1,801,022.33 \pm 7383.63$  \\ 
Power law exp. &  $1.86$  &  $1.81 \pm 0.00$  &  $1.82 \pm 0.00$  &  $2.45$  &  $2.32 \pm 0.01$  &  $2.39 \pm 0.00$  &  $1.44$  &  $1.41 \pm 0.00$  &  $1.43 \pm 0.00$  \\ 
Clustering coeff. &  $0.08$  &  $0.06 \pm 0.00$  &  $0.05 \pm 0.00$  &  $0.04$  &  $0.03 \pm 0.00$  &  $0.02 \pm 0.00$  &  $0.19$  &  $0.13 \pm 0.00$  &  $0.15 \pm 0.00$  \\ 
Charc. path len. &  $5.63$  &  $5.23 \pm 0.03$  &  $5.24 \pm 0.05$  &  $8.02$  &  $6.49 \pm 0.20$  &  $6.33 \pm 0.04$  &  $2.82$  &  $2.67 \pm 0.00$  &  $2.77 \pm 0.01$  \\ 
\midrule
$s_{\wDCA}$ -- Manual & $1.00$  &  $0.11 \pm 0.01$  &  $0.08 \pm 0.00$  &  $1.00$  &  $0.22 \pm 0.01$  &  $0.17 \pm 0.03$  &  $1.00$  &  $0.00 \pm 0.00$  &  $0.00 \pm 0.00$ \\
$s_{\wDCA}$ -- GCC & $0.73$  &  $0.51 \pm 0.00$  &  $0.43 \pm 0.00$  &  $0.75$  &  $0.50 \pm 0.01$  &  $0.44 \pm 0.01$  &  $0.66$  &  $0.38 \pm 0.02$  &  $0.48 \pm 0.04$ \\
$s_{\wDCA}$ -- Graphwave & $1.00$  &  $0.52 \pm 0.01$  &  $0.46 \pm 0.02$  &  $1.00$  &  $0.53 \pm 0.03$  &  $0.50 \pm 0.02$  &  $1.00$  &  $0.50 \pm 0.01$  &  $0.70 \pm 0.01$ \\
\bottomrule
\end{tabular}
\end{sc}
\end{small}

%% file: tables/mainpaper_synthetic_graphwave_gm.tex
\begin{small}
\begin{sc}
\begin{tabular}{@{}r|ccc|ccc|ccc@{}}
\toprule
  & \multicolumn{3}{c|}{diamond}   & \multicolumn{3}{c|}{friendship} & \multicolumn{3}{c}{wheel}      \\
\midrule
 &  \thead{Ground \\ truth} & \thead{NetGAN} & \thead{CELL} & \thead{Ground \\ truth} & NetGAN & CELL & \thead{Ground \\ truth} & NetGAN & CELL \\
\midrule
Max degree &  $5$  &  $6.33 \pm 0.58$  &  $7.00 \pm 1.00$  &  $6$  &  $6.67 \pm 0.58$  &  $7.0 \pm 0.00$  &  $6$  &  $8.00 \pm 0.00$  &  $8.00 \pm 1.00$  \\ 
Assortativity &  $0.05$  &  $-0.10 \pm 0.03$  &  $0.08 \pm 0.02$  &  $0.19$  &  $-0.10 \pm 0.01$  &  $0.08 \pm 0.02$  &  $0.46$  &  $-0.00 \pm 0.00$  &  $0.14 \pm 0.01$  \\ 
Triangle count &  $0$  &  $116.67 \pm 3.79$  &  $0.00 \pm 0.00$  &  $100$  &  $123.00 \pm 3.00$  &  $34.67 \pm 6.11$  &  $47$  &  $251.33 \pm 13.01$  &  $17.33 \pm 0.58$  \\ 
Square count &  $0$  &  $51.0 \pm 9.85$  &  $14.00 \pm 5.57$  &  $0$  &  $65.00 \pm 6.24$  &  $24.67 \pm 4.04$  &  $284$  &  $192.67 \pm 2.08$  &  $127.33 \pm 11.24$  \\ 
Power law exp. &  $2.51$  &  $2.72 \pm 0.01$  &  $2.66 \pm 0.01$  &  $2.45$  &  $2.59 \pm 0.00$  &  $2.56 \pm 0.01$  &  $2.16$  &  $2.25 \pm 0.01$  &  $2.23 \pm 0.00$  \\ 
Clustering coeff. &  $0.00$  &  $0.14 \pm 0.01$  &  $0.00 \pm 0.00$  &  $0.11$  &  $0.12 \pm 0.00$  &  $0.03 \pm 0.01$  &  $0.04$  &  $0.16 \pm 0.01$  &  $0.01 \pm 0.00$  \\ 
Charc. path len. &  $82.72$  &  $11.71 \pm 1.28$  &  $19.60 \pm 0.91$  &  $51.44$  &  $27.77 \pm 0.32$  &  $15.56 \pm 0.26$  &  $39.85$  &  $19.44 \pm 3.33$  &  $11.25 \pm 0.06$  \\ 
\midrule
$s_{\wDCA}$ --  Manual&  $1.00 \pm 0.00$  &  $0.37 \pm 0.04$  &  $0.68 \pm 0.04$  &  $1.00 \pm 0.00$  &  $0.45 \pm 0.02$  &  $0.48 \pm 0.01$  &  $1.00 \pm 0.00$  &  $0.22 \pm 0.06$  &  $0.29 \pm 0.01$  \\ 
$s_{\wDCA}$ --  GCC &  $0.80 \pm 0.01$  &  $0.38 \pm 0.01$  &  $0.64 \pm 0.03$  &  $0.77 \pm 0.00$  &  $0.52 \pm 0.01$  &  $0.57 \pm 0.01$  &  $0.70 \pm 0.02$  &  $0.34 \pm 0.01$  &  $0.42 \pm 0.01$  \\ 
$s_{\wDCA}$ --  Graphwave &  $1.00 \pm 0.00$  &  $0.17 \pm 0.02$  &  $0.34 \pm 0.01$  &  $1.00 \pm 0.00$  &  $0.31 \pm 0.02$  &  $0.39 \pm 0.03$  &  $1.00 \pm 0.00$  &  $0.22 \pm 0.01$  &  $0.20 \pm 0.01$   \\

\bottomrule
\end{tabular}
\end{sc}
\end{small}

%% file: Sections/discussion.tex
\section{Conclusion}
We establish GraphDCA as a graph similarity evaluation framework able to compensate the deficiencies of global statistics by instead comparing graphs in terms of their local structural properties.
Using three different feature extractors producing node role representations, GraphDCA is able to recognize graphs with similar subgraph structures and differentiate between graphs not sharing such similarities.
We showed how GraphDCA can be applied to evaluate generative models for graphs and concluded that while current state-of-the-art models are able to capture many global statistics of training graphs, improvements for reproducing local structures are needed.
A natural next step is to use the insights provided by GraphDCA to further develop and enhance generative models for graphs with this capability. 
Though GraphDCA can be applied to any graphs given suitable feature extractors, we have in this work focused on plain, undirected and unweighted graphs and leave evaluation using attributed, directed and/or weighted graphs to future work.


%% file: Sections/appendix.tex
\section{Graph Statistics}
\label{sec:app:gstats}
We provide the exact definitions of global graph statistics used in our experiments in Table~\ref{tab:graph_stats}.

\begin{table}[b]
\begin{small}
    \centering
    \vskip -0.2in
    \caption{Graph statistics for a graph $G = (V, E)$ with $N=|V|$ nodes. The table is extracted from \cite{bojchevski_netgan_2018} and \cite{rendsburg_netgan_2020}.}
    \label{tab:graph_stats}
    \begin{tabular}{p{0.25\linewidth} c p{0.45\linewidth}}
        \textbf{\textsc{Statistic name}} & \textbf{\textsc{Computation}} & \textbf{\textsc{Description}} \\
        \midrule
        \textsc{Max degree} & $\displaystyle \max_{v \in V} d(v)$ & Maximum degree of all nodes in a graph with $d(v)$ denoting the degree of node $v$. \\
        \textsc{Assortativity} & $ \frac{\mathrm{Cov}(X, Y)}{\sigma_X \sigma_Y}$ & Pearson correlation of degrees of connected nodes where the $(x_i,y_i)$ pairs are the degrees of connected nodes \cite{newman2003mixing}.\\
        \textsc{Global clustering coef.} & $3 \times \frac{\mathrm{Tr} \mathbf{A}^3}{\sum_{i \neq j} [\mathbf{A}^2]_{i,j}}$ & The ratio of number of closed triplets to the total number of triplets. Measures the degree to which nodes in a graph tend to cluster together \cite{newman2003structure}. \\
        \textsc{Power law exponent} & $1 + N \left( \sum_{v \in V} \log \frac{d(v)}{d_{\mathrm{min}}}\right)^{-1}$ & Exponent of the power law distribution where $d_{\mathrm{min}}$ denotes the minimum degree in the graph \cite{newman2018networks}. \\
        \textsc{Characteristic path length} & $\frac{1}{N(N-1)} \sum_{u \neq v} d(u, v)$ & Average shortest path length where $d(u, v)$ is the shortest path length between nodes $u$ and $v$.
    \end{tabular}
\end{small}
\end{table}

\section{Additional Details}


\subsection{GraphDCA Hyperparameters} \label{sec:app:hyperparams}
The GraphDCA framework naturally inherits hyperparameters of the chosen feature extractor model $f$ and hyperparameters associated to the Delaunay graph approximation performed in DCA~\cite{anonymous2022delaunay}. Originally, DCA has four hyperparameters: $T$ affecting the number of found Delaunay edges, $B$ used as an optional parameter for reducing the number of Delaunay edges, and $mcs$ determining the number of points needed to form a cluster. In  our experiments, we use default options for $T = 10,000$ and $B = 1.0$, and set $mcs = 2$.

Similarly, we used the default hyperparameters for GraphWave, namely: $d=50$ (i.e.\ $25$ sample points of the empirical characteristic function), order $30$ Chebyshev polynomials and two filter values $t$ selected automatically as described in their paper \cite{donnat_learning_2018}. The resulting representations are of dimension $100$.

For GCC \cite{qiu_gcc_2020}, we used the model pretrained using MOCO available via the GCC Github repository \url{https://github.com/THUDM/GCC}\footnote{Specifically, the model specified as \path{Pretrain_moco_True_dgl_gin_layer_5_lr_0.005_decay_1e-05_bsz_32_hid_64_samples_2000_nce_t_0.07_nce_k_16384_rw_hops_256_restart_prob_0.8_aug_1st_ft_False_deg_16_pos_32_momentum_0.999} was used.}. This model outputs representations of dimension $d=64$.

The manual features, described in Section~\ref{sec:gdca:feature_extractors}, use egonet radii $\rho \in \{1, 2, 3, 4\}$ and compute the following commonly used egonet graph statistics as features for each $\rho$: number of nodes, average and max node degree, number of triangles, global clustering coefficient \cite{newman2003structure} and assortativity coefficient \cite{newman2003mixing}.
Thus, the resulting representations are of dimension $24$.

\subsection{Gradual Structure Perturbation}
\label{sec:app:rewire}

In this section, we provide detailed descriptions of the rewiring procedures used in the experiments in Section \ref{sec:rewiring}.

{\bf Configuration model.} The configuration rewiring algorithm iterates through the edges in the graph and for each edge attempts to swap its target or source with the target or source of another edge in the graph such that the degrees of the involved nodes remain unchanged.
This is repeated until a given fraction of the total number of edges has been rewired.

{\bf Rewiring subgraphs.} Given an induced subgraph $H=(V_H, E_H)$ of a graph $G=(V_G, E_G)$ and a perturbation fraction $\eta \in [0, 1]$, the goal is to add or remove a total of $m_{\pm} = \floor{\eta |E_H|}$ edges between nodes in the subgraph only without disconnecting the subgraph. 
Thus, a random minimum spanning tree of the subgraph containing $|V_H| - 1$ edges is first extracted. 
Thereafter, $m_{-} = \min(0.5 m_{\pm}, |E_H| - (|V_H| - 1) )$ randomly chosen edges not in the tree are removed, and an additional $m_{+} = m_{\pm} - m_{-}$ randomly chosen edges, not present in the unperturbed subgraph, are added.
This procedure may fail for very dense subgraphs, specifically if the graph density exceeds 0.5, i.e.\ $|E_H| > 0.5 {|V_H| \choose 2}$,  as then a sufficient number of new edges cannot be added. 
However, no such subgraphs are part of the \textsc{Groletest} dataset and it is therefore ignored.

\subsection{Generative Models Hyperparameters}
\label{sec:app:generative_hyper}
\textbf{NetGAN.} We used the following default parameters for NetGAN recommended by the authors~\citet{bojchevski_netgan_2018} or provided in their official Github repository:  $3$ discriminator iterations are performed per generator iteration, learning rate is set to $0.0003$ with initial temperature of $5$ and temperature decay $0.99998$. Generator and discriminator weight matrix sizes were set to $32$ and $128$, $L2$ penalty values to $10^{-7}$ and $5\cdot10^{-5}$, layer counts to $40$ and $30$, respectively. Training was performed with an edge overlap $0.5$ stopping criterion, or with a validation score criterion stopped after $20$ evaluations without improvement. Stopping criteria were evaluated every $2000$ iterations.
Random walk length during training and generation was set to $16$. In total, $60,000$ random walks with Gumbel-Softmax temperature parameter $0.5$ were sampled to create an edge score matrix.

\textbf{CELL.} We used hyperparameters provided in the original publication~\cite{rendsburg_netgan_2020}. Logit space rank (H) was set to $9$, stopping criterion of $0.5$ edge overlap was invoked every $10$ iterations, otherwise the training was stopped after $300$ iterations. Learning rate was set to $0.1$ with weight decay $10^{-7}$.

\section{Additional Results}
\label{sec:app:results}

\begin{figure*}[t]
\vskip -0.1in
\newcommand\colscale{0.33}
\begin{subfigure}{\colscale\columnwidth}
    \begin{center}
    \centerline{\resizebox{\linewidth}{!}{
    \includegraphics{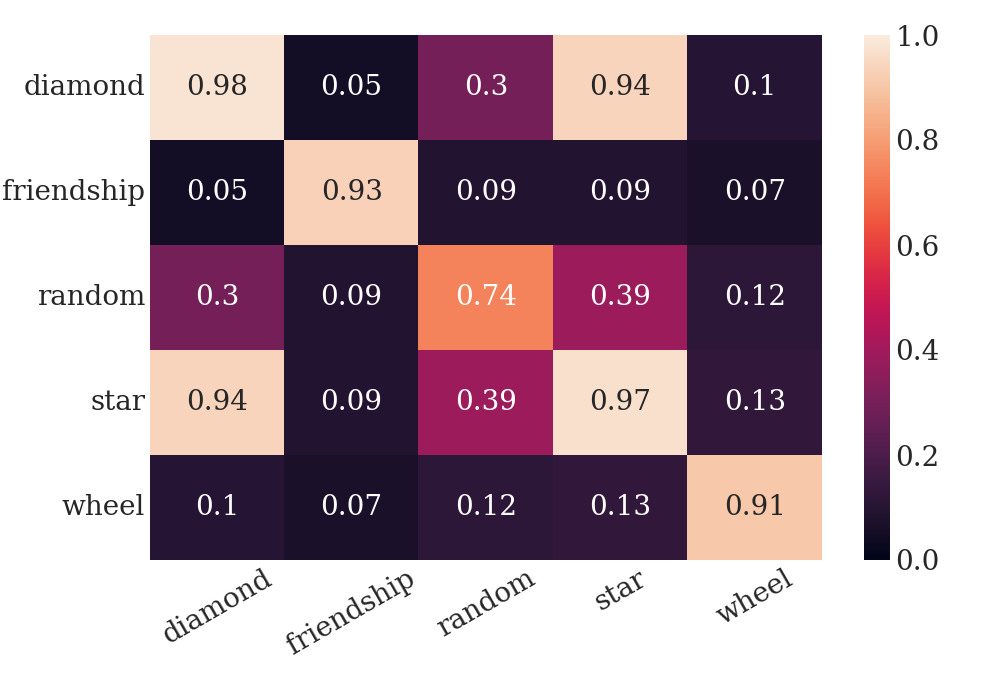}
    }}
    \caption{Manual features}
    \label{fig:synthetic_cm_egonet_count_weight}
    \end{center}
\end{subfigure}
~
\begin{subfigure}{\colscale\columnwidth}
    \begin{center}
    \centerline{\resizebox{\linewidth}{!}{
    \includegraphics{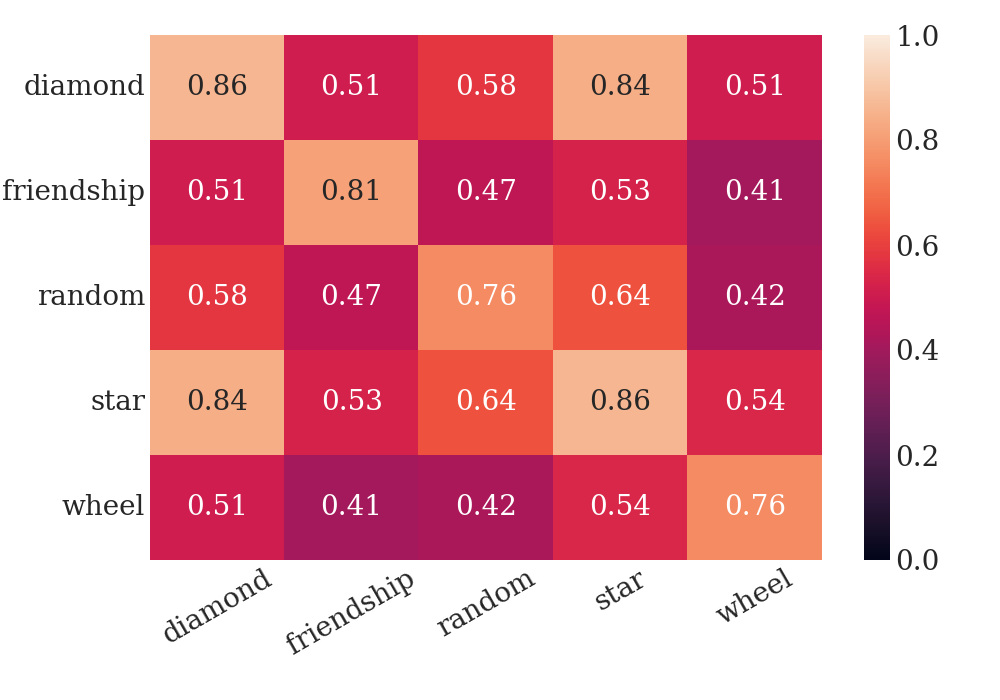}
    }}
    \caption{GCC}
    \label{fig:synthetic_cm_gcc_count_weight}
    \end{center}
\end{subfigure}
~
\begin{subfigure}{\colscale\columnwidth}
    \begin{center}
    \centerline{\resizebox{\linewidth}{!}{
    \includegraphics{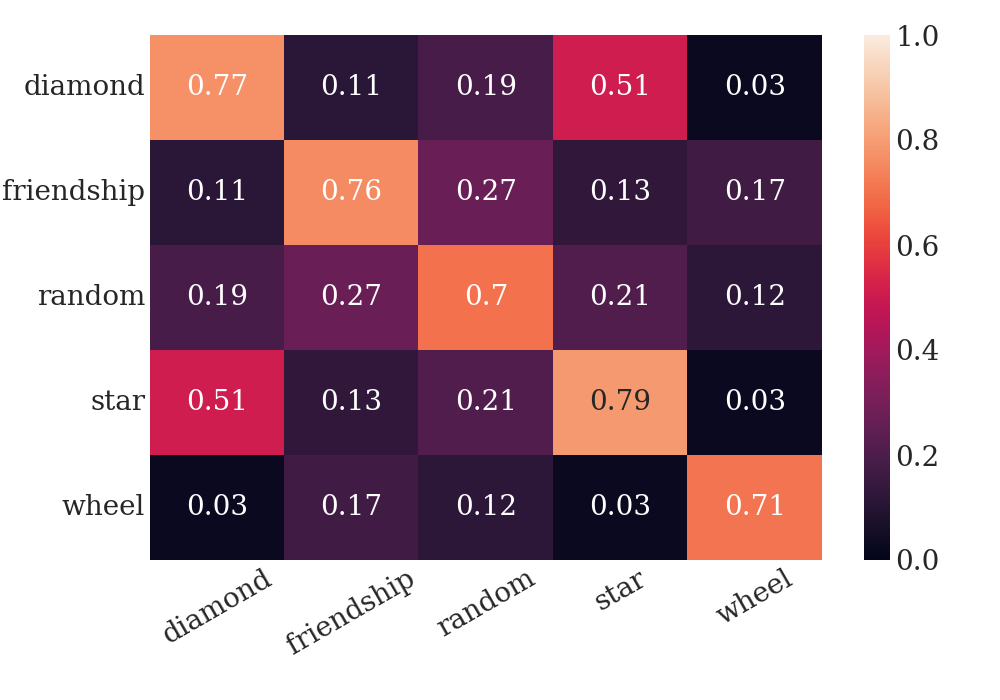}
    }}
    \caption{GraphWave}
    \label{fig:synthetic_cm_graphwave_count_weight}
    \end{center}
\end{subfigure}
\vskip -0.1in
\caption{Matrices with values representing the average $s_{\wDCA}$ scores obtained for various feature extractors $f$ and $G_1, G_2 \in \{G_h^\textit{cycle} | h = \textit{diamond, friendship, random, star, wheel} \}$. The result is averaged over 3 runs. Here, $s_{\wDCA}$ is calculated using uniform weights $w_i=1$ for all nodes in $G_1, G_2$.}
\label{fig:synthetic_confusion_matrices_count_weight}
\end{figure*}

\begin{figure}[t]
\newcommand\colscale{0.24}
\begin{subfigure}{\colscale\columnwidth}
    \begin{center}
    \Large{\centerline{\resizebox{\linewidth}{!}{
    \input{figures/tex/synthetic_change_main/change_main_graph_egonet_f1_h_count_weights_mcs2_fix}
    }}}
    \caption{Manual features}
    \label{fig:change_main_egonet_count_weight}
    \end{center}
\end{subfigure}
~
\begin{subfigure}{\colscale\columnwidth}
    \begin{center}
    \Large{\centerline{\resizebox{\linewidth}{!}{
    \input{figures/tex/synthetic_change_main/change_main_graph_gcc_f1_h_count_weights_mcs2_fix}
    }}}
    \caption{GCC}
    \label{fig:change_main_gcc_count_weight}
    \end{center}
\end{subfigure}
~
\begin{subfigure}{\colscale\columnwidth}
    \begin{center}
    \Large{\centerline{\resizebox{\linewidth}{!}{
    \input{figures/tex/synthetic_change_main/change_main_graph_graphwave_f1_h_count_weights_mcs2_fix}
    }}}
    \caption{GraphWave}
    \label{fig:change_main_graphwave_count_weight}
    \end{center}
\end{subfigure}
~
\begin{subfigure}{\colscale\columnwidth}
    \begin{center}
    \Large{\centerline{\resizebox{\linewidth}{!}{
    \input{figures/tex/synthetic_change_main/change_main_graph_gstats_rel_hmean_mcs2_fix}
    }}}
    \caption{Global statistics}
    \label{fig:change_main_gstats_count_weight}
    \end{center}
\end{subfigure}
\vskip -0.1in
\caption{Average $s_{\wDCA}$ score (\subref{fig:change_main_egonet_count_weight}, \subref{fig:change_main_gcc_count_weight}, \subref{fig:change_main_graphwave_count_weight}) obtained between two input graphs with fixed type of subgraphs $h$ having same main graph, i.e., $G_1 = G_h^{\textit{cycle}}, G_2 = G_h^{\textit{cycle}}$ (solid lines), and having different main graphs, i.e., $G_1 = G_h^{\textit{cycle}}, G_2 = G_h^{\textit{tree}}$ (dashed lines). The corresponding relative global statistics of input graphs $s_{\gstats}(G_1, G_2)$ are shown in \subref{fig:change_main_gstats_count_weight}. The result is averaged over 3 runs.
 Here, $s_{\wDCA}$ is calculated using uniform weights $w_i=1$ for all nodes in $G_1, G_2$.}
\label{fig:synthetic_varying_main_graph_count_weight}
\vskip -0.1in
\end{figure}

In this section, we report additional results supporting the discussions in the main paper. 
Specifically, we present the results of our experiments when using uniform weighting, $w_i =1$ for all $i$, in the $s_{\wDCA}$ score.

\subsection{Local Structural Similarity} \label{sec:app:struct_sim}
In Figures \ref{fig:synthetic_confusion_matrices_count_weight} and \ref{fig:synthetic_varying_main_graph_count_weight} we show $s_{\wDCA}$ scores corresponding to the local structural similarity experiments, as described in Section \ref{sec:struct_sim}, using uniform weighting.
Comparing Figures \ref{fig:synthetic_confusion_matrices_count_weight} and \ref{fig:synthetic_confusion_matrices}, displaying the $s_{\wDCA}$ scores obtained for \textsc{Groletest} graphs with \textit{cycle} main graph and varying subgraph types,
we observe that off-diagonal elements are generally higher when using the uniform weighting.
This is expected since the \textsc{Groletest} graphs in this experiment share the same main graph (\textit{cycle}) and the similarity of the local structure of main graph nodes is reflected in the $s_{\wDCA}$ scores.
We note that off-diagonal elements are particularly pronounced for GCC. We hypothesize that this could be an effect of the random walk subgraph sampling which results in more uniform distribution of node representations.

The opposite effect is observed for similarities calculated on \textsc{Groletest} graphs with different main graphs, shown in Figure \ref{fig:synthetic_varying_main_graph_count_weight}.
Since all nodes receive equal weight, the differences in the main graph structure are also accounted for which results in lower values for $s_{\wDCA}(G_h^{\textit{cycle}}, G_h^{\textit{tree}})$ compared to values obtained with weighting of central nodes reported in Figure \ref{fig:synthetic_varying_main_graph}.

\subsection{Gradual Structure Perturbation} \label{sec:app:edge_pert}

\begin{figure}[b]
\newcommand\colscale{0.33}
\begin{subfigure}{\colscale\columnwidth}
    \begin{center}
    \Large{\centerline{\resizebox{\linewidth}{!}{
    \input{figures/tex/synthetic_rewire/frac_rewired_diamond_counts_F1_qual_rel_score_mcs_2}
    }}}
    \caption{Diamond}
    \label{fig:rewire counts diamond}
    \end{center}
\end{subfigure}
\hfill
\begin{subfigure}{\colscale\columnwidth}
    \begin{center}
    \Large{\centerline{\resizebox{\linewidth}{!}{
    \input{figures/tex/synthetic_rewire/frac_rewired_friendship_counts_F1_qual_rel_score_mcs_2}}
    }}
    \caption{Friendship}
    \label{fig:rewire counts friendship}
    \end{center}
\end{subfigure}
\hfill
\begin{subfigure}{\colscale\columnwidth}
    \begin{center}
    \Large{\centerline{\resizebox{\linewidth}{!}{
    \input{figures/tex/synthetic_rewire/frac_rewired_samplededges_counts_F1_qual_rel_score_mcs_2}}
    }}
    \caption{Random}
    \label{fig:rewire counts samplededges}
    \end{center}
\end{subfigure}
\\
\begin{subfigure}{\colscale\columnwidth}
    \begin{center}
    \Large{\centerline{\resizebox{\linewidth}{!}{
    \input{figures/tex/synthetic_rewire/frac_rewired_star_counts_F1_qual_rel_score_mcs_2}
    }}}
    \caption{Star}
    \label{fig:rewire counts star}
    \end{center}
\end{subfigure}
\hfill
\begin{subfigure}{\colscale\columnwidth}
    \begin{center}
    \Large{\centerline{\resizebox{\linewidth}{!}{
    \input{figures/tex/synthetic_rewire/frac_rewired_wheel_counts_F1_qual_rel_score_mcs_2}
    }}}
    \caption{Wheel}
    \label{fig:rewire counts wheel}
    \end{center}
\end{subfigure}
\hfill
\begin{subfigure}{\colscale\columnwidth}
    \begin{center}
    \Large{\centerline{\resizebox{\linewidth}{!}{
    \input{figures/tex/synthetic_rewire/frac_rewired_samplededges_marked_F1_qual_rel_score_mcs_2}
    }}}
    \caption{Random with node importance weighting}
    \label{fig:rewire samplededges}
    \end{center}
\end{subfigure}
\vskip -0.1in
\caption{Rewiring of \textsc{Groletest} subgraphs via the addition and removal of edges. The x-axis specifies the fraction of changed edges. The y-axis shows $s_{\gstats}$ scores (in black) and $s_{\wDCA}$  scores for three different feature extractors (in color), averaged over $5$ different rewirings. For Figures \subref{fig:rewire counts diamond}-\subref{fig:rewire counts wheel}, $s_{\wDCA}$ is calculated using uniform weights $w_i=1$ for all nodes. 
In Figure \subref{fig:rewire samplededges}, the $s_{\wDCA}$ score is calculated using weighting of the central nodes where $w_i=0$ for all nodes except for the 20 central nodes in the subgraphs which receive $w_i=1$ (see also Figure \ref{fig:synthetic rewire}).}
\label{fig:synthetic counts rewire}
\vskip -0.2in
\end{figure}
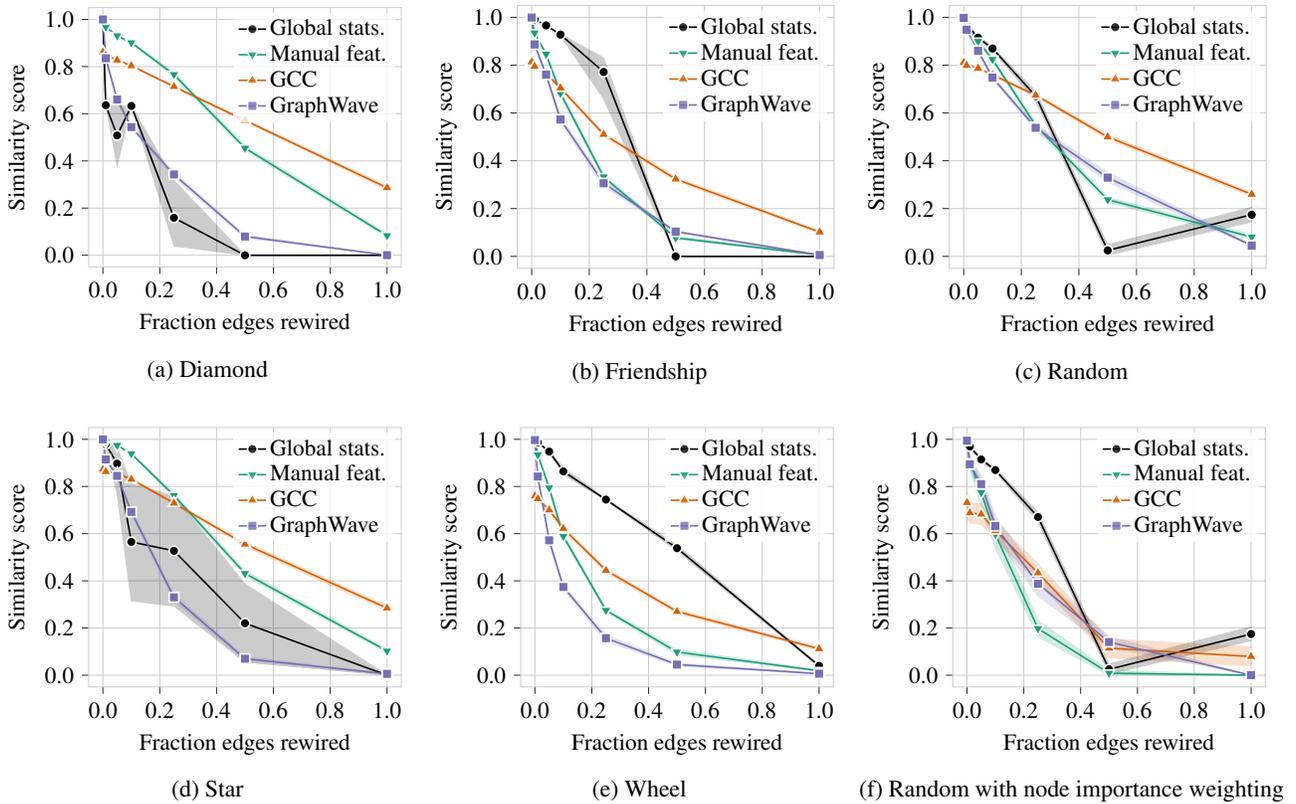

In Figures \ref{fig:rewire counts diamond}-\ref{fig:rewire counts wheel}, we report the results of the subgraph rewiring experiment, described in Section \ref{sec:rewiring}, using uniform weighting for the $s_{\wDCA}$ scores.
Additionally, the rewiring results using the \textit{random} subgraphs omitted from Figure \ref{fig:synthetic rewire} are shown in Figure \ref{fig:rewire samplededges}. 

When using uniform weighting, the $s_{\wDCA}$ scores increase for all feature extractors and for all rewiring fractions compared to the scores obtained with central node weighting shown in Figure \ref{fig:synthetic rewire}. 
This is expected since only the edges of the subgraphs are perturbed meaning that the graphs share the same main graph for which the local structural similarity is preserved.

\subsection{Evaluation of Graph Generative Models} \label{sec:app:gen_models}
In Table~\ref{table:app_real_netgan_val_vs_eo:dca}, we report the global graph statistics and $s_{wDCA}$ scores obtained on real-world datasets for NetGAN models trained with {Val} and {EO} stopping criteria. The results are on par with prior conclusions about higher efficiency of EO criterion in generating new graphs. We highlight that $s_{wDCA}$ scores provide a more clear comparison between two models in cases where all global statistics are either both similar or both different from the training values (see for example \textsc{Cora-ML}).

In Table~\ref{table:app_synthetic_gen_models_random_star:dca}, we report the global statistics and $s_{\wDCA}$ obtained on \textsc{Groletest} for subgraphs $h = \textit{random, star}$. Similar to the results obtained for $h = \textit{diamond, friendship, wheel}$ in Table~\ref{table:synthetic_gen_models:dca}, we observe that the models struggle to produce graphs that would well reflect all the global statistics with CELL being more efficient in replicating the egocentric networks among the two.
\begin{table*}[b]
\centering
\vskip -0.3in
\caption{
Global graph statistics of the real-world training graphs $G_1$ and $s_{\DCA}$ scores obtained on representations extracted from the considered feature extractors $f$ of $G_1$ and graphs $G_2$ generated by NetGAN trained with VAL and EO stopping criteria. The scores are averaged over 3 independently generated graphs with the same training split.}
\vspace{2ex}
\begin{adjustbox}{width=\linewidth,center}
\input{tables/appendix_real_val_vs_eo_netgan}
\end{adjustbox}
\label{table:app_real_netgan_val_vs_eo:dca}
\end{table*}

\begin{table*}[b]
\centering
\vskip -0.3in
\caption{
Global graph statistics of \textsc{Groletest}  \textit{random} and \textit{star} training graphs $G_1$ and $s_{\wDCA}$ scores obtained on representations extracted from the considered feature extractors $f$ of $G_1$ and graphs $G_2$ generated by NetGAN and CELL (50\% EO stopping criterion). The scores are averaged over 3 independently generated graphs with the same training split. }
\vspace{2ex}
\begin{adjustbox}{width=0.8\linewidth,center}
\input{tables/appendix_synthetic_random_star_gm}
\end{adjustbox}
\label{table:app_synthetic_gen_models_random_star:dca}
\end{table*}

%% file: figures/tex/synthetic_change_main/change_main_graph_egonet_f1_h_count_weights_mcs2_fix.tex
\begin{tikzpicture}

\definecolor{color0}{rgb}{0.105882352941176,0.619607843137255,0.466666666666667}
\definecolor{color1}{rgb}{0.0705882352941176,0.431372549019608,0.325490196078431}

\begin{axis}[
axis line style={white!80!black},
legend cell align={left},
legend style={
  fill opacity=0.8,
  draw opacity=1,
  text opacity=1,
  at={(0.03,0.03)},
  anchor=south west,
  draw=none
},
tick align=outside,
tick pos=left,
x grid style={white!80!black},
xmajorgrids,
xmin=-0.2, xmax=4.2,
xtick style={color=white!15!black},
xtick={0,1,2,3,4},
xtick={0,1,2,3,4},
xtick={0,1,2,3,4},
xticklabel style={rotate=30.0},
xticklabels={diamond,friendship,random,star,wheel},
xticklabels={diamond,friendship,random,star,wheel},
xticklabels={diamond,friendship,random,star,wheel},
y grid style={white!80!black},
ylabel={Similarity score},
ymajorgrids,
ymin=0, ymax=1,
ytick style={color=white!15!black},
ytick={0,0.2,0.4,0.6,0.8,1},
yticklabels={0.0,0.2,0.4,0.6,0.8,1.0}
]
\path [draw=color0, line width=1pt]
(axis cs:0,0.972)
--(axis cs:0,0.984);

\path [draw=color0, line width=1pt]
(axis cs:1,0.923)
--(axis cs:1,0.945);

\path [draw=color0, line width=1pt]
(axis cs:2,0.709)
--(axis cs:2,0.763);

\path [draw=color0, line width=1pt]
(axis cs:3,0.973)
--(axis cs:3,0.975);

\path [draw=color0, line width=1pt]
(axis cs:4,0.896)
--(axis cs:4,0.914);

\path [draw=color1, line width=1pt]
(axis cs:0,0.374)
--(axis cs:0,0.46);

\path [draw=color1, line width=1pt]
(axis cs:1,0.382)
--(axis cs:1,0.412);

\path [draw=color1, line width=1pt]
(axis cs:2,0.388)
--(axis cs:2,0.412);

\path [draw=color1, line width=1pt]
(axis cs:3,0.464)
--(axis cs:3,0.492);

\path [draw=color1, line width=1pt]
(axis cs:4,0.397)
--(axis cs:4,0.431);

\addplot [line width=1pt, color0, mark=triangle*, mark size=3, mark options={solid,rotate=180}]
table {%
0 0.978
1 0.934
2 0.736
3 0.974
4 0.905
};
\addlegendentry{cycle-cycle}
\addplot [line width=1pt, color1, dashed, mark=diamond*, mark size=3, mark options={solid}]
table {%
0 0.417
1 0.397
2 0.4
3 0.478
4 0.414
};
\addlegendentry{cycle-tree}
\end{axis}

\end{tikzpicture}

%% file: figures/tex/synthetic_change_main/change_main_graph_gcc_f1_h_count_weights_mcs2_fix.tex
\begin{tikzpicture}

\definecolor{color0}{rgb}{0.850980392156863,0.372549019607843,0.00784313725490196}
\definecolor{color1}{rgb}{0.592156862745098,0.258823529411765,0.00392156862745098}

\begin{axis}[
axis line style={white!80!black},
legend cell align={left},
legend style={
  fill opacity=0.8,
  draw opacity=1,
  text opacity=1,
  at={(0.03,0.03)},
  anchor=south west,
  draw=none
},
tick align=outside,
tick pos=left,
x grid style={white!80!black},
xmajorgrids,
xmin=-0.2, xmax=4.2,
xtick style={color=white!15!black},
xtick={0,1,2,3,4},
xtick={0,1,2,3,4},
xtick={0,1,2,3,4},
xticklabel style={rotate=30.0},
xticklabels={diamond,friendship,random,star,wheel},
xticklabels={diamond,friendship,random,star,wheel},
xticklabels={diamond,friendship,random,star,wheel},
y grid style={white!80!black},
ylabel={Similarity score},
ymajorgrids,
ymin=0, ymax=1,
ytick style={color=white!15!black},
ytick={0,0.2,0.4,0.6,0.8,1},
yticklabels={0.0,0.2,0.4,0.6,0.8,1.0}
]
\path [draw=color0, line width=1pt]
(axis cs:0,0.859)
--(axis cs:0,0.867);

\path [draw=color0, line width=1pt]
(axis cs:1,0.789)
--(axis cs:1,0.823);

\path [draw=color0, line width=1pt]
(axis cs:2,0.736)
--(axis cs:2,0.782);

\path [draw=color0, line width=1pt]
(axis cs:3,0.852)
--(axis cs:3,0.868);

\path [draw=color0, line width=1pt]
(axis cs:4,0.753)
--(axis cs:4,0.757);

\path [draw=color1, line width=1pt]
(axis cs:0,0.515)
--(axis cs:0,0.531);

\path [draw=color1, line width=1pt]
(axis cs:1,0.368)
--(axis cs:1,0.38);

\path [draw=color1, line width=1pt]
(axis cs:2,0.397)
--(axis cs:2,0.399);

\path [draw=color1, line width=1pt]
(axis cs:3,0.584)
--(axis cs:3,0.588);

\path [draw=color1, line width=1pt]
(axis cs:4,0.294)
--(axis cs:4,0.346);

\addplot [line width=1pt, color0, mark=triangle*, mark size=3, mark options={solid}]
table {%
0 0.863
1 0.806
2 0.759
3 0.86
4 0.755
};
\addlegendentry{cycle-cycle}
\addplot [line width=1pt, color1, dashed, mark=diamond*, mark size=3, mark options={solid}]
table {%
0 0.523
1 0.374
2 0.398
3 0.586
4 0.32
};
\addlegendentry{cycle-tree}
\end{axis}

\end{tikzpicture}

%% file: figures/tex/synthetic_change_main/change_main_graph_graphwave_f1_h_count_weights_mcs2_fix.tex
\begin{tikzpicture}

\definecolor{color0}{rgb}{0.458823529411765,0.43921568627451,0.701960784313725}
\definecolor{color1}{rgb}{0.317647058823529,0.305882352941176,0.490196078431373}

\begin{axis}[
axis line style={white!80!black},
legend cell align={left},
legend style={
  fill opacity=0.8,
  draw opacity=1,
  text opacity=1,
  at={(0.03,0.97)},
  anchor=north west,
  draw=none
},
tick align=outside,
tick pos=left,
x grid style={white!80!black},
xmajorgrids,
xmin=-0.2, xmax=4.2,
xtick style={color=white!15!black},
xtick={0,1,2,3,4},
xtick={0,1,2,3,4},
xtick={0,1,2,3,4},
xticklabel style={rotate=30.0},
xticklabels={diamond,friendship,random,star,wheel},
xticklabels={diamond,friendship,random,star,wheel},
xticklabels={diamond,friendship,random,star,wheel},
y grid style={white!80!black},
ylabel={Similarity score},
ymajorgrids,
ymin=0, ymax=1,
ytick style={color=white!15!black},
ytick={0,0.2,0.4,0.6,0.8,1},
yticklabels={0.0,0.2,0.4,0.6,0.8,1.0}
]
\path [draw=color0, line width=1pt]
(axis cs:0,0.766)
--(axis cs:0,0.78);

\path [draw=color0, line width=1pt]
(axis cs:1,0.755)
--(axis cs:1,0.769);

\path [draw=color0, line width=1pt]
(axis cs:2,0.683)
--(axis cs:2,0.719);

\path [draw=color0, line width=1pt]
(axis cs:3,0.766)
--(axis cs:3,0.806);

\path [draw=color0, line width=1pt]
(axis cs:4,0.689)
--(axis cs:4,0.727);

\path [draw=color1, line width=1pt]
(axis cs:0,0.2)
--(axis cs:0,0.232);

\path [draw=color1, line width=1pt]
(axis cs:1,0.242)
--(axis cs:1,0.328);

\path [draw=color1, line width=1pt]
(axis cs:2,0.343)
--(axis cs:2,0.391);

\path [draw=color1, line width=1pt]
(axis cs:3,0.276)
--(axis cs:3,0.3);

\path [draw=color1, line width=1pt]
(axis cs:4,0.187)
--(axis cs:4,0.261);

\addplot [line width=1pt, color0, mark=square*, mark size=3, mark options={solid}]
table {%
0 0.773
1 0.762
2 0.701
3 0.786
4 0.708
};
\addlegendentry{cycle-cycle}
\addplot [line width=1pt, color1, dashed, mark=diamond*, mark size=3, mark options={solid}]
table {%
0 0.216
1 0.285
2 0.367
3 0.288
4 0.224
};
\addlegendentry{cycle-tree}
\end{axis}

\end{tikzpicture}

%% file: tables/appendix_real_val_vs_eo_netgan.tex
\begin{small}
\begin{sc}
\begin{tabular}{@{}r|ccc|ccc|ccc@{}}
\toprule
  & \multicolumn{3}{c|}{Cora-ML}   & \multicolumn{3}{c|}{Citeseer} & \multicolumn{3}{c}{PolBlogs}      \\
\midrule
 &  \thead{Ground \\ truth} & \thead{NetGAN \\ VAL} & \thead{NetGAN \\ EO} & \thead{Ground \\ truth} & \thead{NetGAN \\ VAL} & \thead{NetGAN \\ EO}  & \thead{Ground \\ truth} & \thead{NetGAN \\ VAL} & \thead{NetGAN \\ EO}  \\
\midrule
Max degree &  $238$  &  $256.00 \pm 10.82$  &  $216.00 \pm 11.14$  &  $76$  &  $70.33 \pm 12.22$  &  $88.00 \pm 1.73$  &  $303$  &  $260.67 \pm 5.13$  &  $279.33 \pm 14.57$  \\ 
Assortativity &  $-0.08$  &  $-0.04 \pm 0.00$  &  $-0.08 \pm 0.00$  &  $-0.19$  &  $-0.08 \pm 0.00$  &  $-0.16 \pm 0.00$  &  $-0.22$  &  $-0.25 \pm 0.01$  &  $-0.25 \pm 0.01$  \\ 
Triangle count &  $2,802$  &  $659.00 \pm 28.36$  &  $1,772.33 \pm 17.39$  &  $304$  &  $74.67 \pm 11.59$  &  $195.33 \pm 5.86$  &  $61,108$  &  $33,105.33 \pm 360.05$  &  $36,294.33 \pm 348.62$  \\ 
Square count &  $14,268$  &  $2,461 \pm 7.81$  &  $6,741.33 \pm 229.38$  &  $1,441$  &  $166.00 \pm 17.44$  &  $371.33 \pm 35.22$  &  $2,654,319$  &  $1,260,882.00 \pm 7,415.16$  &  $1,350,075.67 \pm 17,073.43$  \\ 
Power law exp. &  $1.86$  &  $1.78 \pm 0.00$  &  $1.81 \pm 0.00$  &  $2.45$  &  $2.28 \pm 0.00$  &  $2.32 \pm 0.01$  &  $1.44$  &  $1.40 \pm 0.00$  &  $1.41 \pm 0.00$  \\ 
Clustering coeff. &  $0.08$  &  $0.02 \pm 0.00$  &  $0.06 \pm 0.00$  &  $0.04$  &  $0.02 \pm 0.00$  &  $0.03 \pm 0.00$  &  $0.19$  &  $0.12 \pm 0.00$  &  $0.13 \pm 0.00$  \\ 
Charc. path len. &  $5.63$  &  $4.80 \pm 0.02$  &  $5.23 \pm 0.03$  &  $8.02$  &  $6.00 \pm 0.09$  &  $6.49 \pm 0.20$  &  $2.82$  &  $2.64 \pm 0.01$  &  $2.67 \pm 0.00$  \\ 

\midrule
$s_{\wDCA}$ -- Manual &  $1.00$  &  $0.02 \pm 0.01$  &  $0.11 \pm 0.01$  &  $1.00$  &  $0.13 \pm 0.01$  &  $0.22 \pm 0.01$  &  $1.00$  &  $0.00 \pm 0.00$  &  $0.00 \pm 0.00$  \\ 
$s_{\wDCA}$ -- GCC &  $0.73$  &  $0.23 \pm 0.02$  &  $0.51 \pm 0.00$  &  $0.75$  &  $0.35 \pm 0.02$  &  $0.50 \pm 0.01$  &  $0.66$  &  $0.34 \pm 0.03$  &  $0.38 \pm 0.02$  \\ 
$s_{\wDCA}$ -- Graphwave &  $1.00$  &  $0.18 \pm 0.01$  &  $0.52 \pm 0.01$  &  $1.00$  &  $0.32 \pm 0.01$  &  $0.53 \pm 0.03$  &  $1.00$  &  $0.43 \pm 0.02$  &  $0.50 \pm 0.01$  \\
\bottomrule
\end{tabular}
\end{sc}
\end{small}

%% file: tables/appendix_synthetic_random_star_gm.tex
\begin{small}
\begin{sc}
\begin{tabular}{@{}r|ccc|ccc@{}}
\toprule
  & \multicolumn{3}{c|}{random}   & \multicolumn{3}{c}{star}      \\
\midrule
 &  \thead{Ground \\ truth} & \thead{NetGAN} & \thead{CELL} & \thead{Ground \\ truth} & NetGAN & CELL   \\
\midrule
Max degree &  $6$  &  $7.33 \pm 1.53$  &  $8.00 \pm 1.0$  &  $5$  &  $6.67 \pm 0.58$  &  $6.33 \pm 0.58$  \\ 
Assortativity &  $-0.04$  &  $-0.14 \pm 0.02$  &  $-0.02 \pm 0.03$  &  $0.08$  &  $-0.14 \pm 0.03$  &  $0.10 \pm 0.01$  \\ 
Triangle count &  $0$  &  $103.67 \pm 9.45$  &  $0.00 \pm 0.00$  &  $0$  &  $107.33 \pm 9.29$  &  $0.67 \pm 0.58$  \\ 
Square count &  $0$  &  $46.00 \pm 3.61$  &  $9.33 \pm 0.58$  &  $0$  &  $46.67 \pm 2.08$  &  $12.67 \pm 2.08$  \\ 
Power law exp. &  $2.59$  &  $2.73 \pm 0.01$  &  $2.70 \pm 0.00$  &  $2.50$  &  $2.71 \pm 0.01$  &  $2.65 \pm 0.01$  \\ 
Clustering coeff. &  $0.00$  &  $0.11 \pm 0.01$  &  $0.00 \pm 0.00$  &  $0.00$  &  $0.12 \pm 0.01$  &  $0.00 \pm 0.00$  \\ 
Charc. path len. &  $83.73$  &  $14.97 \pm 0.75$  &  $17.85 \pm 0.81$  &  $79.50$  &  $10.98 \pm 3.16$  &  $18.84 \pm 0.37$  \\ 
\midrule
$s_{\wDCA}$ -- Manual &  $1.00 \pm 0.00$  &  $0.41 \pm 0.03$  &  $0.73 \pm 0.01$  &  $1.00 \pm 0.00$  &  $0.32 \pm 0.03$  &  $0.66 \pm 0.02$  \\ 
$s_{\wDCA}$ --  GCC &  $0.76 \pm 0.02$  &  $0.49 \pm 0.02$  &  $0.70 \pm 0.00$  &  $0.80 \pm 0.02$  &  $0.33 \pm 0.03$  &  $0.62 \pm 0.02$  \\ 
$s_{\wDCA}$ --  Graphwave &  $1.00 \pm 0.00$  &  $0.32 \pm 0.00$  &  $0.47 \pm 0.02$  &  $1.00 \pm 0.00$  &  $0.11 \pm 0.01$  &  $0.26 \pm 0.02$  \\ 
\bottomrule
\end{tabular}
\end{sc}
\end{small}

%% file: figures/tex/synthetic_rewire/frac_rewired_diamond_counts_F1_qual_rel_score_mcs_2.tex
\begin{tikzpicture}

\definecolor{color0}{rgb}{0.105882352941176,0.619607843137255,0.466666666666667}
\definecolor{color1}{rgb}{0.850980392156863,0.372549019607843,0.00784313725490196}
\definecolor{color2}{rgb}{0.458823529411765,0.43921568627451,0.701960784313725}

\begin{axis}[
axis line style={white!80!black},
legend cell align={left},
legend style={fill opacity=0.8, draw opacity=1, text opacity=1, draw=none},
tick align=outside,
tick pos=left,
x grid style={white!80!black},
xlabel={Fraction edges rewired},
xmajorgrids,
xmin=-0.05, xmax=1.05,
xtick style={color=white!15!black},
xtick={-0.2,0,0.2,0.4,0.6,0.8,1,1.2},
xticklabels={−0.2,0.0,0.2,0.4,0.6,0.8,1.0,1.2},
y grid style={white!80!black},
ylabel={Similarity score},
ymajorgrids,
ymin=-0.05, ymax=1.05,
ytick style={color=white!15!black},
ytick={-0.2,0,0.2,0.4,0.6,0.8,1,1.2},
yticklabels={−0.2,0.0,0.2,0.4,0.6,0.8,1.0,1.2}
]
\path [draw=black, fill=black, opacity=0.2]
(axis cs:0,1)
--(axis cs:0,1)
--(axis cs:0.01,0.634917176039136)
--(axis cs:0.05,0.379812819258392)
--(axis cs:0.1,0.631130331665339)
--(axis cs:0.25,0.0395746438507714)
--(axis cs:0.5,0)
--(axis cs:1,0)
--(axis cs:1,0)
--(axis cs:1,0)
--(axis cs:0.5,0)
--(axis cs:0.25,0.3174501379442)
--(axis cs:0.1,0.634938261877516)
--(axis cs:0.05,0.633215635684729)
--(axis cs:0.01,0.637771347687718)
--(axis cs:0,1)
--cycle;

\path [draw=color0, fill=color0, opacity=0.2]
(axis cs:0,1)
--(axis cs:0,1)
--(axis cs:0.01,0.96675203455291)
--(axis cs:0.05,0.928448799831522)
--(axis cs:0.1,0.897919635091546)
--(axis cs:0.25,0.7552353804992)
--(axis cs:0.5,0.445583144406938)
--(axis cs:1,0.0754667632209763)
--(axis cs:1,0.0919273334052237)
--(axis cs:1,0.0919273334052237)
--(axis cs:0.5,0.46386504544455)
--(axis cs:0.25,0.777576386515288)
--(axis cs:0.1,0.904548396305324)
--(axis cs:0.05,0.933017597434835)
--(axis cs:0.01,0.969460251784619)
--(axis cs:0,1)
--cycle;

\path [draw=color1, fill=color1, opacity=0.2]
(axis cs:0,0.865356551257788)
--(axis cs:0,0.853936751935299)
--(axis cs:0.01,0.843061717410825)
--(axis cs:0.05,0.822465808987263)
--(axis cs:0.1,0.799621563154453)
--(axis cs:0.25,0.711664406333127)
--(axis cs:0.5,0.566293996280961)
--(axis cs:1,0.278616651654487)
--(axis cs:1,0.294197429592186)
--(axis cs:1,0.294197429592186)
--(axis cs:0.5,0.576952845143296)
--(axis cs:0.25,0.719377269707374)
--(axis cs:0.1,0.807203671576845)
--(axis cs:0.05,0.832115913089113)
--(axis cs:0.01,0.850617486289317)
--(axis cs:0,0.865356551257788)
--cycle;

\path [draw=color2, fill=color2, opacity=0.2]
(axis cs:0,1)
--(axis cs:0,0.999629355081545)
--(axis cs:0.01,0.828252807009784)
--(axis cs:0.05,0.650452313986162)
--(axis cs:0.1,0.533337428200064)
--(axis cs:0.25,0.331531481293667)
--(axis cs:0.5,0.072419671266781)
--(axis cs:1,0.000152148575533121)
--(axis cs:1,0.000963335526377594)
--(axis cs:1,0.000963335526377594)
--(axis cs:0.5,0.0862633353385973)
--(axis cs:0.25,0.353778057595964)
--(axis cs:0.1,0.553225971972271)
--(axis cs:0.05,0.669430772337772)
--(axis cs:0.01,0.841542674109485)
--(axis cs:0,1)
--cycle;

\addplot [line width=1pt, black, mark=*, mark size=3, mark options={solid,draw=white}]
table {%
0 1
0.01 0.636260571940902
0.05 0.508188089368127
0.1 0.632957289490248
0.25 0.158946080369185
0.5 0
1 0
};
\addlegendentry{Global stats.}
\addplot [line width=1pt, color0, mark=triangle*, mark size=4, mark options={solid,rotate=180,draw=white}]
table {%
0 1
0.01 0.968172791558383
0.05 0.930704976840305
0.1 0.901186879772098
0.25 0.766301712143074
0.5 0.454445180978149
1 0.0831455867464761
};
\addlegendentry{Manual feat.}
\addplot [line width=1pt, color1, mark=triangle*, mark size=4, mark options={solid,draw=white}]
table {%
0 0.861209772963663
0.01 0.846729332363764
0.05 0.827242924423598
0.1 0.803337647723791
0.25 0.715256929094734
0.5 0.571850368428576
1 0.286182330207307
};
\addlegendentry{GCC}
\addplot [line width=1pt, color2, mark=square*, mark size=3, mark options={solid,draw=white}]
table {%
0 0.999752903387697
0.01 0.835030187067974
0.05 0.660256759374751
0.1 0.54332525070218
0.25 0.342531222667465
0.5 0.0796388496631418
1 0.000517271489061784
};
\addlegendentry{GraphWave}
\end{axis}

\end{tikzpicture}

%% file: figures/tex/synthetic_rewire/frac_rewired_friendship_counts_F1_qual_rel_score_mcs_2.tex
\begin{tikzpicture}

\definecolor{color0}{rgb}{0.105882352941176,0.619607843137255,0.466666666666667}
\definecolor{color1}{rgb}{0.850980392156863,0.372549019607843,0.00784313725490196}
\definecolor{color2}{rgb}{0.458823529411765,0.43921568627451,0.701960784313725}

\begin{axis}[
axis line style={white!80!black},
legend cell align={left},
legend style={fill opacity=0.8, draw opacity=1, text opacity=1, draw=none},
tick align=outside,
tick pos=left,
x grid style={white!80!black},
xlabel={Fraction edges rewired},
xmajorgrids,
xmin=-0.05, xmax=1.05,
xtick style={color=white!15!black},
xtick={-0.2,0,0.2,0.4,0.6,0.8,1,1.2},
xticklabels={−0.2,0.0,0.2,0.4,0.6,0.8,1.0,1.2},
y grid style={white!80!black},
ylabel={Similarity score},
ymajorgrids,
ymin=-0.05, ymax=1.05,
ytick style={color=white!15!black},
ytick={-0.2,0,0.2,0.4,0.6,0.8,1,1.2},
yticklabels={−0.2,0.0,0.2,0.4,0.6,0.8,1.0,1.2}
]
\path [draw=black, fill=black, opacity=0.2]
(axis cs:0,1)
--(axis cs:0,1)
--(axis cs:0.01,0.990579561646301)
--(axis cs:0.05,0.963841870939877)
--(axis cs:0.1,0.924164701627066)
--(axis cs:0.25,0.658053388363704)
--(axis cs:0.5,0)
--(axis cs:1,0)
--(axis cs:1,0)
--(axis cs:1,0)
--(axis cs:0.5,0)
--(axis cs:0.25,0.829928719253325)
--(axis cs:0.1,0.930805711855663)
--(axis cs:0.05,0.967700390609681)
--(axis cs:0.01,0.99152497141665)
--(axis cs:0,1)
--cycle;

\path [draw=color0, fill=color0, opacity=0.2]
(axis cs:0,1)
--(axis cs:0,1)
--(axis cs:0.01,0.930845109160376)
--(axis cs:0.05,0.838080632560312)
--(axis cs:0.1,0.669092440883689)
--(axis cs:0.25,0.323635448777525)
--(axis cs:0.5,0.0715823873944458)
--(axis cs:1,0.00471410537074296)
--(axis cs:1,0.00758175630015974)
--(axis cs:1,0.00758175630015974)
--(axis cs:0.5,0.0844663980645491)
--(axis cs:0.25,0.343538425771055)
--(axis cs:0.1,0.695529784973413)
--(axis cs:0.05,0.854509953776928)
--(axis cs:0.01,0.938161269723857)
--(axis cs:0,1)
--cycle;

\path [draw=color1, fill=color1, opacity=0.2]
(axis cs:0,0.813753292517228)
--(axis cs:0,0.812435376826348)
--(axis cs:0.01,0.791458220295486)
--(axis cs:0.05,0.759654946983557)
--(axis cs:0.1,0.698689821675424)
--(axis cs:0.25,0.501906654495973)
--(axis cs:0.5,0.314950321005402)
--(axis cs:1,0.0961998588164272)
--(axis cs:1,0.108348136893012)
--(axis cs:1,0.108348136893012)
--(axis cs:0.5,0.330914443151482)
--(axis cs:0.25,0.518069630646072)
--(axis cs:0.1,0.710094917321814)
--(axis cs:0.05,0.770646962138078)
--(axis cs:0.01,0.798803909774589)
--(axis cs:0,0.813753292517228)
--cycle;

\path [draw=color2, fill=color2, opacity=0.2]
(axis cs:0,1)
--(axis cs:0,0.997393894259083)
--(axis cs:0.01,0.879897887673165)
--(axis cs:0.05,0.748920528122157)
--(axis cs:0.1,0.55870393795974)
--(axis cs:0.25,0.292026645251971)
--(axis cs:0.5,0.0993094533714479)
--(axis cs:1,0.0042326630385186)
--(axis cs:1,0.00742935780138296)
--(axis cs:1,0.00742935780138296)
--(axis cs:0.5,0.108066255118126)
--(axis cs:0.25,0.318345149400792)
--(axis cs:0.1,0.584909287225895)
--(axis cs:0.05,0.770851446885422)
--(axis cs:0.01,0.892821242935798)
--(axis cs:0,1)
--cycle;

\addplot [line width=1pt, black, mark=*, mark size=3, mark options={solid,draw=white}]
table {%
0 1
0.01 0.991073521500448
0.05 0.96600000648364
0.1 0.927591009935351
0.25 0.771339257316674
0.5 0
1 0
};
\addlegendentry{Global stats.}
\addplot [line width=1pt, color0, mark=triangle*, mark size=4, mark options={solid,rotate=180,draw=white}]
table {%
0 1
0.01 0.934521702088604
0.05 0.846145168147045
0.1 0.682079161823359
0.25 0.332582539307842
0.5 0.0777005444986635
1 0.00613954067114054
};
\addlegendentry{Manual feat.}
\addplot [line width=1pt, color1, mark=triangle*, mark size=4, mark options={solid,draw=white}]
table {%
0 0.813091945138669
0.01 0.795059306383016
0.05 0.764891326413005
0.1 0.704304105360205
0.25 0.5103005733018
0.5 0.32336439103556
1 0.10179931564362
};
\addlegendentry{GCC}
\addplot [line width=1pt, color2, mark=square*, mark size=3, mark options={solid,draw=white}]
table {%
0 0.999131298086361
0.01 0.886609555769418
0.05 0.760340722389563
0.1 0.572640378677474
0.25 0.305817292384937
0.5 0.103667940372759
1 0.00575969519257242
};
\addlegendentry{GraphWave}
\end{axis}

\end{tikzpicture}

%% file: figures/tex/synthetic_rewire/frac_rewired_samplededges_counts_F1_qual_rel_score_mcs_2.tex
\begin{tikzpicture}

\definecolor{color0}{rgb}{0.105882352941176,0.619607843137255,0.466666666666667}
\definecolor{color1}{rgb}{0.850980392156863,0.372549019607843,0.00784313725490196}
\definecolor{color2}{rgb}{0.458823529411765,0.43921568627451,0.701960784313725}

\begin{axis}[
axis line style={white!80!black},
legend cell align={left},
legend style={fill opacity=0.8, draw opacity=1, text opacity=1, draw=none},
tick align=outside,
tick pos=left,
x grid style={white!80!black},
xlabel={Fraction edges rewired},
xmajorgrids,
xmin=-0.05, xmax=1.05,
xtick style={color=white!15!black},
xtick={-0.2,0,0.2,0.4,0.6,0.8,1,1.2},
xticklabels={−0.2,0.0,0.2,0.4,0.6,0.8,1.0,1.2},
y grid style={white!80!black},
ylabel={Similarity score},
ymajorgrids,
ymin=-0.05, ymax=1.05,
ytick style={color=white!15!black},
ytick={-0.2,0,0.2,0.4,0.6,0.8,1,1.2},
yticklabels={−0.2,0.0,0.2,0.4,0.6,0.8,1.0,1.2}
]
\path [draw=black, fill=black, opacity=0.2]
(axis cs:0,1)
--(axis cs:0,1)
--(axis cs:0.01,0.964817112427004)
--(axis cs:0.05,0.903560166636392)
--(axis cs:0.1,0.862583563960745)
--(axis cs:0.25,0.650579224884621)
--(axis cs:0.5,0.00537711913871193)
--(axis cs:1,0.144299316088724)
--(axis cs:1,0.205297668316049)
--(axis cs:1,0.205297668316049)
--(axis cs:0.5,0.0488946075341628)
--(axis cs:0.25,0.69055025262354)
--(axis cs:0.1,0.876098316816707)
--(axis cs:0.05,0.924526798541603)
--(axis cs:0.01,0.974105716846484)
--(axis cs:0,1)
--cycle;

\path [draw=color0, fill=color0, opacity=0.2]
(axis cs:0,1)
--(axis cs:0,1)
--(axis cs:0.01,0.959031410146174)
--(axis cs:0.05,0.889358503103173)
--(axis cs:0.1,0.816907564592109)
--(axis cs:0.25,0.531677384192712)
--(axis cs:0.5,0.226253635935745)
--(axis cs:1,0.0712454844880665)
--(axis cs:1,0.0928293300857368)
--(axis cs:1,0.0928293300857368)
--(axis cs:0.5,0.247428340206031)
--(axis cs:0.25,0.562753116017247)
--(axis cs:0.1,0.830532126209178)
--(axis cs:0.05,0.9087176752206)
--(axis cs:0.01,0.968663084263096)
--(axis cs:0,1)
--cycle;

\path [draw=color1, fill=color1, opacity=0.2]
(axis cs:0,0.82346826302162)
--(axis cs:0,0.803936573000983)
--(axis cs:0.01,0.793566626917544)
--(axis cs:0.05,0.782199978674782)
--(axis cs:0.1,0.754234591251452)
--(axis cs:0.25,0.667954635179978)
--(axis cs:0.5,0.489600388228296)
--(axis cs:1,0.251098593741174)
--(axis cs:1,0.266370478176669)
--(axis cs:1,0.266370478176669)
--(axis cs:0.5,0.509935819557404)
--(axis cs:0.25,0.680997115059614)
--(axis cs:0.1,0.764533272599073)
--(axis cs:0.05,0.791785093453973)
--(axis cs:0.01,0.803317217691207)
--(axis cs:0,0.82346826302162)
--cycle;

\path [draw=color2, fill=color2, opacity=0.2]
(axis cs:0,0.999258778640362)
--(axis cs:0,0.994571321606181)
--(axis cs:0.01,0.946150721205418)
--(axis cs:0.05,0.855005974604623)
--(axis cs:0.1,0.740098199757691)
--(axis cs:0.25,0.52508214035178)
--(axis cs:0.5,0.311255108556013)
--(axis cs:1,0.0405876946852538)
--(axis cs:1,0.0504530640685191)
--(axis cs:1,0.0504530640685191)
--(axis cs:0.5,0.351473832995703)
--(axis cs:0.25,0.550249477813396)
--(axis cs:0.1,0.75575864177392)
--(axis cs:0.05,0.864871459105809)
--(axis cs:0.01,0.950479682430264)
--(axis cs:0,0.999258778640362)
--cycle;

\addplot [line width=1pt, black, mark=*, mark size=3, mark options={solid,draw=white}]
table {%
0 1
0.01 0.969532065263505
0.05 0.914678861879259
0.1 0.869597290871642
0.25 0.670508546059215
0.5 0.0250395680411839
1 0.174431686175543
};
\addlegendentry{Global stats.}
\addplot [line width=1pt, color0, mark=triangle*, mark size=4, mark options={solid,rotate=180,draw=white}]
table {%
0 1
0.01 0.963760815321191
0.05 0.899201843915289
0.1 0.823737947879117
0.25 0.547563728974026
0.5 0.237074395941527
1 0.0812419869877402
};
\addlegendentry{Manual feat.}
\addplot [line width=1pt, color1, mark=triangle*, mark size=4, mark options={solid,draw=white}]
table {%
0 0.81069053767475
0.01 0.798427226308332
0.05 0.786863258279745
0.1 0.759219208185446
0.25 0.674232115138313
0.5 0.499560128742688
1 0.25965991074982
};
\addlegendentry{GCC}
\addplot [line width=1pt, color2, mark=square*, mark size=3, mark options={solid,draw=white}]
table {%
0 0.997448072053468
0.01 0.94828388045014
0.05 0.859909610428327
0.1 0.747876322873772
0.25 0.537782633325452
0.5 0.329456558508715
1 0.0455616954115764
};
\addlegendentry{GraphWave}
\end{axis}

\end{tikzpicture}

%% file: figures/tex/synthetic_rewire/frac_rewired_star_counts_F1_qual_rel_score_mcs_2.tex
\begin{tikzpicture}

\definecolor{color0}{rgb}{0.105882352941176,0.619607843137255,0.466666666666667}
\definecolor{color1}{rgb}{0.850980392156863,0.372549019607843,0.00784313725490196}
\definecolor{color2}{rgb}{0.458823529411765,0.43921568627451,0.701960784313725}

\begin{axis}[
axis line style={white!80!black},
legend cell align={left},
legend style={fill opacity=0.8, draw opacity=1, text opacity=1, draw=none},
tick align=outside,
tick pos=left,
x grid style={white!80!black},
xlabel={Fraction edges rewired},
xmajorgrids,
xmin=-0.05, xmax=1.05,
xtick style={color=white!15!black},
xtick={-0.2,0,0.2,0.4,0.6,0.8,1,1.2},
xticklabels={−0.2,0.0,0.2,0.4,0.6,0.8,1.0,1.2},
y grid style={white!80!black},
ylabel={Similarity score},
ymajorgrids,
ymin=-0.05, ymax=1.05,
ytick style={color=white!15!black},
ytick={-0.2,0,0.2,0.4,0.6,0.8,1,1.2},
yticklabels={−0.2,0.0,0.2,0.4,0.6,0.8,1.0,1.2}
]
\path [draw=black, fill=black, opacity=0.2]
(axis cs:0,1)
--(axis cs:0,1)
--(axis cs:0.01,0.975820336334377)
--(axis cs:0.05,0.766842425013876)
--(axis cs:0.1,0.314782800430164)
--(axis cs:0.25,0.293189318467202)
--(axis cs:0.5,0.0549698081676743)
--(axis cs:1,0)
--(axis cs:1,0)
--(axis cs:1,0)
--(axis cs:0.5,0.386587426104397)
--(axis cs:0.25,0.759971904688599)
--(axis cs:0.1,0.811813633733332)
--(axis cs:0.05,0.964966900463716)
--(axis cs:0.01,0.982006850558652)
--(axis cs:0,1)
--cycle;

\path [draw=color0, fill=color0, opacity=0.2]
(axis cs:0,1)
--(axis cs:0,1)
--(axis cs:0.01,0.988576809250022)
--(axis cs:0.05,0.972601004411987)
--(axis cs:0.1,0.93566298387258)
--(axis cs:0.25,0.752429377263925)
--(axis cs:0.5,0.420753768194978)
--(axis cs:1,0.0947213163381487)
--(axis cs:1,0.111493492355488)
--(axis cs:1,0.111493492355488)
--(axis cs:0.5,0.442894663305572)
--(axis cs:0.25,0.772517887451307)
--(axis cs:0.1,0.943009500527618)
--(axis cs:0.05,0.976647815512532)
--(axis cs:0.01,0.990602637982413)
--(axis cs:0,1)
--cycle;

\path [draw=color1, fill=color1, opacity=0.2]
(axis cs:0,0.882657469424493)
--(axis cs:0,0.862539397001047)
--(axis cs:0.01,0.860304793334321)
--(axis cs:0.05,0.850170353853756)
--(axis cs:0.1,0.824995834418318)
--(axis cs:0.25,0.724282620004335)
--(axis cs:0.5,0.545503801363022)
--(axis cs:1,0.274234596168555)
--(axis cs:1,0.294486500114278)
--(axis cs:1,0.294486500114278)
--(axis cs:0.5,0.564899829551983)
--(axis cs:0.25,0.734953544066758)
--(axis cs:0.1,0.836370945001342)
--(axis cs:0.05,0.859356780431708)
--(axis cs:0.01,0.865981863065615)
--(axis cs:0,0.882657469424493)
--cycle;

\path [draw=color2, fill=color2, opacity=0.2]
(axis cs:0,1)
--(axis cs:0,0.999443826462724)
--(axis cs:0.01,0.910757478031169)
--(axis cs:0.05,0.837011813933992)
--(axis cs:0.1,0.680298696659936)
--(axis cs:0.25,0.316439681360832)
--(axis cs:0.5,0.0613446034758427)
--(axis cs:1,0.00408493951060559)
--(axis cs:1,0.00734079414378548)
--(axis cs:1,0.00734079414378548)
--(axis cs:0.5,0.0781492398519219)
--(axis cs:0.25,0.3431820769542)
--(axis cs:0.1,0.703323913411392)
--(axis cs:0.05,0.852487790437513)
--(axis cs:0.01,0.918458364839466)
--(axis cs:0,1)
--cycle;

\addplot [line width=1pt, black, mark=*, mark size=3, mark options={solid,draw=white}]
table {%
0 1
0.01 0.979074593048863
0.05 0.896963026075371
0.1 0.565215987162407
0.25 0.527217952775744
0.5 0.220334085620527
1 0
};
\addlegendentry{Global stats.}
\addplot [line width=1pt, color0, mark=triangle*, mark size=4, mark options={solid,rotate=180,draw=white}]
table {%
0 1
0.01 0.989608421942838
0.05 0.974757667860287
0.1 0.939097358549671
0.25 0.762106412175217
0.5 0.431457803161162
1 0.102597614188293
};
\addlegendentry{Manual feat.}
\addplot [line width=1pt, color1, mark=triangle*, mark size=4, mark options={solid,draw=white}]
table {%
0 0.872512335991058
0.01 0.863223425657141
0.05 0.854885694785942
0.1 0.830637312032184
0.25 0.729571412534191
0.5 0.55560004467221
1 0.284434460505283
};
\addlegendentry{GCC}
\addplot [line width=1pt, color2, mark=square*, mark size=3, mark options={solid,draw=white}]
table {%
0 0.999814608820908
0.01 0.91454114279243
0.05 0.845083607630062
0.1 0.692173104636682
0.25 0.330088738435741
0.5 0.0699267576054999
1 0.00562547528057131
};
\addlegendentry{GraphWave}
\end{axis}

\end{tikzpicture}

%% file: figures/tex/synthetic_rewire/frac_rewired_wheel_counts_F1_qual_rel_score_mcs_2.tex
\begin{tikzpicture}

\definecolor{color0}{rgb}{0.105882352941176,0.619607843137255,0.466666666666667}
\definecolor{color1}{rgb}{0.850980392156863,0.372549019607843,0.00784313725490196}
\definecolor{color2}{rgb}{0.458823529411765,0.43921568627451,0.701960784313725}

\begin{axis}[
axis line style={white!80!black},
legend cell align={left},
legend style={fill opacity=0.8, draw opacity=1, text opacity=1, draw=none},
tick align=outside,
tick pos=left,
x grid style={white!80!black},
xlabel={Fraction edges rewired},
xmajorgrids,
xmin=-0.05, xmax=1.05,
xtick style={color=white!15!black},
xtick={-0.2,0,0.2,0.4,0.6,0.8,1,1.2},
xticklabels={−0.2,0.0,0.2,0.4,0.6,0.8,1.0,1.2},
y grid style={white!80!black},
ylabel={Similarity score},
ymajorgrids,
ymin=-0.05, ymax=1.05,
ytick style={color=white!15!black},
ytick={-0.2,0,0.2,0.4,0.6,0.8,1,1.2},
yticklabels={−0.2,0.0,0.2,0.4,0.6,0.8,1.0,1.2}
]
\path [draw=black, fill=black, opacity=0.2]
(axis cs:0,1)
--(axis cs:0,1)
--(axis cs:0.01,0.986377043269554)
--(axis cs:0.05,0.944079712644922)
--(axis cs:0.1,0.850748328973916)
--(axis cs:0.25,0.740395606732396)
--(axis cs:0.5,0.52770387112104)
--(axis cs:1,0.034462845710456)
--(axis cs:1,0.0444350982176521)
--(axis cs:1,0.0444350982176521)
--(axis cs:0.5,0.551475492353372)
--(axis cs:0.25,0.747949850456112)
--(axis cs:0.1,0.879749181237904)
--(axis cs:0.05,0.951673446066714)
--(axis cs:0.01,0.988187433122725)
--(axis cs:0,1)
--cycle;

\path [draw=color0, fill=color0, opacity=0.2]
(axis cs:0,1)
--(axis cs:0,1)
--(axis cs:0.01,0.931613032229805)
--(axis cs:0.05,0.783469909924584)
--(axis cs:0.1,0.575672166915339)
--(axis cs:0.25,0.267749833276623)
--(axis cs:0.5,0.0864307530131902)
--(axis cs:1,0.0162126182201025)
--(axis cs:1,0.021943873470723)
--(axis cs:1,0.021943873470723)
--(axis cs:0.5,0.109466842793478)
--(axis cs:0.25,0.282663554580811)
--(axis cs:0.1,0.601680418795965)
--(axis cs:0.05,0.805122553916048)
--(axis cs:0.01,0.936697341007028)
--(axis cs:0,1)
--cycle;

\path [draw=color1, fill=color1, opacity=0.2]
(axis cs:0,0.773177878061203)
--(axis cs:0,0.743182431100982)
--(axis cs:0.01,0.742587280554245)
--(axis cs:0.05,0.691328527219814)
--(axis cs:0.1,0.612978733541303)
--(axis cs:0.25,0.435937089403263)
--(axis cs:0.5,0.260617686467754)
--(axis cs:1,0.105303602941701)
--(axis cs:1,0.119580542836625)
--(axis cs:1,0.119580542836625)
--(axis cs:0.5,0.277811377768691)
--(axis cs:0.25,0.452956081841453)
--(axis cs:0.1,0.631944454089953)
--(axis cs:0.05,0.708390412800024)
--(axis cs:0.01,0.754005529347118)
--(axis cs:0,0.773177878061203)
--cycle;

\path [draw=color2, fill=color2, opacity=0.2]
(axis cs:0,0.998327759197324)
--(axis cs:0,0.995152739786651)
--(axis cs:0.01,0.836832288210084)
--(axis cs:0.05,0.55864402740851)
--(axis cs:0.1,0.364669244897659)
--(axis cs:0.25,0.142342003119948)
--(axis cs:0.5,0.0392560586561394)
--(axis cs:1,0.00480570627557074)
--(axis cs:1,0.00810474240768208)
--(axis cs:1,0.00810474240768208)
--(axis cs:0.5,0.0514891069131303)
--(axis cs:0.25,0.170774001162077)
--(axis cs:0.1,0.383211283112309)
--(axis cs:0.05,0.584421359003295)
--(axis cs:0.01,0.848810020406885)
--(axis cs:0,0.998327759197324)
--cycle;

\addplot [line width=1pt, black, mark=*, mark size=3, mark options={solid,draw=white}]
table {%
0 1
0.01 0.987314186972972
0.05 0.948085154632807
0.1 0.863974719831894
0.25 0.744471578093861
0.5 0.538575694715956
1 0.0395868609001286
};
\addlegendentry{Global stats.}
\addplot [line width=1pt, color0, mark=triangle*, mark size=4, mark options={solid,rotate=180,draw=white}]
table {%
0 1
0.01 0.934205081150278
0.05 0.794444949769759
0.1 0.588608883750729
0.25 0.274976947860417
0.5 0.0982161143213701
1 0.0188520492841856
};
\addlegendentry{Manual feat.}
\addplot [line width=1pt, color1, mark=triangle*, mark size=4, mark options={solid,draw=white}]
table {%
0 0.759801931657652
0.01 0.748176296526399
0.05 0.700507082093317
0.1 0.621889788619139
0.25 0.444325969316694
0.5 0.26992656806187
1 0.112100177536397
};
\addlegendentry{GCC}
\addplot [line width=1pt, color2, mark=square*, mark size=3, mark options={solid,draw=white}]
table {%
0 0.996837091428444
0.01 0.842375047642133
0.05 0.571917387554785
0.1 0.373969063049765
0.25 0.156908409360552
0.5 0.0451338261660876
1 0.00642306521472264
};
\addlegendentry{GraphWave}
\end{axis}

\end{tikzpicture}

%% file: figures/tex/synthetic_rewire/frac_rewired_samplededges_marked_F1_qual_rel_score_mcs_2.tex
\begin{tikzpicture}

\definecolor{color0}{rgb}{0.105882352941176,0.619607843137255,0.466666666666667}
\definecolor{color1}{rgb}{0.850980392156863,0.372549019607843,0.00784313725490196}
\definecolor{color2}{rgb}{0.458823529411765,0.43921568627451,0.701960784313725}

\begin{axis}[
axis line style={white!80!black},
legend cell align={left},
legend style={fill opacity=0.8, draw opacity=1, text opacity=1, draw=none},
tick align=outside,
tick pos=left,
x grid style={white!80!black},
xlabel={Fraction edges rewired},
xmajorgrids,
xmin=-0.05, xmax=1.05,
xtick style={color=white!15!black},
xtick={-0.2,0,0.2,0.4,0.6,0.8,1,1.2},
xticklabels={−0.2,0.0,0.2,0.4,0.6,0.8,1.0,1.2},
y grid style={white!80!black},
ylabel={Similarity score},
ymajorgrids,
ymin=-0.05, ymax=1.05,
ytick style={color=white!15!black},
ytick={-0.2,0,0.2,0.4,0.6,0.8,1,1.2},
yticklabels={−0.2,0.0,0.2,0.4,0.6,0.8,1.0,1.2}
]
\path [draw=black, fill=black, opacity=0.2]
(axis cs:0,1)
--(axis cs:0,1)
--(axis cs:0.01,0.964642194044375)
--(axis cs:0.05,0.903920445093041)
--(axis cs:0.1,0.862424653136414)
--(axis cs:0.25,0.649409610991528)
--(axis cs:0.5,0.00558206798745324)
--(axis cs:1,0.146221875439402)
--(axis cs:1,0.205088373771649)
--(axis cs:1,0.205088373771649)
--(axis cs:0.5,0.047812446696528)
--(axis cs:0.25,0.692503133067439)
--(axis cs:0.1,0.8759748526055)
--(axis cs:0.05,0.924316571657076)
--(axis cs:0.01,0.974206216702584)
--(axis cs:0,1)
--cycle;

\path [draw=color0, fill=color0, opacity=0.2]
(axis cs:0,1)
--(axis cs:0,1)
--(axis cs:0.01,0.878757070277545)
--(axis cs:0.05,0.712955542836495)
--(axis cs:0.1,0.538270475671664)
--(axis cs:0.25,0.162149346621193)
--(axis cs:0.5,0)
--(axis cs:1,0)
--(axis cs:1,0)
--(axis cs:1,0)
--(axis cs:0.5,0.021444803578179)
--(axis cs:0.25,0.23150320089691)
--(axis cs:0.1,0.655103578010859)
--(axis cs:0.05,0.82799830017275)
--(axis cs:0.01,0.926526285096652)
--(axis cs:0,1)
--cycle;

\path [draw=color1, fill=color1, opacity=0.2]
(axis cs:0,0.777777777777778)
--(axis cs:0,0.661016949152542)
--(axis cs:0.01,0.645753748827141)
--(axis cs:0.05,0.638877173398257)
--(axis cs:0.1,0.575433908337497)
--(axis cs:0.25,0.394516928278375)
--(axis cs:0.5,0.0734014489861859)
--(axis cs:1,0.0392631847479832)
--(axis cs:1,0.120110152571063)
--(axis cs:1,0.120110152571063)
--(axis cs:0.5,0.153098870434641)
--(axis cs:0.25,0.471529089073111)
--(axis cs:0.1,0.653318087944964)
--(axis cs:0.05,0.726499993229661)
--(axis cs:0.01,0.732798224088363)
--(axis cs:0,0.777777777777778)
--cycle;

\path [draw=color2, fill=color2, opacity=0.2]
(axis cs:0,1)
--(axis cs:0,0.982758620689655)
--(axis cs:0.01,0.867415847767422)
--(axis cs:0.05,0.772961306736317)
--(axis cs:0.1,0.590648905671789)
--(axis cs:0.25,0.335492224661466)
--(axis cs:0.5,0.116339763316026)
--(axis cs:1,0)
--(axis cs:1,0)
--(axis cs:1,0)
--(axis cs:0.5,0.165137449630382)
--(axis cs:0.25,0.428508220358637)
--(axis cs:0.1,0.674109457535624)
--(axis cs:0.05,0.849478914405493)
--(axis cs:0.01,0.91842044403281)
--(axis cs:0,1)
--cycle;

\addplot [line width=1pt, black, mark=*, mark size=3, mark options={solid,draw=white}]
table {%
0 1
0.01 0.969532065263505
0.05 0.914678861879259
0.1 0.869597290871642
0.25 0.670508546059215
0.5 0.0250395680411839
1 0.174431686175543
};
\addlegendentry{Global stats.}
\addplot [line width=1pt, color0, mark=triangle*, mark size=4, mark options={solid,rotate=180,draw=white}]
table {%
0 1
0.01 0.903590378641334
0.05 0.774222024411916
0.1 0.597177264068061
0.25 0.197657799010477
0.5 0.00859854810740742
1 0
};
\addlegendentry{Manual feat.}
\addplot [line width=1pt, color1, mark=triangle*, mark size=4, mark options={solid,draw=white}]
table {%
0 0.731675804636977
0.01 0.688417800850711
0.05 0.682527639959027
0.1 0.612384165858724
0.25 0.43374810460541
0.5 0.114561472790123
1 0.0791406987662209
};
\addlegendentry{GCC}
\addplot [line width=1pt, color2, mark=square*, mark size=3, mark options={solid,draw=white}]
table {%
0 0.994252873563218
0.01 0.894194631166528
0.05 0.810283497445754
0.1 0.632338092840697
0.25 0.387745255077151
0.5 0.14054048459296
1 0
};
\addlegendentry{GraphWave}
\end{axis}

\end{tikzpicture}

%% file: main.bbl
\begin{thebibliography}{43}
\providecommand{\natexlab}[1]{#1}
\providecommand{\url}[1]{\texttt{#1}}
\expandafter\ifx\csname urlstyle\endcsname\relax
  \providecommand{\doi}[1]{doi: #1}\else
  \providecommand{\doi}{doi: \begingroup \urlstyle{rm}\Url}\fi

\bibitem[Adamic \& Glance(2005)Adamic and Glance]{divided_blog}
Adamic, L.~A. and Glance, N.
\newblock The political blogosphere and the 2004 u.s. election: Divided they
  blog.
\newblock In \emph{Proceedings of the 3rd International Workshop on Link
  Discovery}, LinkKDD '05, pp.\  36–43, New York, NY, USA, 2005. Association
  for Computing Machinery.
\newblock ISBN 1595932151.
\newblock \doi{10.1145/1134271.1134277}.
\newblock URL \url{https://doi.org/10.1145/1134271.1134277}.

\bibitem[Akoglu et~al.(2010)Akoglu, McGlohon, and
  Faloutsos]{akoglu_oddball_2010}
Akoglu, L., McGlohon, M., and Faloutsos, C.
\newblock oddball: {Spotting} {Anomalies} in {Weighted} {Graphs}.
\newblock In \emph{{PAKDD}}, 2010.
\newblock \doi{10.1007/978-3-642-13672-6_40}.

\bibitem[Bojchevski et~al.(2018)Bojchevski, Shchur, Zügner, and
  Günnemann]{bojchevski_netgan_2018}
Bojchevski, A., Shchur, O., Zügner, D., and Günnemann, S.
\newblock {NetGAN}: {Generating} {Graphs} via {Random} {Walks}.
\newblock In \emph{International {Conference} on {Machine} {Learning}}, pp.\
  610--619. PMLR, July 2018.
\newblock URL \url{http://proceedings.mlr.press/v80/bojchevski18a.html}.
\newblock ISSN: 2640-3498.

\bibitem[Bunke \& Shearer(1998)Bunke and Shearer]{BUNKE1998255}
Bunke, H. and Shearer, K.
\newblock A graph distance metric based on the maximal common subgraph.
\newblock \emph{Pattern Recognition Letters}, 19\penalty0 (3):\penalty0
  255--259, 1998.
\newblock ISSN 0167-8655.
\newblock \doi{https://doi.org/10.1016/S0167-8655(97)00179-7}.
\newblock URL
  \url{https://www.sciencedirect.com/science/article/pii/S0167865597001797}.

\bibitem[Chen et~al.(2020)Chen, Jacob, and Mairal]{chen2020convolutional}
Chen, D., Jacob, L., and Mairal, J.
\newblock Convolutional kernel networks for graph-structured data.
\newblock In \emph{International Conference on Machine Learning}, pp.\
  1576--1586. PMLR, 2020.

\bibitem[Chiang et~al.(2019)Chiang, Liu, Si, Li, Bengio, and
  Hsieh]{chiang2019cluster}
Chiang, W.-L., Liu, X., Si, S., Li, Y., Bengio, S., and Hsieh, C.-J.
\newblock Cluster-gcn: An efficient algorithm for training deep and large graph
  convolutional networks.
\newblock In \emph{Proceedings of the 25th ACM SIGKDD International Conference
  on Knowledge Discovery \& Data Mining}, pp.\  257--266, 2019.

\bibitem[Donnat et~al.(2018)Donnat, Zitnik, Hallac, and
  Leskovec]{donnat_learning_2018}
Donnat, C., Zitnik, M., Hallac, D., and Leskovec, J.
\newblock Learning {Structural} {Node} {Embeddings} {Via} {Diffusion}
  {Wavelets}.
\newblock In \emph{Proceedings of the 24th ACM SIGKDD International Conference
  on Knowledge Discovery \& Data Mining}, {KDD}'18, pp.\  1320--1329, New York,
  NY, USA, July 2018. Association for Computing Machinery.
\newblock \doi{10.1145/3219819.3220025}.
\newblock URL \url{http://arxiv.org/abs/1710.10321}.
\newblock arXiv: 1710.10321.

\bibitem[Everett et~al.(1990)Everett, Boyd, and Borgatti]{everett1990ego}
Everett, M.~G., Boyd, J.~P., and Borgatti, S.~P.
\newblock Ego-centered and local roles: A graph theoretic approach.
\newblock \emph{Journal of Mathematical Sociology}, 15\penalty0 (3-4):\penalty0
  163--172, 1990.

\bibitem[Faloutsos et~al.(2011)Faloutsos, Faloutsos, and
  Faloutsos]{powerlaw_Faloutsos_11}
Faloutsos, M., Faloutsos, P., and Faloutsos, C.
\newblock \emph{On power-law relationships of the internet topology}, pp.\
  195--206.
\newblock Princeton University Press, 2011.
\newblock \doi{doi:10.1515/9781400841356.195}.
\newblock URL \url{https://doi.org/10.1515/9781400841356.195}.

\bibitem[Gao et~al.(2010)Gao, Xiao, Tao, and Li]{gao2010survey}
Gao, X., Xiao, B., Tao, D., and Li, X.
\newblock A survey of graph edit distance.
\newblock \emph{Pattern Analysis and applications}, 13\penalty0 (1):\penalty0
  113--129, 2010.

\bibitem[Grover \& Leskovec(2016)Grover and Leskovec]{grover_node2vec_2016}
Grover, A. and Leskovec, J.
\newblock node2vec: {Scalable} {Feature} {Learning} for {Networks}.
\newblock In \emph{Proceedings of the 22nd {ACM} {SIGKDD} {International}
  {Conference} on {Knowledge} {Discovery} and {Data} {Mining} - {KDD} '16},
  pp.\  855--864, San Francisco, California, USA, 2016. ACM Press.
\newblock ISBN 978-1-4503-4232-2.
\newblock \doi{10.1145/2939672.2939754}.
\newblock URL \url{http://dl.acm.org/citation.cfm?doid=2939672.2939754}.

\bibitem[Hamilton et~al.(2017)Hamilton, Ying, and
  Leskovec]{graphsage_hamilton17}
Hamilton, W., Ying, Z., and Leskovec, J.
\newblock Inductive representation learning on large graphs.
\newblock In Guyon, I., Luxburg, U.~V., Bengio, S., Wallach, H., Fergus, R.,
  Vishwanathan, S., and Garnett, R. (eds.), \emph{Advances in Neural
  Information Processing Systems}, volume~30. Curran Associates, Inc., 2017.
\newblock URL
  \url{https://proceedings.neurips.cc/paper/2017/file/5dd9db5e033da9c6fb5ba83c7a7ebea9-Paper.pdf}.

\bibitem[Henderson et~al.(2011)Henderson, Gallagher, Li, Akoglu, Eliassi-Rad,
  Tong, and Faloutsos]{henderson_its_2011}
Henderson, K., Gallagher, B., Li, L., Akoglu, L., Eliassi-Rad, T., Tong, H.,
  and Faloutsos, C.
\newblock It's who you know: graph mining using recursive structural features.
\newblock In \emph{Proceedings of the 17th {ACM} {SIGKDD} international
  conference on {Knowledge} discovery and data mining}, {KDD} '11, pp.\
  663--671, New York, NY, USA, August 2011. Association for Computing
  Machinery.
\newblock ISBN 978-1-4503-0813-7.
\newblock \doi{10.1145/2020408.2020512}.
\newblock URL \url{https://doi.org/10.1145/2020408.2020512}.

\bibitem[Henderson et~al.(2012)Henderson, Gallagher, Eliassi-Rad, Tong, Basu,
  Akoglu, Koutra, Faloutsos, and Li]{henderson_rolx_2012}
Henderson, K., Gallagher, B., Eliassi-Rad, T., Tong, H., Basu, S., Akoglu, L.,
  Koutra, D., Faloutsos, C., and Li, L.
\newblock {RolX}: structural role extraction \& mining in large graphs.
\newblock In \emph{Proceedings of the 18th {ACM} {SIGKDD} international
  conference on {Knowledge} discovery and data mining}, {KDD} '12, pp.\
  1231--1239, New York, NY, USA, August 2012. Association for Computing
  Machinery.
\newblock ISBN 978-1-4503-1462-6.
\newblock \doi{10.1145/2339530.2339723}.
\newblock URL \url{https://doi.org/10.1145/2339530.2339723}.

\bibitem[Heusel et~al.(2017)Heusel, Ramsauer, Unterthiner, Nessler, and
  Hochreiter]{fid_heusel17}
Heusel, M., Ramsauer, H., Unterthiner, T., Nessler, B., and Hochreiter, S.
\newblock Gans trained by a two time-scale update rule converge to a local nash
  equilibrium.
\newblock In Guyon, I., Luxburg, U.~V., Bengio, S., Wallach, H., Fergus, R.,
  Vishwanathan, S., and Garnett, R. (eds.), \emph{Advances in Neural
  Information Processing Systems}, volume~30. Curran Associates, Inc., 2017.
\newblock URL
  \url{https://proceedings.neurips.cc/paper/2017/file/8a1d694707eb0fefe65871369074926d-Paper.pdf}.

\bibitem[Hočevar \& Demšar(2014)Hočevar and Demšar]{orbits_hocevar14}
Hočevar, T. and Demšar, J.
\newblock {A combinatorial approach to graphlet counting}.
\newblock \emph{Bioinformatics}, 30\penalty0 (4):\penalty0 559--565, 12 2014.
\newblock ISSN 1367-4803.
\newblock \doi{10.1093/bioinformatics/btt717}.
\newblock URL \url{https://doi.org/10.1093/bioinformatics/btt717}.

\bibitem[Hu et~al.(2020)Hu, Fey, Zitnik, Dong, Ren, Liu, Catasta, and
  Leskovec]{hu2020ogb}
Hu, W., Fey, M., Zitnik, M., Dong, Y., Ren, H., Liu, B., Catasta, M., and
  Leskovec, J.
\newblock Open graph benchmark: Datasets for machine learning on graphs.
\newblock \emph{arXiv preprint arXiv:2005.00687}, 2020.

\bibitem[Khrulkov \& Oseledets(2018)Khrulkov and
  Oseledets]{khrulkov2018geometry}
Khrulkov, V. and Oseledets, I.
\newblock Geometry score: A method for comparing generative adversarial
  networks.
\newblock In \emph{International Conference on Machine Learning}, pp.\
  2621--2629. PMLR, 2018.

\bibitem[Kipf \& Welling(2017)Kipf and Welling]{KipfW17}
Kipf, T.~N. and Welling, M.
\newblock Semi-supervised classification with graph convolutional networks.
\newblock In \emph{5th International Conference on Learning Representations,
  {ICLR} 2017, Toulon, France, April 24-26, 2017, Conference Track
  Proceedings}. OpenReview.net, 2017.
\newblock URL \url{https://openreview.net/forum?id=SJU4ayYgl}.

\bibitem[Kynk\"{a}\"{a}nniemi et~al.(2019)Kynk\"{a}\"{a}nniemi, Karras, Laine,
  Lehtinen, and Aila]{improved_prec_recall19}
Kynk\"{a}\"{a}nniemi, T., Karras, T., Laine, S., Lehtinen, J., and Aila, T.
\newblock Improved precision and recall metric for assessing generative models.
\newblock In Wallach, H., Larochelle, H., Beygelzimer, A., d\textquotesingle
  Alch\'{e}-Buc, F., Fox, E., and Garnett, R. (eds.), \emph{Advances in Neural
  Information Processing Systems}, volume~32. Curran Associates, Inc., 2019.
\newblock URL
  \url{https://proceedings.neurips.cc/paper/2019/file/0234c510bc6d908b28c70ff313743079-Paper.pdf}.

\bibitem[Liao et~al.(2019)Liao, Li, Song, Wang, Hamilton, Duvenaud, Urtasun,
  and Zemel]{gran_liao19}
Liao, R., Li, Y., Song, Y., Wang, S., Hamilton, W., Duvenaud, D.~K., Urtasun,
  R., and Zemel, R.
\newblock Efficient graph generation with graph recurrent attention networks.
\newblock In Wallach, H., Larochelle, H., Beygelzimer, A., d\textquotesingle
  Alch\'{e}-Buc, F., Fox, E., and Garnett, R. (eds.), \emph{Advances in Neural
  Information Processing Systems}, volume~32. Curran Associates, Inc., 2019.
\newblock URL
  \url{https://proceedings.neurips.cc/paper/2019/file/d0921d442ee91b896ad95059d13df618-Paper.pdf}.

\bibitem[Newman(2018)]{newman2018networks}
Newman, M.
\newblock \emph{Networks}.
\newblock Oxford university press, 2nd edition, 2018.
\newblock ISBN 978-0-19-880509-0.
\newblock \doi{https://doi.org/10.1093/oso/9780198805090.001.0001}.

\bibitem[Newman(2003{\natexlab{a}})]{newman2003structure}
Newman, M.~E.
\newblock The structure and function of complex networks.
\newblock \emph{SIAM review}, 45\penalty0 (2):\penalty0 167--256,
  2003{\natexlab{a}}.

\bibitem[Newman(2003{\natexlab{b}})]{newman2003mixing}
Newman, M. E.~J.
\newblock Mixing patterns in networks.
\newblock \emph{Phys. Rev. E}, 67:\penalty0 026126, Feb 2003{\natexlab{b}}.
\newblock \doi{10.1103/PhysRevE.67.026126}.
\newblock URL \url{https://link.aps.org/doi/10.1103/PhysRevE.67.026126}.

\bibitem[Nikolentzos et~al.(2021)Nikolentzos, Siglidis, and
  Vazirgiannis]{nikolentzos2021graph}
Nikolentzos, G., Siglidis, G., and Vazirgiannis, M.
\newblock Graph kernels: A survey.
\newblock \emph{Journal of Artificial Intelligence Research}, 72:\penalty0
  943--1027, 2021.

\bibitem[O'Bray et~al.(2021)O'Bray, Horn, Rieck, and
  Borgwardt]{obray2021evaluation}
O'Bray, L., Horn, M., Rieck, B., and Borgwardt, K.
\newblock Evaluation metrics for graph generative models: Problems, pitfalls,
  and practical solutions, 2021.

\bibitem[Perozzi et~al.(2014)Perozzi, Al-Rfou, and
  Skiena]{perozzi_deepwalk_2014}
Perozzi, B., Al-Rfou, R., and Skiena, S.
\newblock {DeepWalk}: online learning of social representations.
\newblock In \emph{Proceedings of the 20th {ACM} {SIGKDD} international
  conference on {Knowledge} discovery and data mining}, {KDD} '14, pp.\
  701--710, New York, NY, USA, August 2014. Association for Computing
  Machinery.
\newblock ISBN 978-1-4503-2956-9.
\newblock \doi{10.1145/2623330.2623732}.
\newblock URL \url{https://doi.org/10.1145/2623330.2623732}.

\bibitem[Poklukar et~al.(2021)Poklukar, Varava, and
  Kragic]{pmlr-v139-poklukar21a}
Poklukar, P., Varava, A., and Kragic, D.
\newblock Geomca: Geometric evaluation of data representations.
\newblock In Meila, M. and Zhang, T. (eds.), \emph{Proceedings of the 38th
  International Conference on Machine Learning}, volume 139 of
  \emph{Proceedings of Machine Learning Research}, pp.\  8588--8598. PMLR,
  18--24 Jul 2021.
\newblock URL \url{https://proceedings.mlr.press/v139/poklukar21a.html}.

\bibitem[Poklukar et~al.(2022)Poklukar, Polianskii, Varava, Pokorny, and
  Jensfelt]{anonymous2022delaunay}
Poklukar, P., Polianskii, V., Varava, A., Pokorny, F.~T., and Jensfelt, D.~K.
\newblock Delaunay component analysis for evaluation of data representations.
\newblock In \emph{International Conference on Learning Representations}, 2022.
\newblock URL \url{https://openreview.net/forum?id=HTVch9AMPa}.

\bibitem[Qiu et~al.(2018)Qiu, Dong, Ma, Li, Wang, and Tang]{qiu_network_2018}
Qiu, J., Dong, Y., Ma, H., Li, J., Wang, K., and Tang, J.
\newblock Network {Embedding} as {Matrix} {Factorization}: {Unifying}
  {DeepWalk}, {LINE}, {PTE}, and {Node2vec}.
\newblock In \emph{Proceedings of the eleventh ACM international conference on
  web search and data mining}, {WSDM} '18, pp.\  459--467, New York, NY, USA,
  2018. Association for Computing Machinery.
\newblock ISBN 978-1-4503-5581-0.
\newblock \doi{10.1145/3159652.3159706}.
\newblock URL \url{https://doi.org/10.1145/3159652.3159706}.
\newblock event-place: Marina Del Rey, CA, USA.

\bibitem[Qiu et~al.(2020)Qiu, Chen, Dong, Zhang, Yang, Ding, Wang, and
  Tang]{qiu_gcc_2020}
Qiu, J., Chen, Q., Dong, Y., Zhang, J., Yang, H., Ding, M., Wang, K., and Tang,
  J.
\newblock {GCC}: {Graph} {Contrastive} {Coding} for {Graph} {Neural} {Network}
  {Pre}-{Training}.
\newblock In \emph{Proceedings of the 26th {ACM} {SIGKDD} {International}
  {Conference} on {Knowledge} {Discovery} \& {Data} {Mining}}, {KDD}'20, pp.\
  1150--1160. Association for Computing Machinery, New York, NY, USA, August
  2020.
\newblock ISBN 978-1-4503-7998-4.
\newblock URL \url{https://doi.org/10.1145/3394486.3403168}.

\bibitem[Rendsburg et~al.(2020)Rendsburg, Heidrich, and
  Luxburg]{rendsburg_netgan_2020}
Rendsburg, L., Heidrich, H., and Luxburg, U.~V.
\newblock {NetGAN} without {GAN}: {From} {Random} {Walks} to {Low}-{Rank}
  {Approximations}.
\newblock In III, H.~D. and Singh, A. (eds.), \emph{Proceedings of the 37th
  {International} {Conference} on {Machine} {Learning}}, volume 119 of
  \emph{Proceedings of {Machine} {Learning} {Research}}, pp.\  8073--8082.
  PMLR, July 2020.
\newblock URL \url{http://proceedings.mlr.press/v119/rendsburg20a.html}.

\bibitem[Ribeiro et~al.(2017)Ribeiro, Saverese, and
  Figueiredo]{ribeiro2017struc2vec}
Ribeiro, L.~F., Saverese, P.~H., and Figueiredo, D.~R.
\newblock struc2vec: Learning node representations from structural identity.
\newblock In \emph{Proceedings of the 23rd ACM SIGKDD international conference
  on knowledge discovery and data mining}, pp.\  385--394, 2017.

\bibitem[Rossi \& Ahmed(2015)Rossi and Ahmed]{rossi_role_2015}
Rossi, R.~A. and Ahmed, N.~K.
\newblock Role {Discovery} in {Networks}.
\newblock \emph{IEEE Transactions on Knowledge and Data Engineering},
  27\penalty0 (4):\penalty0 1112--1131, April 2015.
\newblock ISSN 1558-2191.
\newblock \doi{10.1109/TKDE.2014.2349913}.
\newblock Conference Name: IEEE Transactions on Knowledge and Data Engineering.

\bibitem[Sajjadi et~al.(2018)Sajjadi, Bachem, Lucic, Bousquet, and
  Gelly]{prec_recall_sajjadi18}
Sajjadi, M. S.~M., Bachem, O., Lucic, M., Bousquet, O., and Gelly, S.
\newblock Assessing generative models via precision and recall.
\newblock In Bengio, S., Wallach, H., Larochelle, H., Grauman, K.,
  Cesa-Bianchi, N., and Garnett, R. (eds.), \emph{Advances in Neural
  Information Processing Systems}, volume~31. Curran Associates, Inc., 2018.
\newblock URL
  \url{https://proceedings.neurips.cc/paper/2018/file/f7696a9b362ac5a51c3dc8f098b73923-Paper.pdf}.

\bibitem[Salimans et~al.(2016)Salimans, Goodfellow, Zaremba, Cheung, Radford,
  and Chen]{salimans2016improved}
Salimans, T., Goodfellow, I., Zaremba, W., Cheung, V., Radford, A., and Chen,
  X.
\newblock Improved techniques for training gans.
\newblock \emph{Advances in neural information processing systems},
  29:\penalty0 2234--2242, 2016.

\bibitem[Sanfeliu \& Fu(1983)Sanfeliu and Fu]{Sanfeliu_graph_edit83}
Sanfeliu, A. and Fu, K.-S.
\newblock A distance measure between attributed relational graphs for pattern
  recognition.
\newblock \emph{IEEE Transactions on Systems, Man, and Cybernetics},
  SMC-13\penalty0 (3):\penalty0 353--362, 1983.
\newblock \doi{10.1109/TSMC.1983.6313167}.

\bibitem[Sarajli{\'c} et~al.(2016)Sarajli{\'c}, Malod-Dognin, Yavero{\u{g}}lu,
  and Pr{\v{z}}ulj]{sarajlic2016graphlet}
Sarajli{\'c}, A., Malod-Dognin, N., Yavero{\u{g}}lu, {\"O}.~N., and
  Pr{\v{z}}ulj, N.
\newblock Graphlet-based characterization of directed networks.
\newblock \emph{Scientific reports}, 6\penalty0 (1):\penalty0 1--14, 2016.

\bibitem[Shervashidze et~al.(2011)Shervashidze, Schweitzer, Van~Leeuwen,
  Mehlhorn, and Borgwardt]{shervashidze2011weisfeiler}
Shervashidze, N., Schweitzer, P., Van~Leeuwen, E.~J., Mehlhorn, K., and
  Borgwardt, K.~M.
\newblock Weisfeiler-lehman graph kernels.
\newblock \emph{Journal of Machine Learning Research}, 12\penalty0 (9), 2011.
\newblock URL
  \url{https://www.jmlr.org/papers/volume12/shervashidze11a/shervashidze11a.pdf}.

\bibitem[Togninalli et~al.(2019)Togninalli, Ghisu, Llinares-L\'{o}pez, Rieck,
  and Borgwardt]{wwl_graph_kernel_togninalli19}
Togninalli, M., Ghisu, E., Llinares-L\'{o}pez, F., Rieck, B., and Borgwardt, K.
\newblock Wasserstein weisfeiler-lehman graph kernels.
\newblock In Wallach, H., Larochelle, H., Beygelzimer, A., d\textquotesingle
  Alch\'{e}-Buc, F., Fox, E., and Garnett, R. (eds.), \emph{Advances in Neural
  Information Processing Systems}, volume~32. Curran Associates, Inc., 2019.
\newblock URL
  \url{https://proceedings.neurips.cc/paper/2019/file/73fed7fd472e502d8908794430511f4d-Paper.pdf}.

\bibitem[Veli{\v{c}}kovi{\'{c}} et~al.(2018)Veli{\v{c}}kovi{\'{c}}, Cucurull,
  Casanova, Romero, Li{\`{o}}, and Bengio]{velickovic2018graph}
Veli{\v{c}}kovi{\'{c}}, P., Cucurull, G., Casanova, A., Romero, A., Li{\`{o}},
  P., and Bengio, Y.
\newblock {Graph Attention Networks}.
\newblock \emph{International Conference on Learning Representations}, 2018.
\newblock URL \url{https://openreview.net/forum?id=rJXMpikCZ}.

\bibitem[Vishwanathan et~al.(2010)Vishwanathan, Schraudolph, Kondor, and
  Borgwardt]{vishwanathan2010graph}
Vishwanathan, S. V.~N., Schraudolph, N.~N., Kondor, R., and Borgwardt, K.~M.
\newblock Graph kernels.
\newblock \emph{Journal of Machine Learning Research}, 11:\penalty0 1201--1242,
  2010.
\newblock URL
  \url{https://www.jmlr.org/papers/volume11/vishwanathan10a/vishwanathan10a.pdf}.

\bibitem[You et~al.(2018)You, Ying, Ren, Hamilton, and
  Leskovec]{you2018graphrnn}
You, J., Ying, R., Ren, X., Hamilton, W., and Leskovec, J.
\newblock Graphrnn: Generating realistic graphs with deep auto-regressive
  models.
\newblock In \emph{International {Conference} on {Machine} {Learning}}, pp.\
  5708--5717. PMLR, 2018.

\end{thebibliography}
